\documentclass{article}

\usepackage[nonatbib, final]{neurips_2025}
\usepackage[numbers,sort&compress]{natbib}
\usepackage{marvosym}

\usepackage[utf8]{inputenc} 
\usepackage[T1]{fontenc}    
\usepackage{hyperref}       
\usepackage{url}            
\usepackage{booktabs}       
\usepackage{amsfonts}       
\usepackage{nicefrac}       
\usepackage{microtype}      
\usepackage{wrapfig}
\usepackage{graphicx}
\usepackage{enumitem}
\usepackage{amsmath}
\usepackage{multirow}
\usepackage{rotating}

\usepackage[table]{xcolor}

\title{Reason-RFT: Reinforcement Fine-Tuning for Visual Reasoning of Vision Language Models}

\author{\small Huajie Tan$^{1,2,*}$, Yuheng Ji$^{2,3,4,*}$, Xiaoshuai Hao$^{2,*}$, Xiansheng Chen$^{2,*}$, \\
\small \textbf{Pengwei Wang$^{2,\dagger}$, Zhongyuan Wang$^2$, Shanghang Zhang$^{1,2,\text{\Letter}}$} \\
$^1$ \small State Key Laboratory of Multimedia Information Processing, School of Computer Science, Peking University \\ 
$^2$ \small Beijing Academy of Artificial Intelligence
$^3$ \small Institute of Automation, Chinese Academy of Sciences \\
$^4$ \small School of Artificial Intelligence, University of Chinese Academy of Sciences
}

\begin{document}

\maketitle

\let\thefootnote\relax\footnotetext{$^{*}$ Equal contribution. $^{\dagger}$ Project leader. $^{\text{\Letter}}$ Corresponding author: \href{shanghang@pku.edu.cn}{shanghang@pku.edu.cn}.}


\vspace{-0.5cm}
\begin{abstract}
Visual reasoning abilities play a crucial role in understanding complex multimodal data, advancing both domain-specific applications and artificial general intelligence (AGI).
Existing methods improve Vision-Language Models (VLMs) reasoning via Chain-of-Thought (CoT) supervised fine-tuning, using meticulously annotated training data to enhance visual reasoning capabilities.
However, this training paradigm may lead to overfitting and cognitive rigidity, restricting the model's generalization ability to transfer visual reasoning skills under domain shift and limiting its real-world applicability.
To address these limitations, we propose \textit{\textbf{Reason-RFT}}, the first two-stage reinforcement fine-tuning framework for visual reasoning: (1) Supervised Fine-Tuning (SFT) with curated CoT data activates the reasoning potential of VLMs, followed by (2) Group Relative Policy Optimization (GRPO)-based reinforcement learning that generates multiple reasoning-response pairs, significantly enhancing the capability to address ubiquitous domain shift in visual reasoning tasks.
To evaluate the visual reasoning capabilities of \textit{Reason-RFT}, we reconstructed a comprehensive dataset encompassing visual counting, structural perception, and spatial transformation, serving as a benchmark for systematic assessment across three core dimensions.
Experimental results demonstrate three key advantages: 
(1) \textit{{Performance Enhancement}}: achieving state-of-the-art results across multiple tasks, outperforming mainstream open-source and proprietary models; 
(2) \textit{{Generalization Superiority}}: consistently maintaining robust performance in addressing domain shift in typical visual reasoning tasks, outperforming alternative paradigms; 
(3) \textit{{Data Efficiency}}: excelling in few-shot learning scenarios while surpassing full-dataset SFT baselines. 
\textit{{Reason-RFT}} introduces a rebust training paradigm in visual reasoning, and please refer to project website: \href{https://tanhuajie.github.io/ReasonRFT/}{Reason-RFT}.
\end{abstract}
    
\section{Introduction}
\label{sec:intro}
Visual reasoning is pivotal for understanding complex multimodal data and advancing artificial general intelligence (AGI)~\cite{clevr-math,OpenAI2024o1}, making it a central focus in intelligent systems research. Recent advancements in image recognition~\cite{image_processing,tan2024joint,zhang2025beyond,ji2024advlora,ji2025enhancing,ji2023learning,lyu2025egoprompt}, interactive security~\cite{mu2023configurable,mu2023energy,bai2025alleviating} and scene understanding~\cite{cordts2016cityscapes,vsl_bench} have enabled transformative applications in healthcare~\cite{zhan2020medical,MedVLM}, robotics~\cite{robobrain,team2025robobrain,liu2024robomamba,tan2025roboos,zhou2025code,zhou2025roborefer,li2024foundation,song2025maniplvm}, and autonomous driving~\cite{HAO2025103018,hao2024your,hao2024mapdistill,hao2024mbfusion,hao2025msc,zhao2025fastrsr,hao2025really}. Consequently, enhancing visual reasoning capabilities has garnered significant attention from both industry and academia for its potential to drive transformative advancements.

Researchers have explored two primary categories of methods to enhance visual reasoning capabilities: (1) neural-symbolic methods~\cite{symbolic_computing,symbolic,neuro-symbolic,zhang2024take,visual_programming}, which integrate symbolic reasoning with neural networks to improve interpretability and modularity, and (2) Supervised Fine-Tuning (SFT) based on vision-language models (VLMs)~\cite{llava-cot,llamav-o1}, which utilize end-to-end training to strengthen reasoning abilities. However, both approaches face significant limitations. Neural-symbolic methods are hindered by high complexity and a strong reliance on program generation, while SFT is constrained by its dependence on high-quality Chain-of-Thought (CoT) annotated data and meticulously designed data mixing strategies, leading to issues such as overfitting, cognitive rigidity, and limited adaptability to domain shift. These challenges reduce their effectiveness in real-world applications.

Recent advances such as GPT-o1~\cite{OpenAI2024o1}, DeepSeek-R1~\cite{guo2025deepseek}, and Kimi-1.5~\cite{team2025kimi} show that reinforcement learning (RL) during post-training enhances reasoning in coding and mathematics. RL offers a dynamic alternative to SFT by enabling exploration and feedback-driven optimization, which can improve performance with limited labeled data. However, pure RL methods often lack robustness to domain shifts—such as changes in visual appearance or configuration, limiting their generalization capacity in real-world visual reasoning scenarios. 

\begin{figure*}[t]
    \centering
   \vspace{-0.2cm}
    \includegraphics[width=0.98\linewidth]{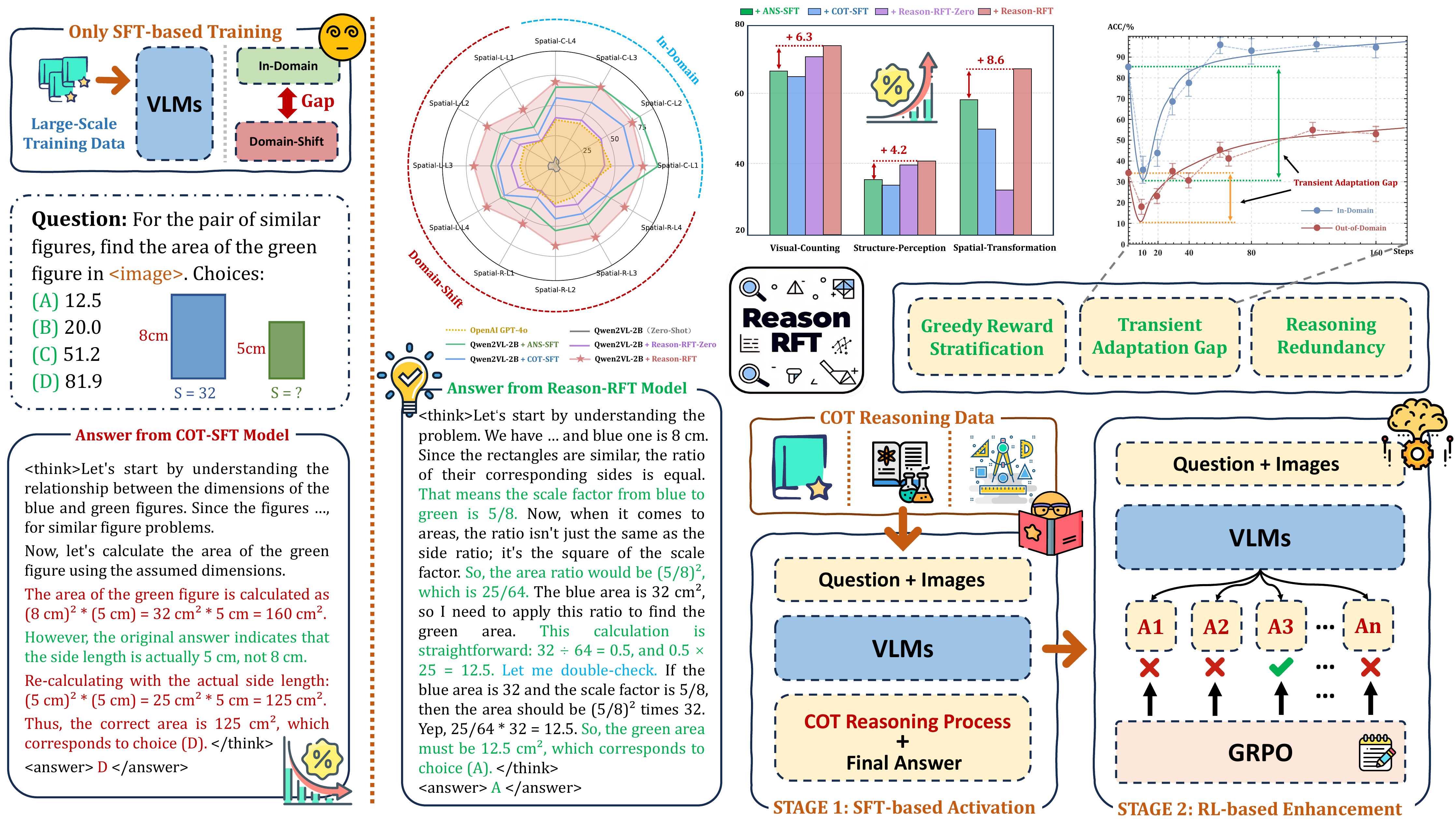}
    \caption{
    \textbf{Overview of Reason-RFT.} 
    Compared to traditional SFT-based methods, our proposed Reason-RFT framework demonstrates superior generalization in visual reasoning tasks, excelling in reasoning improvement, domain shift adaptability, and data efficiency.
    }
    \vspace{-0.5cm}
    \label{fig:intro}
\end{figure*}

To address this, we propose \textit{\textbf{Reason-RFT}}, the first two-stage reinforcement fine-tuning framework designed to enhance generalization in visual reasoning tasks. First, we employ SFT with CoT reasoning to activate the model's potential reasoning capabilities, using a high-quality domain-specific visual reasoning dataset tailored to stimulate related reasoning abilities. Subsequently, we further enhance reasoning potential through Group Relative Policy Optimization (GRPO), demonstrating that \textit{{Reason-RFT}} improves robustness under distribution shifts by enhancing the model's reasoning capabilities. To evaluate its effectiveness, we constructed a high-quality dataset covering visual counting, structure perception, and spatial transformation, serving as a benchmark for evaluating three core capabilities of visual reasoning.
Extensive experiments highlight three key advantages of \textit{Reason-RFT}:
(1) \textbf{\textit{Performance Improvement}}: It achieves strong results on visual reasoning tasks, including visual counting, structure perception, and spatial transformation, outperforming mainstream VLMs;
(2) \textbf{\textit{Enhanced Generalization}}: It consistently exceeds both SFT-only and RL-only baselines under domain shift conditions, as demonstrated through comprehensive evaluations;
(3) \textbf{\textit{Data Efficiency}}: It reaches over 90\% of the SFT-only performance while using less than 5\% of the data.
These results underscore the effectiveness and efficiency of \textit{Reason-RFT}, establishing it as a robust framework for advancing visual reasoning.
Our main contributions are summarized as follows.
\begin{itemize}[left=1em]
\item We introduce \textit{\textbf{Reason-RFT}}, a two-stage reinforcement fine-tuning framework that significantly enhances the visual reasoning capabilities of VLMs by effectively combining the complementary strengths of SFT-based and RL-based methods.

\item We provide a systematic analysis of SFT-based and RL-based paradigms on visual reasoning tasks, identifying the limitations of SFT and the advantages of RL in improving reasoning ability, handling domain shifts, and achieving data-efficient learning.

\item We reconstruct a comprehensive dataset spanning three core domains: visual counting, structure perception, and spatial transformation, serving as a benchmark for evaluating visual cognition, geometric understanding, and spatial generalization.

\item Extensive experiments validate the proposed framework, demonstrating its practicality and effectiveness, and providing a new perspective for reinforcement-driven multi-modal training.

\end{itemize}

\section{Related Work}
\label{sec:formatting}

\textbf{Visual Reasoning} 
Visual reasoning is a core challenge in advancing AGI, requiring models to perform complex cognitive tasks grounded in visual perception~\cite{clevr-math,mathvista,hao2023mixgen,ma2025followyourmotion,ma2024followyouremoji,ma2023magicstick,ma2024followpose,ma2022visual,yan2025eedit,feng2025follow,long2025follow}. It underpins a wide range of applications, including visual counting~\cite{clevr-math, super-clevr}, geometric problem-solving~\cite{geo170k, kazemi2023geomverse, mathvista, mavis, math360k}, visual transformation reasoning~\cite{hong2021transformation,ji2025visualtrans}, scientific analysis~\cite{sci, ai2d}, and robotic task planning~\cite{hu2023look, robobrain, hao2025tla}. Traditional approaches rely on program generation~\cite{programs, visual_programming, suris2023vipergpt} or neural-symbolic reasoning~\cite{symbolic_computing, symbolic, neuro-symbolic, zhang2024take}, while recent advances in VLMs leverage large language models (LLMs) to enhance reasoning capabilities. For instance, LLaVA-CoT~\cite{llava-cot} employs multi-stage SFT with CoT prompting~\cite{cot}, and Insight-V~\cite{insight-v} integrates SFT with RL. DeepSeek-R1-Zero~\cite{deepseek-r1} further introduces a rule-guided RL framework that substantially improves reasoning performance. Building upon the DeepSeek-R1~\cite{deepseek-r1}, our work provides a comparative analysis of SFT-based and RL-based paradigms, demonstrating the advantages of R1-style methods in enhancing visual reasoning.

\textbf{Post-Training}  
Post-Training is a crucial phase for enhancing the performance of LLMs and VLMs, bridging pre-trained models and their real-world applications~\cite{tang2025affordgrasp,zhang2025mapnav,post_training,post_training_2}. It primarily involves two methodologies: \textit{SFT}~\cite{cot_sft,math_sft} and \textit{RL}~\cite{rl_1,rlhf,rl_2,rl_math,rl_3}. SFT adapts pre-trained models to specific tasks using task-oriented datasets, often formatted as instructions. Research like FLAN~\cite{flan} highlights the importance of diverse instruction-tuning datasets for improving zero-shot performance, while iterative processes, such as Llama 3.1's six-round strategy~\cite{llama3_1}, integrate rejection sampling, synthetic data, and human annotations. RL aligns models with human preferences or task-specific goals through feedback mechanisms. Reinforcement Learning from Human Feedback (RLHF)~\cite{rlhf} refines models using human preference data, as seen in Llama 3.1~\cite{llama3_1} and Nemotron-4~\cite{nemotron}, which use reward modeling techniques like DPO~\cite{dpo} and RPO~\cite{nemotron}. For example, TÜLU3~\cite{tulu3} employs length-normalized DPO, while DeepSeek-V3~\cite{deepseekv3} combines rule-based and model-based reward systems. Recently, DeepSeek-R1~\cite{deepseek-r1} achieved significant text reasoning improvements through pure RL~\cite{grpo}. Our work first adapts R1 methodologies to VLMs, enhancing visual reasoning, and systematically compares SFT-based and RL-based paradigms in visual reasoning tasks.

\section{Methodology}
\label{main-sec3}

In this section, we introduce \textit{\textbf{Reason-RFT}}, a novel two-stage training strategy to enhance the reasoning capabilities of VLMs in complex visual reasoning tasks. As shown in Fig.~\ref{fig:method}, the framework comprises two stages: (1) \textit{SFT-based Visual Reasoning Activation}, which uses SFT with high-quality CoT reasoning data to activate the model's domain-specific reasoning capabilities, and (2) \textit{RL-based Reasoning Enhancement}, which employs the GRPO algorithm with rule-based rewards to further push the upper limits of the model's reasoning potential.

\subsection{STAGE 1: SFT-based Reasoning Activation}
In the initial stage, we employ SFT on a structured visual reasoning dataset containing step-by-step reasoning processes. This stage trains the model to decompose complex tasks into logical steps. Each sample is represented as \( (x, q, r, a) \), where \( x \) denotes the input images, \( q \) is the question, \( r \) is the reasoning steps, and \( a \) is the final answer. The training objective maximizes the likelihood of generating both \( r \) and \( a \) given \( (x, q) \):
\begin{equation}
\mathcal{L}_{\text{SFT}} = -\mathbb{E}_{(x,q,r,a) \sim \mathcal{D}} \sum_{t=1}^{T} \log \pi_{\theta}(y_t \mid x, q, y_{<t}),
\end{equation}
where \( \mathcal{D} \) denotes the dataset, \( y \) represents the concatenated sequence of \( r \) and \( a \), and \( \pi_{\theta} \) denotes the model’s token distribution. The resulting model \( \pi_{\text{CoT}} \) is used to initialize the subsequent stage, providing a stable foundation for RL-based reasoning enhancement.

\subsection{STAGE 2: RL-based Reasoning Enhancement}

In the second stage, we refine \( \pi_{\text{CoT}} \) using GRPO, leveraging RL for its efficiency and scalability. Unlike Proximal Policy Optimization (PPO), which requires a computationally expensive value network, GRPO calculates relative advantages by comparing rewards within a group of sampled actions, reducing computational overhead and simplifying optimization. This makes GRPO particularly suitable for visual reasoning tasks.

\textbf{Sampling Action Groups} For each input state \( s = (x, q) \), 
GRPO samples a group of actions \( \{a_1, a_2, \dots, a_G\} \) from the current policy \( \pi_{\theta} \), initialized from \( \pi_{\text{CoT}} \). The sampling process is:
\begin{equation}
a_i \sim \pi_{\theta}(a \mid x, q), \quad \text{for } i = 1, 2, \dots, G.
\end{equation}
This strategy ensures diverse responses, promoting exploration and preventing premature convergence.

\textbf{Reward Evaluation.} Each sampled action \( a_i \) receives a reward \( R(a_i) \) based on verifiable criteria, yielding a reward set \( \{r_1, r_2, \dots, r_G\} \). For visual reasoning tasks, the reward \( R(a_i) \) is composed of a format reward \( R_{\text{format}}(a_i) \), which enforces structured outputs, and an accuracy reward \( R_{\text{acc}}(a_i) \), which measures correctness. This formulation balances structural alignment and factual precision in reasoning. The reward function is defined as:
\begin{equation}
R(a_i) = R_{\text{format}}(a_i) + R_{\text{acc}}(a_i).
\end{equation}

\textbf{Policy Update with Relative Advantage} Rewards are normalized within the sampled group to compute relative advantages \( \{A_1, A_2, \dots, A_G\} \), defined as:
\begin{equation}
A_i = \frac{r_i - \text{mean}\{r_1, r_2, \dots, r_G\}}{\text{std}\{r_1, r_2, \dots, r_G\}}.
\end{equation}
Based on these advantages, the policy is updated to reinforce actions with positive advantages and reduce the probability of less effective ones. To maintain training stability, the update is constrained by minimizing the KL divergence between the updated and reference policies.

\begin{figure*}[t]
\setlength{\abovecaptionskip}{-0.1em}
    \centering
    \vspace{-0.2cm}
    \includegraphics[width=0.99\linewidth]{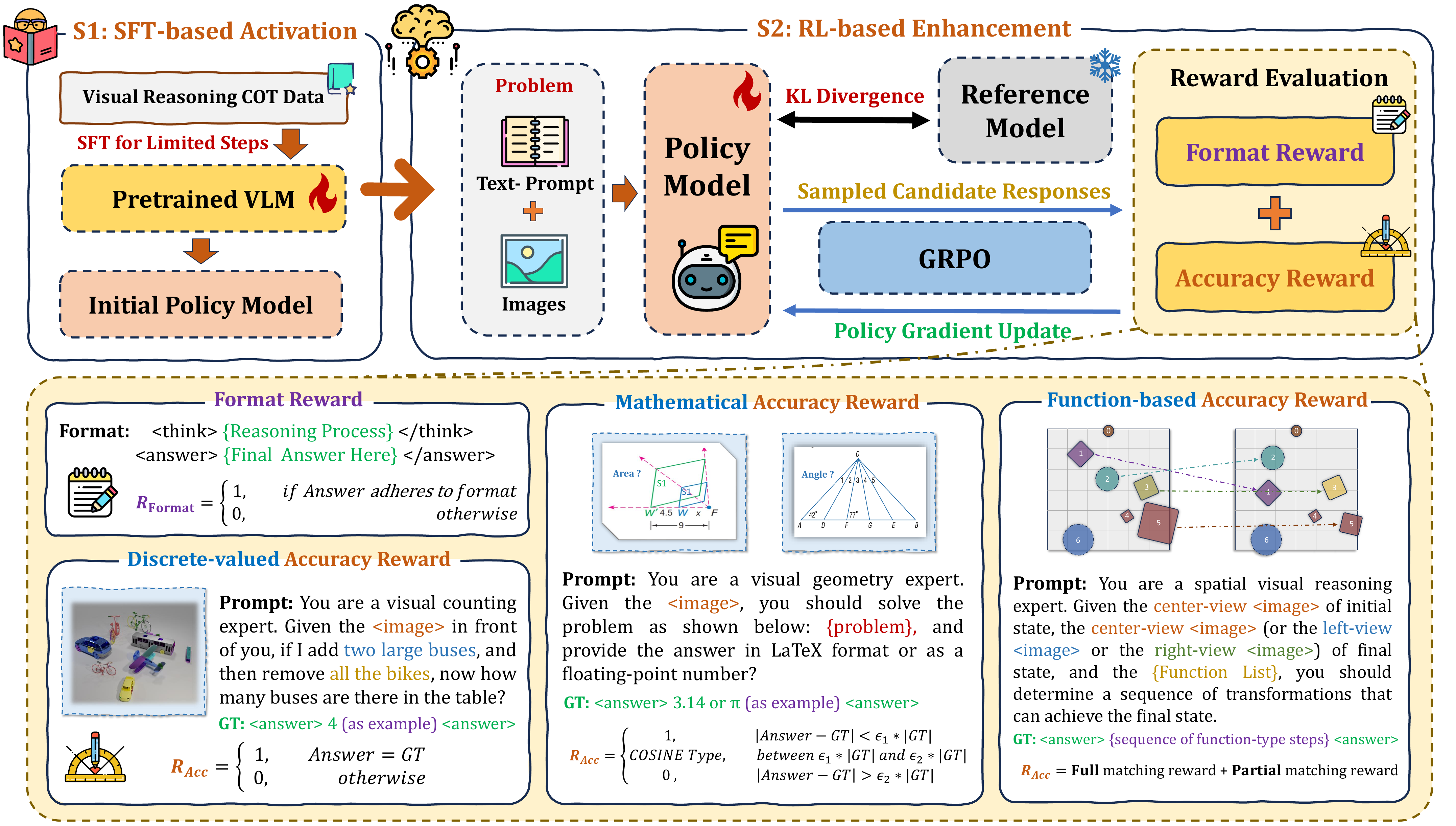}
    \vspace{0.1cm}
\caption{\textbf{Framework of Reason-RFT}. Reason-RFT adopts a two-stage training paradigm for visual reasoning. The first stage applies SFT with CoT reasoning to establish strong task-specific priors. In the second stage, GRPO is employed to further enhance reasoning capability and generalization.}
   \vspace{-0.3cm}
    \label{fig:method}
\end{figure*}

\subsection{Reward Design for Visual Reasoning Tasks}
For the diverse requirements of visual reasoning tasks, including visual counting, structure perception, and spatial transformation, our reward design integrates two essential components: \textit{format reward} and \textit{accuracy reward}. The format reward is uniformly applied across all tasks, ensuring that the model strictly adheres to a structured response format for consistency. For the accuracy reward, we carefully tailor the design to the specific characteristics of each task, as shown in Fig.~\ref{fig:method}, creating task-specific reward mechanisms to evaluate the correctness of the model's responses.

\textbf{Format Reward} This component ensures structured and interpretable responses by requiring the model to adhere to a predefined template: reasoning within \texttt{<think>} and \texttt{</think>} and the final answer within \texttt{<answer>} and \texttt{</answer>}. A reward of 1 is only given for strict adherence.

\textbf{Accuracy Reward} This component evaluates the correctness of the model's responses, ensuring alignment with ground truth across diverse visual reasoning tasks. To address task diversity, we design tailored reward mechanisms for discrete-valued, mathematical, and function-based problems. Each mechanism is crafted to handle the unique characteristics of its problem category, enabling precise and fair evaluation. Below, we introduce the three reward forms.

\begin{itemize} [left=1em]
\item \textbf{Discrete-valued Type} This reward type applies to visual counting and structure perception tasks, where answers are discrete values (\textit{e.g.}, multiple-choice or integer-based responses). The accuracy reward \( R_{\text{acc}}(a_i) \) is defined as:
\begin{small}
\begin{equation}
R_{\text{acc}}(a_i) = \begin{cases} 
1, & \text{if } a_{\text{pred}} = a_{\text{gt}} \\
0, & \text{otherwise},
\end{cases}
\end{equation}
\end{small}
where \( a_{\text{pred}} \) is the predicted answer and \( a_{\text{gt}} \) is the ground truth. This discrete reward penalizes deviations from ground truth, ensuring precision in tasks requiring unambiguous answers.

\item \textbf{Mathematical Type} This reward type is designed for structure perception tasks involving numerical answers, such as floating-point values or LaTeX-formatted expressions. It uses a tolerance-based evaluation to account for minor numerical deviations, which is defined as:
\begin{small}
\begin{equation}
R_{\text{acc}}(a_i) = \frac{1}{2}\left[\cos\left(\pi \times \frac{|a_{\text{pred}} - a_{\text{gt}}| - \epsilon_1 \times |a_{\text{gt}}|}{(\epsilon_2-\epsilon_1) \times |a_{\text{gt}}|}\right)+1\right],
\end{equation}
\end{small}
where \( a_{\text{pred}} \) is the predicted answer, \( a_{\text{gt}} \) is the ground truth, \( \epsilon_1 \) is the tolerance threshold for an exact match (\textit{e.g.}, 0.05), and \( \epsilon_2 \) is the upper bound for partial rewards (\textit{e.g.}, 0.20). If \( |a_{\text{pred}} - a_{\text{gt}}| < \epsilon_1 \times |a_{\text{gt}}| \), the reward is 1 (exact match); if \( |a_{\text{pred}} - a_{\text{gt}}| > \epsilon_2 \times |a_{\text{gt}}| \), the reward is 0 (incorrect). This formulation ensures smooth transitions between full and partial rewards, enabling fair evaluation of numerical accuracy.

\item \textbf{Function-based Type} This reward type is designed for spatial transformation tasks requiring a sequence of transformation functions. The accuracy reward \( R_{\text{acc}}(a_i) \) evaluates the alignment between the predicted sequence \( T_{\text{pred}} \) and the ground truth \( T_{\text{gt}} \), computed as:
\begin{small}
\begin{equation}
\label{eq_task3}
R_{\text{acc}}(a_i) = \frac{\text{len}(T_{\text{pred}}^{f+o+v}) + \alpha \cdot \text{len}(T_{\text{pred}}^{f+o/v}) + \beta \cdot \text{len}(T_{\text{pred}}^{f})}{\max(\text{len}(T_{\text{pred}}), \text{len}(T_{\text{gt}}))},
\end{equation}
\end{small}
where \( T_{\text{pred}}^{f+o+v} \) is the subset of transformation steps with complete matches (w/ function, object, and value), \( T_{\text{pred}}^{f+o/v} \) are the subsets with partial and only-function matches (w/ function and object, or w/ function and value), \( T_{\text{pred}}^{f} \) is the subset with only-function matches. \( \alpha \) and \( \beta \) are the weighting coefficients for partial matches. This formulation ensures nuanced evaluation for flexible adjustment of partial match contributions.
\end{itemize}

\section{Experiments}
\label{main-sec4}
We design experiments to investigate the following key research questions:

\begin{itemize}[left=1em]

\item \textbf{RQ1}: How effective is \textit{Reason-RFT} in reasoning, generalization, and data efficiency?

\item \textbf{RQ2}: Why is the STAGE 1 of SFT with CoT reasoning necessary?

\item \textbf{RQ3}: Why is the STAGE 2 of reinforcement fine-tuning necessary?

\item \textbf{RQ4}: How does reward design affect \textit{Reason-RFT}'s performance?

\item \textbf{RQ5}: What training dynamics emerge during reinforcement fine-tuning, and how do they shape the reasoning behavior of \textit{Reason-RFT}?

\end{itemize}
\subsection{Experimental Details}

\textbf{Datasets} 
In this paper, we comprehensively evaluate the visual reasoning capabilities of our method by leveraging six existing datasets, enhanced through subtask categorization, error-prone data filtering, and dataset restructuring. Detailed protocols for data filtering and restructuring are provided in Sec.~\ref{sec1}. Specifically, we define three task categories as follows.

\begin{itemize}[left=1em]
\item{\textbf{Visual Counting}} is a multimodal reasoning task evaluating the integration of linguistic, visual, and mathematical skills by solving arithmetic problems in 3D block-based scenes. Specifically, we filtered and corrected 35K samples from CLEVR-Math~\cite{clevr-math} for training and 1K test samples for in-domain (ID) evaluation. To assess generalization under domain-shift (DS), we constructed 1K new samples using 3D assets from Super-CLEVR~\cite{super-clevr}, including two subsets: direct arithmetic (DS-D) and mixed arithmetic (DS-M). Refer to the Appendix Sec.~\ref{subsec:vc} for details.

\item{\textbf{Structure Perception}} is a structural reasoning task requiring models to analyze relationships in geometries, imaging structures, chart layouts, and architectural designs. We filtered  4.5K training samples and 820 ID test samples from Geo170K~\cite{geo170k} and Math360K~\cite{math360k}, along with 800 samples from Geometry3K~\cite{lu2021inter} to evaluate DS adaptability. See the Appendix Sec.~\ref{subsec:sp}.

\item{\textbf{Spatial Transformation}} is a spatial-visual reasoning task requiring models to infer single- or multi-step transformations by analyzing initial and final states of 3D scenes from multiple perspectives (\textit{e.g.}, center, left, right). We generated 100K samples using TRANCE~\cite{hong2021transformation}, covering four difficulty levels, and selected 60K for training and 6K for testing through a specific filtering process. For DS evaluation, identical scenes are rendered from left/right viewpoints (DS-L/R) to test perspective-change robustness. Details can be found in the Appendix Sec.~\ref{subsec:st}.

\end{itemize}

\textbf{Evaluation Metrics} We use accuracy-rate (Acc) as the primary metric \cite{zhang2024lmms}. For numerical answers, correctness is verified by mathematical equivalence to the ground truth. For multiple-choice questions, we perform a string match. For function-type sequences, we use stepwise multi-level evaluation.

\textbf{Implementation Details} 
We utilize Qwen2-VL-2B and Qwen2-VL-7B \cite{qwen2vl} as the backbone models for our experiments. Our implementation is built on the open-source frameworks Open-R1~\cite{openr1} and vLLM~\cite{vllm}, ensuring reproducibility and scalability. All experiments were conducted on a cluster of servers, each equipped with 8$\times$A800 GPUs.
For further details, see the Appendix Sec.~\ref{sec2}.

\textbf{Training Paradigms and Baselines} 
To assess the performance and generalization of different training strategies, we compare: (1) SFT-based methods—ANS-SFT, which fine-tunes on answer generation, and CoT-SFT, which uses supervised learning with CoT reasoning; and (2) RL-based methods—Reason-RFT-Zero, which applies RL without reasoning activation stage, and Reason-RFT, which uses limited CoT data for reasoning activation before RL training. For comprehensive experiments, we use Qwen2-VL-Instruct~\cite{qwen2vl} as the base model (both 2B and 7B variants). In addition, we also select the most advanced open-source models ~\cite{Qwen2.5-VL,abdin2024phi,chen2024internvl,meta2024llama3vision,agrawal2024pixtral} and the proprietary models ~\cite{hurst2024gpt4o,team2024gemini} as baselines to evaluate the performance of different paradigms.

\subsection{Overall Evaluation of Reason-RFT Framework (RQ1)}
To evaluate {Reason-RFT}, we evaluate {Reason-RFT} using 2B- and 7B-parameter models on three visual reasoning tasks. The results are summarized as follows. 

\textbf{Strong reasoning performance across ID tasks.}
As shown in Tab.~\ref{tab:main_tab}, {Reason-RFT} achieves performance comparable to or better than both SFT- and RL-based methods across all tasks. In \textit{visual counting}, Reason-RFT-Zero achieves the best performance among all models in the 7B setting. In \textit{structure perception}, Reason-RFT outperforms most open-source and proprietary baselines in the 7B setting and remains competitive with top models such as InternVL-2.5-8B~\cite{chen2024internvl}. In \textit{spatial transformation}, Reason-RFT matches or exceeds SFT-based methods while consistently outperforming all baselines. These results demonstrate that Reason-RFT effectively integrates the strengths of both SFT and RL in structured reasoning tasks.

\textbf{Superior generalization under DS.} Under DS settings, {Reason-RFT} shows substantial gains over both traditional baselines and alternative training paradigms. In \textit{visual counting}, it outperforms ANS-SFT by 10.95\% (2B) and 13.93\% (7B). In \textit{structure perception}, Reason-RFT achieves the highest performance in the 2B model, with an 6.93\% gain over CoT-SFT, and remains highly competitive in the 7B model. Most notably, in \textit{the spatial transformation} task, the 2B Reason-RFT model surpasses GPT-4o~\cite{hurst2024gpt4o} by 31.7\%, showcasing remarkable generalization under DS.

\begin{wrapfigure}{r}{0.52\linewidth}
    \centering
    \vspace{-0.5em}
    \includegraphics[width=1.0\linewidth]{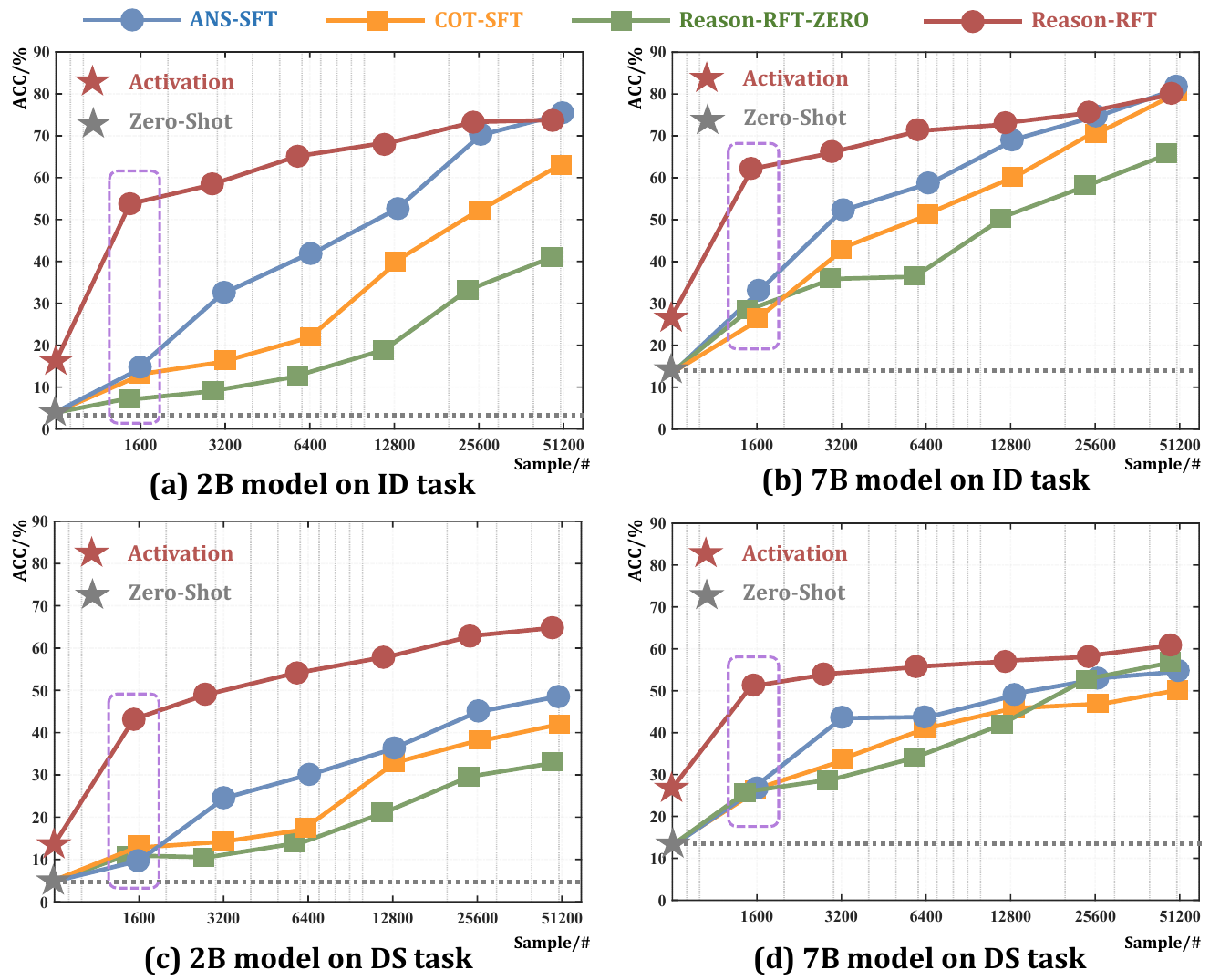}
    \vspace{-1.5em}
    \caption{Results of different methods and model sizes on \textit{spatial transformation} task across training.}
    \vspace{-1.0em}
    \label{fig:data_eff}
\end{wrapfigure}

\textbf{High training efficiency.} To evaluate data efficiency during training, we train all methods on the \textit{spatial transformation} task and monitor intermediate ID and DS performance (Fig.~\ref{fig:data_eff}). Additional results for \textit{visual counting} and \textit{structure perception} are provided in Appendix Sec.~\ref{sec3}. In the 2B model, Reason-RFT achieves 70\% of the final performance of Reason-RFT-Zero using only 3\% of the training data (1,600 samples), and reaches 82.5\% with 9\% of the data. In the 7B model, it achieves over 92\% of Reason-RFT-Zero’s performance using just 3\% of the data. These results confirm that Reason-RFT exhibits strong training efficiency in both ID and DS scenarios. The data-efficient nature of Reason-RFT renders it particularly effective for few-shot learning, offering significant potential for applications where labeled data is limited.

\begin{table*}[t]
    \scriptsize
    \centering
    \caption{\textbf{Results on three visual reasoning tasks.} The best results among different training paradigms are highlighted in \textbf{bold}, while the second-best results are \underline{underlined}. ``ID'' denotes in-domain test data, and ``DS'' denotes domain-shift test data.}
    \vspace{0.5em}
    \scalebox{0.96}{
    \begin{tabular}{l|cccc|ccc|cccc}
        \toprule
        \multicolumn{1}{l|}{\multirow{2}{*}{\textbf{Method}}} & \multicolumn{4}{c|}{\textbf{Visual Counting}} & \multicolumn{3}{c|}{\textbf{Structure Perception}} & \multicolumn{4}{c}{\textbf{Spatial Transformation}} \\ \cmidrule(lr){2-5} \cmidrule(lr){6-8} \cmidrule(lr){9-12}
        & \textbf{ID} & \textbf{DS-D} & \textbf{DS-M} & \textbf{AVG} & \textbf{ID} & \textbf{DS} & \textbf{AVG} & \textbf{ID} & \textbf{DS-L} & \textbf{DS-R} & \textbf{AVG} \\ \midrule
        \rowcolor[HTML]{F2F2F2} \multicolumn{12}{l}{\textbf{Proprietary Models}} \\ \midrule
        GPT-4o-2024-08-06 \cite{hurst2024gpt4o} & 68.10  & 42.54  & 9.60 & 40.08  & 50.18  & 43.49  & 46.83  & 42.55  & 28.67  & 29.76  & 35.88 \\
        Gemini-1.5-Pro \cite{team2024gemini} & 61.80  & 41.20 & 26.40 & 43.13  & 50.12  & 48.38  & 49.45  & 26.22  & 18.76  & 19.88  & 22.77\\
        \midrule
        \rowcolor[HTML]{F2F2F2} \multicolumn{12}{l}{\textbf{Open-Source Models}} \\ \midrule
        Qwen2.5-VL-3B-Inst. \cite{Qwen2.5-VL} & 75.90  & 50.93 & 4.40 & 43.74  & 36.75  & 37.44  & 37.09  & 8.57  & 8.26  &  8.31 & 8.42\\
        Phi-3.5-Vision-4B-Inst. \cite{abdin2024phi} & 21.40  & 18.27 & 6.00 & 15.22  & 36.83  & 50.25  & 43.54  & 7.42  & 2.45  &  4.02 & 5.33 \\
        Qwen2.5-VL-7B-Inst. \cite{Qwen2.5-VL} & 74.60  & 46.00  & 2.80 & 41.13 & 44.00  & 45.61  & 44.80  &  19.63 & 13.12  & 13.42  & 16.45\\
        InternVL-2.5-8B \cite{chen2024internvl}  & 93.50  & 46.13  & 2.80 & 47.48 & 63.00  &  47.32 & 51.60  & 7.19  &  6.62 & 6.63  & 6.91\\
        Llama-3.2-11B-Vision \cite{meta2024llama3vision} & 10.30  & 9.60  & 9.20  & 9.70 & 13.75  & 20.85  & 17.30  & 8.22  &  8.40 & 9.03  & 8.47\\
        Pixtral-12B \cite{agrawal2024pixtral} & 42.60  & 25.33  & 15.60 & 27.84  & 30.38  & 36.09  & 33.23  & 7.35  & 5.03  & 5.22  & 6.42\\
        \midrule
        \rowcolor[HTML]{F2F2F2} \multicolumn{12}{l}{\textbf{Qwen2VL-2B-Instruct}} \\ \midrule
        Zero-Shot & 82.40 & 42.67 & 0.00 & 41.69 & 25.86 & 20.63 & 23.25 & 3.78 & 4.60 & 4.67 & 4.35 \\
        $+$ ANS-SFT & 96.20 & 51.07 & 5.20 & 50.82 & \textbf{51.34} & 22.50 & 36.92 & \textbf{77.39} & \underline{49.24} & \underline{50.33} & \underline{58.99} \\
        $+$ CoT-SFT & 85.50 & 49.73 & \textbf{36.80} & \underline{57.34} & 43.05 & 25.25 & 34.15 & 64.37 & 43.19 & 42.86 & 50.14 \\
        $+$ Reason-RFT-Zero & \textbf{98.40} & \underline{58.00} & 5.20 & 53.87 & 47.68 & \underline{32.50} & \underline{40.09} & 42.13 & 34.07 & 33.41 & 33.74\\
        \rowcolor[HTML]{DAEFF9} {$+$ Reason-RFT} & \underline{96.80} & \textbf{60.00} & \underline{28.40} & \textbf{61.77} & \underline{49.03} & \textbf{33.13} & \textbf{41.08} & \underline{74.61} & \textbf{64.05} & \textbf{64.08} & \textbf{67.58} \\ \midrule
        \rowcolor[HTML]{F2F2F2} \multicolumn{12}{l}{\textbf{Qwen2VL-7B-Instruct}} \\ \midrule
        Zero-Shot & \underline{98.60} & 54.53 & 4.80 & 52.64 & 43.30 & 43.88 & 43.59 & 13.53 & 12.72 & 12.78 & 13.01 \\
        $+$ ANS-SFT & 95.00 & 42.53 & 8.00 & 48.51 & 51.34 & 25.38 & 38.36 & \textbf{82.19} & \underline{54.29} & \underline{54.83} & \underline{63.77} \\
        $+$ CoT-SFT & 87.30 & 45.33 & \underline{33.60} & 55.41 & 50.49 & 33.00 & 41.75 & \underline{81.31} & 47.90 & 47.80 & 59.00 \\
        $+$ Reason-RFT-Zero & \textbf{99.40} & \textbf{63.60} & 21.20 & \underline{61.40} & \underline{55.00} & \textbf{54.75} & \textbf{54.88} & 67.67 & 57.20 & 56.15 & 60.34\\
        \rowcolor[HTML]{DAEFF9} {$+$ Reason-RFT} & 95.60 & \underline{56.13} & \textbf{35.60} & \textbf{62.44} & \textbf{59.27} & \underline{49.25} & \underline{54.26} & 79.97 & \textbf{59.36} & \textbf{58.61} & \textbf{65.98}\\
        \bottomrule
    \end{tabular}
    }
    \label{tab:main_tab}
\vspace{-1.5em}
\end{table*}

\subsection{Effect of STAGE 1 on Initialization (RQ2)}
\label{id_task}
To investigate the role of CoT-SFT in initialization, we compare four baselines across three tasks. The results in Tab.~\ref{tab:main_tab} reveal the following: \textbf{\textit{(1) Consistent performance gains from CoT-SFT.}} Across all three tasks and both 2B and 7B model scales, Reason-RFT consistently outperforms Reason-RFT-Zero following stage 1 reasoning activation. This improvement is particularly notable when the model is small and the task involves complex output structures. For example, in the \textit{spatial transformation} task—which requires function-like serialized outputs—the 2B Reason-RFT model surpasses Reason-RFT-Zero by 33.84\%. \textit{\textbf{(2) Smaller models benefit more from CoT-SFT priors.}} In the \textit{visual counting} task under the DS-M setting, the 2B model with CoT-SFT outperforms Reason-RFT-Zero by 31.6\%. Although the gap narrows in the 7B model, CoT-SFT still yields substantial gains. This indicates that pure RL-based methods struggle to adapt from direct arithmetic to mixed arithmetic reasoning under DS, whereas CoT-SFT provides effective inductive priors for such adaptation. Moreover, under the same amount of CoT-SFT data, the 2B model still underperforms its 7B counterpart, highlighting the increased reliance of smaller models on CoT-SFT for acquiring reasoning capabilities.

\begin{wrapfigure}{r}{0.35\linewidth}
    \centering
    \vspace{-3.8em}
    \includegraphics[width=1.0\linewidth]{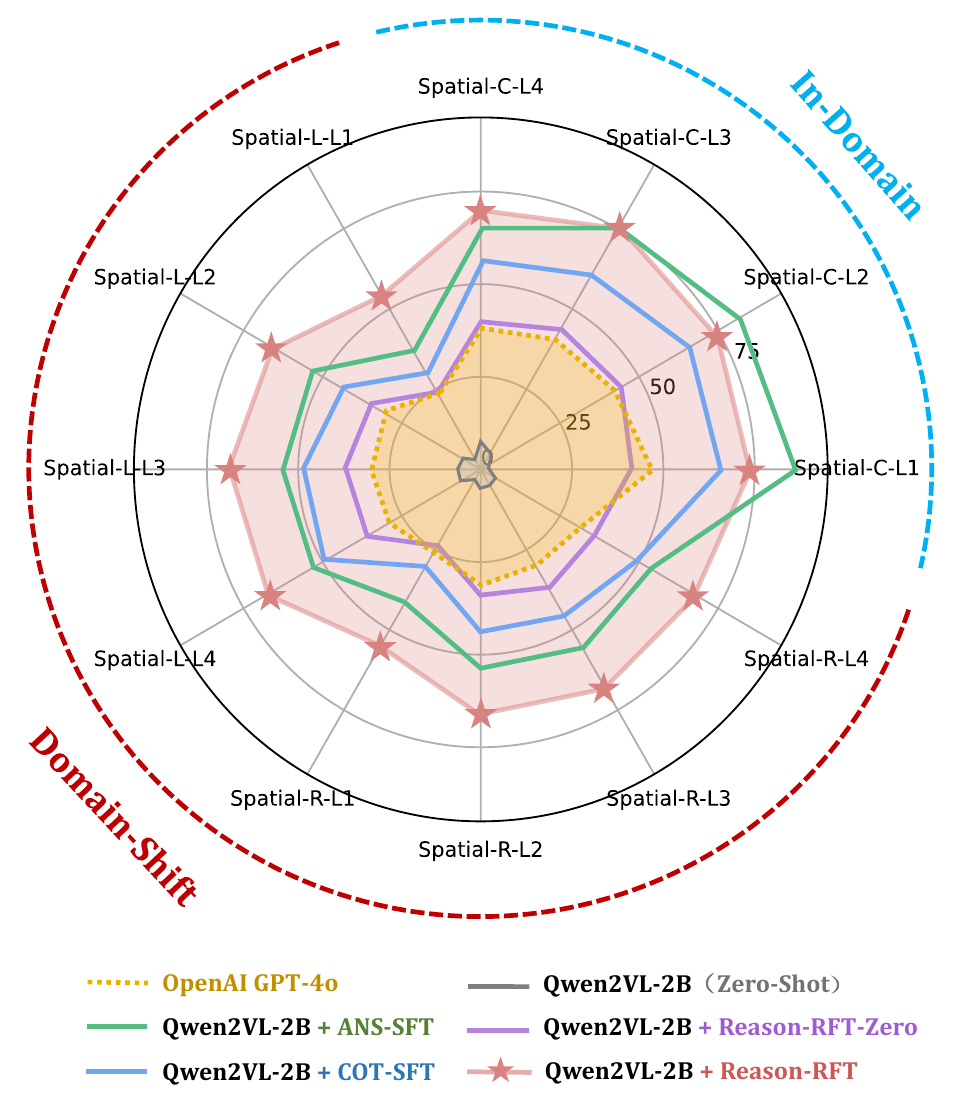}
    \caption{Results of DS \textit{v.s.} ID on \textit{spatial transformation} task.}
    \vspace{-0.8em}
    \label{fig:data_ood}
\end{wrapfigure}

\vspace{-1em}
\subsection{Effect of STAGE 2 on Generalization (RQ3)}
\label{ood_task}

To evaluate the impact of RL in stage 2, we compare the generalization performance of CoT-SFT and Reason-RFT across three visual reasoning tasks under DS. As shown in Tab.~\ref{tab:main_tab} and Fig.~\ref{fig:data_ood}, the results reveal the following:
\textbf{\textit{Reinforcement fine-tuning improves generalization beyond CoT-SFT.}}  
Across all domain-shift settings, Reason-RFT consistently outperforms CoT-SFT, demonstrating that reinforcement learning significantly enhances model robustness. For instance, in the \textit{visual counting} task, the 7B Reason-RFT achieves a combined DS-D and DS-M score 12.8\% higher than CoT-SFT. The improvement is even more pronounced in structure-sensitive tasks such as \textit{spatial transformation}, where the 2B Reason-RFT exceeds CoT-SFT by 21.04\% on average across DS-L and DS-R. These results indicate that CoT-SFT alone yields limited generalization, while reinforcement fine-tuning enables better adaptation to compositional and layout-dependent variations.

\vspace{-1em}
\subsection{Exploration on Reward Design (RQ4)}

\begin{wraptable}{r}{0.45\linewidth}
\centering
\vspace{-3.5em}
\caption{Results of different format reward strategies on the \textit{spatial transformation} task.}
\scriptsize
\begin{tabular}{c|ccll}
\toprule
\textbf{Setting}   & \textbf{ID}  & \textbf{DS-L} & \textbf{DS-R} & \textbf{AVG} \\ \midrule

\rowcolor[HTML]{F2F2F2} \multicolumn{5}{l}{\textbf{Qwen2VL-2B-Instruct}} \\ \midrule
\multicolumn{1}{l|}{Reason-RFT-Zero}                  &  \textbf{42.13}   & 34.07  &  33.41  &  33.74   \\ 
\rowcolor[HTML]{DAEFF9} \multicolumn{1}{r|}{$+$ visual tokens} &  42.01  &  \textbf{36.05} &  \textbf{35.97}  &  \textbf{38.01}  \\
\multicolumn{1}{l|}{Reason-RFT}                  &  \textbf{74.61}   & \textbf{64.05}  &  \textbf{64.08}  &  \textbf{69.33}   \\ 
\rowcolor[HTML]{DAEFF9} \multicolumn{1}{r|}{$+$ visual tokens} &  71.99  &  60.13 &  59.87  &  65.99 \\ \midrule

\rowcolor[HTML]{F2F2F2} \multicolumn{5}{l}{\textbf{Qwen2VL-7B-Instruct}} \\ \midrule
\multicolumn{1}{l|}{Reason-RFT-Zero}                  &  67.67   & 57.2  &  56.15  &  62.17   \\ 
\rowcolor[HTML]{DAEFF9} \multicolumn{1}{r|}{$+$ visual tokens} &  \textbf{70.28}  &  \textbf{59.52} &  \textbf{57.01}  &  \textbf{64.27}  \\
\multicolumn{1}{l|}{Reason-RFT}                  &  \textbf{79.97}   & \textbf{59.36}  &  \textbf{58.61}  &  \textbf{69.48}   \\ 
\rowcolor[HTML]{DAEFF9} \multicolumn{1}{r|}{$+$ visual tokens} &  79.85  &  58.71 &  57.98  &  69.09 \\ \bottomrule

\end{tabular}
\vspace{-0.5em}
\label{format_reward}
\end{wraptable}

\textbf{Format Reward.}  
In DeepSeek-R1~\cite{guo2025deepseek}, the format reward enforces the use of \texttt{<think>} and \texttt{<answer>} tokens to structure reasoning in textual tasks. To better support visual reasoning, we extend this with \texttt{<summary>} and \texttt{<caption>} tokens to incorporate visual observations via caption-style prompts. As shown in Tab.~\ref{format_reward}, this improves Reason-RFT-Zero but has limited effect on Reason-RFT. We attribute this to Reason-RFT's prior CoT supervision, which likely helps it internalize caption-like structures in stage 1, reducing the benefit of explicit tags. In contrast, Reason-RFT-Zero benefits more from such structural cues, indicating greater sensitivity to format-level guidance.

\begin{wraptable}{r}{0.45\linewidth}
\centering
\vspace{-1.8em}
\caption{Results of different accuracy reward strategies on the \textit{spatial transformation} task.}
\scalebox{0.7}{
\begin{tabular}{ccc|cccc}
\toprule
\textbf{Setting} & \textbf{$\alpha$} & \textbf{$\beta$}  & \textbf{ID}     & \textbf{DS-L} & \textbf{DS-R} & \textbf{AVG} \\ \midrule
\rowcolor[HTML]{F2F2F2} \multicolumn{7}{l}{\textbf{Qwen2VL-2B-Instruct}} \\ \midrule
Baseline                          & 0      & 0       &   74.61    &   64.05    &  64.08  &  \textbf{69.33} \\ 
(a)                               & 0.50   & 0.25   &   \textbf{79.18}      &  56.36    &  55.45  &  67.54   \\
(b)                               & -0.25  & -0.50  &   73.69    &   \textbf{64.41}   &  \textbf{64.72}  &  69.13 \\ \midrule
\rowcolor[HTML]{F2F2F2} \multicolumn{7}{l}{\textbf{Qwen2VL-7B-Instruct}} \\ \midrule
Baseline                          & 0      & 0       &   79.97    &   59.36    &  58.61  &  69.48 \\ 
(a)                               & 0.50   & 0.25   &   \textbf{80.89}      &  53.20    &  52.61  &  66.90   \\
(b)                               & -0.25  & -0.50  &   75.03    &   \textbf{64.83}   &  \textbf{63.18}  &  \textbf{69.52} \\\bottomrule
\end{tabular}}
\label{acc_reward}
\vspace{-1em}
\end{wraptable}

\textbf{Accuracy Reward.}  
We explore accuracy reward design in the \textit{spatial transformation} task, which requires predicting transformation sequences in a structured format. The formulation in Eq.~\ref{eq_task3} introduces coefficients \( \alpha \) and \( \beta \) to control tolerance for partial matches. We test three settings: (1) \( \alpha = 0, \beta = 0 \) (exact match only), (2) \( \alpha = 0.50, \beta = 0.25 \) (partial credit), and (3) \( \alpha = -0.25, \beta = -0.50 \) (penalized partial matches). Results on 2B and 7B models (Tab.~\ref{acc_reward}) show that: (1) partial credit improves ID performance but harms generalization, suggesting ``soft rewards'' reduce robustness; (2) penalizing partial matches improves generalization under domain shift, indicating ``hard rewards'' better support serialized reasoning.

\begin{figure}[!t]
\setlength{\abovecaptionskip}{-0.1em}
    \centering
    \includegraphics[width=0.97\linewidth]{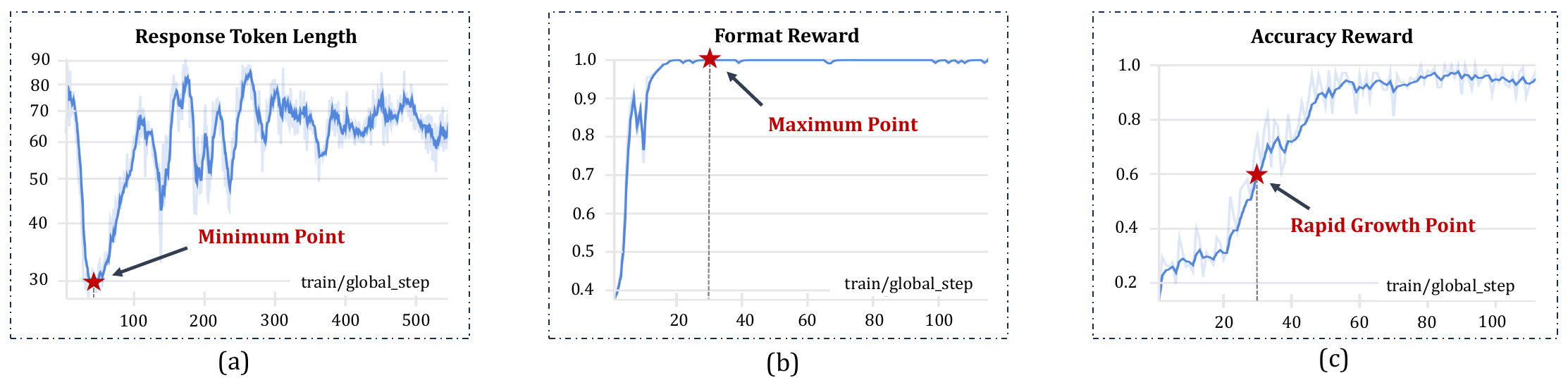}
    \caption{\textbf{Analysis of Greedy Reward Stratification.} The model’s reasoning token length first decreases, then gradually rises and stabilizes during Reason-RFT-Zero training. The peak of the format reward coincides with the accuracy reward’s rapid growth phase.}
    \label{fig:abl_grs}
\end{figure}

\begin{figure}[tbp]
    \centering
    \vspace{-1em}
    \begin{minipage}[t]{0.46\textwidth}
        \centering
        \includegraphics[width=1.0\linewidth]{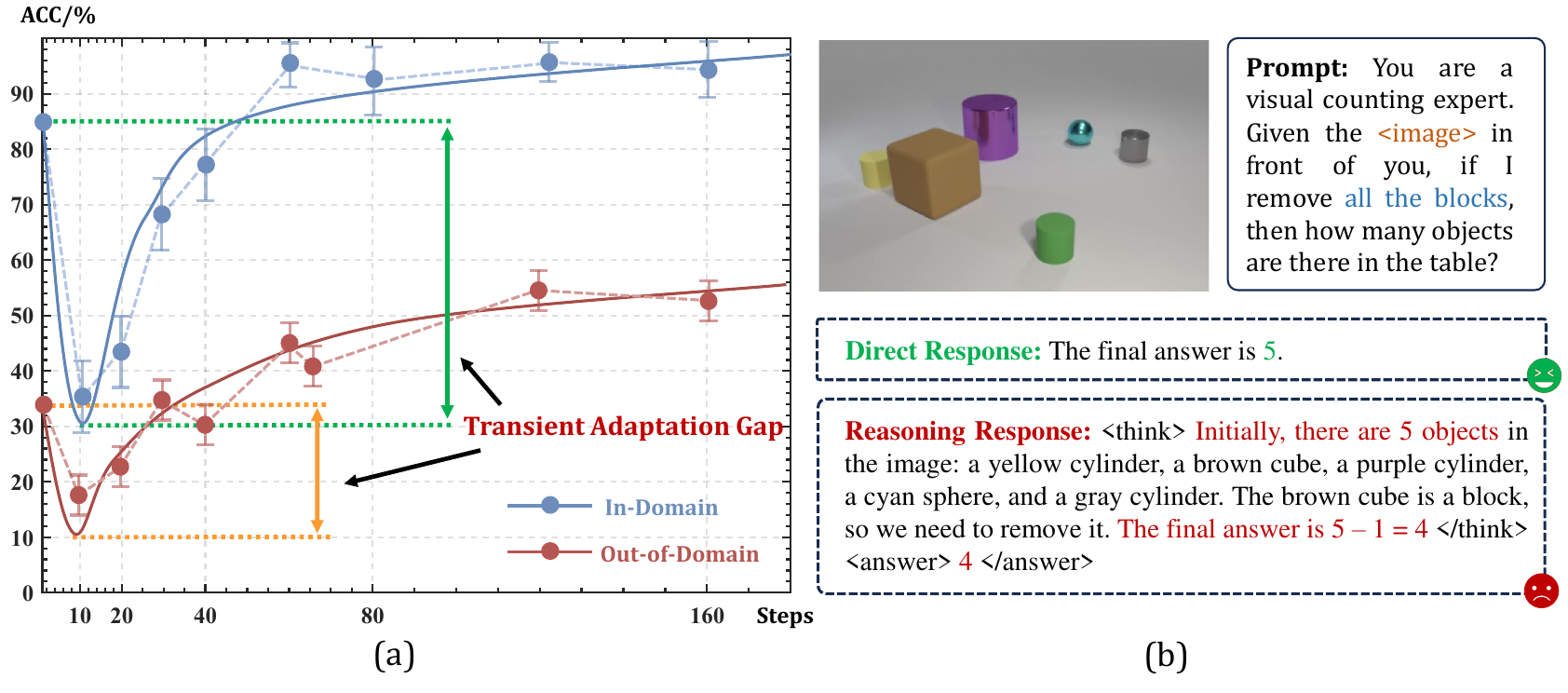}
        \vspace{-1.5em}
        \caption{\textbf{Illustration of the Transient Adaptation Gap.} (a) shows a sharp drop and recovery in both ID and DS test performances within the early training steps. (b) shows a case study of the prediction result on early step.}
        \label{fig:abl_gap}
    \end{minipage}
    \hfill
    \begin{minipage}[t]{0.50\textwidth}
        \centering
        \includegraphics[width=1.0\linewidth]{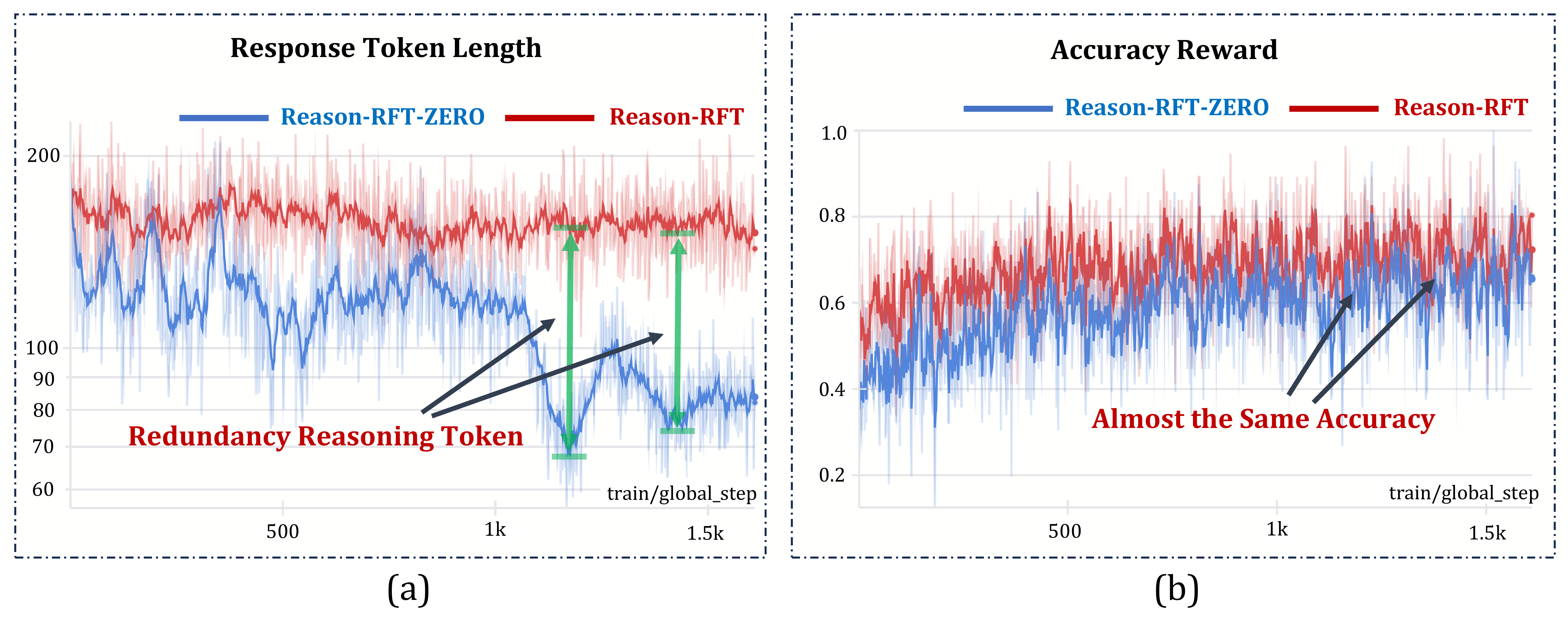}
        \vspace{-1.5em}
        \caption{\textbf{Analysis of Reasoning Redundancy.} (a) shows the reasoning token length curves for Reason-RFT-Zero and Reason-RFT during training. (b) displays their accuracy reward, with both paradigms converging to similar accuracy rate.}
        \label{fig:abl_redun}
    \end{minipage}
    \vspace{-1em}
\end{figure}

\subsection{Training Dynamics and Reasoning Behavior Analysis (RQ5)}

\textbf{Greedy Reward Stratification.}
This phenomenon captures the model’s tendency, particularly in Reason-RFT-Zero, to prioritize easier-to-optimize rewards (e.g., format reward) before addressing more challenging objectives (e.g., accuracy reward). As shown in Fig.~\ref{fig:abl_grs}, the reasoning token length initially drops, then gradually increases and stabilizes. This dynamic correlates with the format reward quickly reaching a plateau, followed by a sharp rise in the accuracy reward. We infer that the model initially simplifies its outputs to adapt rapidly to structured format expectations, and only later allocates learning capacity to improve reasoning correctness.

\textbf{Transient Adaptation Gap.}
This refers to a temporary performance degradation observed in the early training phase of Reason-RFT-Zero. When the model shifts from directly predicting answers to producing structured reasoning traces, it experiences a brief adaptation bottleneck—marked by a sharp decline and gradual recovery in accuracy. Fig.~\ref{fig:abl_gap} (a) illustrates this drop within the first 100 steps on the \textit{visual counting} task under both ID and DS settings. A case analysis in Fig.~\ref{fig:abl_gap} (b) further reveals that forcing structured reasoning prematurely may lead to incorrect outputs, highlighting the sensitivity of early-stage training to reasoning format constraints.

\textbf{Reasoning Redundancy.}
This phenomenon concerns the discrepancy in reasoning token length between models trained with and without CoT activation. In the \textit{structure perception} task, Reason-RFT and Reason-RFT-Zero attain similar accuracy, yet the former generates significantly longer reasoning traces (Fig.~\ref{fig:abl_redun}). This likely stems from Reason-RFT’s use of CoT data distilled from advanced models (e.g., GPT-4o), which encourages verbose reasoning during stage 1. In the absence of penalties or length control in reinforcement fine-tuning, such verbosity persists. By contrast, Reason-RFT-Zero converges to more concise reasoning through reward-driven exploration. We hypothesize that these longer chains in Reason-RFT may introduce unnecessary computational overhead or reflect overthinking relative to task complexity.

\section{Conclusion}
\label{mian-sec5}
In this paper, we propose \textit{\textbf{Reason-RFT}}, a novel reinforcement fine-tuning framework that enhances the generalization capabilities of visual reasoning models. By integrating SFT with CoT reasoning activation data and GRPO-based reinforcement learning, \textit{{Reason-RFT}} effectively mitigates key challenges such as overfitting and cognitive rigidity, thereby improving cross-domain transferability and real-world applicability.
To support systematic evaluation, we reconstruct a comprehensive dataset covering visual counting, structure perception, and spatial transformation tasks, establishing a robust benchmark for assessing model performance across diverse scenarios. Extensive experiments demonstrate the effectiveness of \textit{{Reason-RFT}}, providing valuable insights for advancing visual reasoning research and introducing a new paradigm in multimodal learning.

\section*{Acknowledgments}
This work was supported by the National Natural Science Foundation of China (62476011), and the National Science and Technology Major Project (No. 2022ZD0117800).


\bibliographystyle{unsrt}
\bibliography{main}  

\begin{thebibliography}{100}

\bibitem{clevr-math}
Adam~Dahlgren Lindstr{\"o}m and Savitha~Sam Abraham.
\newblock Clevr-math: A dataset for compositional language, visual and mathematical reasoning.
\newblock {\em arXiv preprint arXiv:2208.05358}, 2022.

\bibitem{OpenAI2024o1}
OpenAI.
\newblock Learning to reason with llms.
\newblock \url{https://openai.com/index/learning-to-reason-with-llms/}, 2024.
\newblock Accessed: 2025-03-02.

\bibitem{image_processing}
Maria~MP Petrou and Costas Petrou.
\newblock {\em Image processing: the fundamentals}.
\newblock John Wiley \& Sons, 2010.

\bibitem{tan2024joint}
Huajie Tan, Guoqing Xiang, Xiaodong Xie, and Huizhu Jia.
\newblock Joint frame-level and block-level rate-perception optimized preprocessing for video coding.
\newblock In {\em Proceedings of the 6th ACM International Conference on Multimedia in Asia}, pages 1--1, 2024.

\bibitem{zhang2025beyond}
Qizhe Zhang, Mengzhen Liu, Lichen Li, Ming Lu, Yuan Zhang, Junwen Pan, Qi~She, and Shanghang Zhang.
\newblock Beyond attention or similarity: Maximizing conditional diversity for token pruning in mllms.
\newblock {\em arXiv preprint arXiv:2506.10967}, 2025.

\bibitem{ji2024advlora}
Yuheng Ji, Yue Liu, Zhicheng Zhang, Zhao Zhang, Yuting Zhao, Gang Zhou, Xingwei Zhang, Xinwang Liu, and Xiaolong Zheng.
\newblock Advlora: Adversarial low-rank adaptation of vision-language models.
\newblock {\em arXiv preprint arXiv:2404.13425}, 2024.

\bibitem{ji2025enhancing}
Yuheng Ji, Yue Liu, Zhicheng Zhang, Zhao Zhang, Yuting Zhao, Xiaoshuai Hao, Gang Zhou, Xingwei Zhang, and Xiaolong Zheng.
\newblock Enhancing adversarial robustness of vision-language models through low-rank adaptation.
\newblock In {\em Proceedings of the 2025 International Conference on Multimedia Retrieval}, pages 550--559, 2025.

\bibitem{ji2023learning}
Yuheng Ji, Xingwei Zhang, Gang Zhou, Xiaolong Zheng, and Daniel~Dajun Zeng.
\newblock Learning hash subspace from large-scale multi-modal pre-training: A clip-based cross-modal hashing framework.
\newblock In {\em China Conference on Command and Control}, pages 514--526. Springer, 2023.

\bibitem{lyu2025egoprompt}
Huaihai Lyu, Chaofan Chen, Yuheng Ji, and Changsheng Xu.
\newblock Egoprompt: Prompt pool learning for egocentric action recognition.
\newblock {\em arXiv preprint arXiv:2508.03266}, 2025.

\bibitem{mu2023configurable}
Jianan Mu, Huajie Tan, Shuai Chen, Min Cai, Jing Ye, Huawei Li, and Xiaowei Li.
\newblock Configurable and high-level pipelined lattice-based post quantum cryptography hardware accelerator design.
\newblock In {\em 2023 IEEE 32nd Asian Test Symposium (ATS)}, pages 1--6. IEEE, 2023.

\bibitem{mu2023energy}
Jianan Mu, Huajie Tan, Jiawen Wu, Haotian Lu, Chip-Hong Chang, Shuai Chen, Shengwen Liang, Jing Ye, Huawei Li, and Xiaowei Li.
\newblock Energy-efficient ntt design with one-bank sram and 2-d pe array.
\newblock In {\em 2023 Design, Automation \& Test in Europe Conference \& Exhibition (DATE)}, pages 1--2. IEEE, 2023.

\bibitem{bai2025alleviating}
Songran Bai, Yuheng Ji, Yue Liu, Xingwei Zhang, Xiaolong Zheng, and Daniel~Dajun Zeng.
\newblock Alleviating performance disparity in adversarial spatiotemporal graph learning under zero-inflated distribution.
\newblock In {\em Proceedings of the AAAI Conference on Artificial Intelligence}, volume~39, pages 11436--11444, 2025.

\bibitem{cordts2016cityscapes}
Marius Cordts, Mohamed Omran, Sebastian Ramos, Timo Rehfeld, Markus Enzweiler, Rodrigo Benenson, Uwe Franke, Stefan Roth, and Bernt Schiele.
\newblock The cityscapes dataset for semantic urban scene understanding.
\newblock In {\em IEEE conference on computer vision and pattern recognition}, pages 3213--3223, 2016.

\bibitem{vsl_bench}
Jihan Yang, Shusheng Yang, Anjali~W Gupta, Rilyn Han, Li~Fei-Fei, and Saining Xie.
\newblock Thinking in space: How multimodal large language models see, remember, and recall spaces.
\newblock {\em arXiv preprint arXiv:2412.14171}, 2024.

\bibitem{zhan2020medical}
Li-Ming Zhan, Bo~Liu, Lu~Fan, Jiaxin Chen, and Xiao-Ming Wu.
\newblock Medical visual question answering via conditional reasoning.
\newblock In {\em ACM International Conference on Multimedia}, pages 2345--2354, 2020.

\bibitem{MedVLM}
Jiazhen Pan, Che Liu, Junde Wu, Fenglin Liu, Jiayuan Zhu, Hongwei~Bran Li, Chen Chen, Cheng Ouyang, and Daniel Rueckert.
\newblock Medvlm-r1: Incentivizing medical reasoning capability of vision-language models (vlms) via reinforcement learning.
\newblock {\em arXiv preprint arXiv:2502.19634}, 2025.

\bibitem{robobrain}
Yuheng Ji, Huajie Tan, Jiayu Shi, Xiaoshuai Hao, Yuan Zhang, Hengyuan Zhang, Pengwei Wang, Mengdi Zhao, Yao Mu, Pengju An, et~al.
\newblock Robobrain: A unified brain model for robotic manipulation from abstract to concrete.
\newblock In {\em Proceedings of the Computer Vision and Pattern Recognition Conference}, pages 1724--1734, 2025.

\bibitem{team2025robobrain}
BAAI~RoboBrain Team, Mingyu Cao, Huajie Tan, Yuheng Ji, Minglan Lin, Zhiyu Li, Zhou Cao, Pengwei Wang, Enshen Zhou, Yi~Han, et~al.
\newblock Robobrain 2.0 technical report.
\newblock {\em arXiv preprint arXiv:2507.02029}, 2025.

\bibitem{liu2024robomamba}
Jiaming Liu, Mengzhen Liu, Zhenyu Wang, Lily Lee, Kaichen Zhou, Pengju An, Senqiao Yang, Renrui Zhang, Yandong Guo, and Shanghang Zhang.
\newblock Robomamba: Multimodal state space model for efficient robot reasoning and manipulation.
\newblock {\em arXiv e-prints}, pages arXiv--2406, 2024.

\bibitem{tan2025roboos}
Huajie Tan, Xiaoshuai Hao, Cheng Chi, Minglan Lin, Yaoxu Lyu, Mingyu Cao, Dong Liang, Zhuo Chen, Mengsi Lyu, Cheng Peng, et~al.
\newblock Roboos: A hierarchical embodied framework for cross-embodiment and multi-agent collaboration.
\newblock {\em arXiv preprint arXiv:2505.03673}, 2025.

\bibitem{zhou2025code}
Enshen Zhou, Qi~Su, Cheng Chi, Zhizheng Zhang, Zhongyuan Wang, Tiejun Huang, Lu~Sheng, and He~Wang.
\newblock Code-as-monitor: Constraint-aware visual programming for reactive and proactive robotic failure detection.
\newblock In {\em Proceedings of the Computer Vision and Pattern Recognition Conference}, pages 6919--6929, 2025.

\bibitem{zhou2025roborefer}
Enshen Zhou, Jingkun An, Cheng Chi, Yi~Han, Shanyu Rong, Chi Zhang, Pengwei Wang, Zhongyuan Wang, Tiejun Huang, Lu~Sheng, et~al.
\newblock Roborefer: Towards spatial referring with reasoning in vision-language models for robotics.
\newblock {\em arXiv preprint arXiv:2506.04308}, 2025.

\bibitem{li2024foundation}
Dingzhe Li, Yixiang Jin, Yuhao Sun, Hongze Yu, Jun Shi, Xiaoshuai Hao, Peng Hao, Huaping Liu, Fuchun Sun, Jianwei Zhang, et~al.
\newblock What foundation models can bring for robot learning in manipulation: A survey.
\newblock {\em arXiv preprint arXiv:2404.18201}, 2024.

\bibitem{song2025maniplvm}
Zirui Song, Guangxian Ouyang, Mingzhe Li, Yuheng Ji, Chenxi Wang, Zixiang Xu, Zeyu Zhang, Xiaoqing Zhang, Qian Jiang, Zhenhao Chen, et~al.
\newblock Maniplvm-r1: Reinforcement learning for reasoning in embodied manipulation with large vision-language models.
\newblock {\em arXiv preprint arXiv:2505.16517}, 2025.

\bibitem{HAO2025103018}
Xiaoshuai Hao, Yunfeng Diao, Mengchuan Wei, Yifan Yang, Peng Hao, Rong Yin, Hui Zhang, Weiming Li, Shu Zhao, and Yu~Liu.
\newblock Mapfusion: A novel bev feature fusion network for multi-modal map construction.
\newblock {\em Information Fusion}, page 103018, 2025.

\bibitem{hao2024your}
Xiaoshuai Hao, Mengchuan Wei, Yifan Yang, Haimei Zhao, Hui Zhang, Yi~Zhou, Qiang Wang, Weiming Li, Lingdong Kong, and Jing Zhang.
\newblock Is your {HD} map constructor reliable under sensor corruptions?
\newblock In {\em Advances in Neural Information Processing Systems}, 2024.

\bibitem{hao2024mapdistill}
Xiaoshuai Hao, Ruikai Li, Hui Zhang, Dingzhe Li, Rong Yin, Sangil Jung, Seung-In Park, ByungIn Yoo, Haimei Zhao, and Jing Zhang.
\newblock Mapdistill: Boosting efficient camera-based hd map construction via camera-lidar fusion model distillation.
\newblock In {\em European Conference on Computer Vision}, pages 166--183, 2024.

\bibitem{hao2024mbfusion}
Xiaoshuai Hao, Hui Zhang, Yifan Yang, Yi~Zhou, Sangil Jung, Seung-In Park, and ByungIn Yoo.
\newblock Mbfusion: A new multi-modal bev feature fusion method for hd map construction.
\newblock In {\em IEEE International Conference on Robotics and Automation}, pages 15922--15928, 2024.

\bibitem{hao2025msc}
Xiaoshuai Hao, Guanqun Liu, Yuting Zhao, Yuheng Ji, Mengchuan Wei, Haimei Zhao, Lingdong Kong, Rong Yin, and Yu~Liu.
\newblock Msc-bench: Benchmarking and analyzing multi-sensor corruption for driving perception.
\newblock {\em arXiv preprint arXiv:2501.01037}, 2025.

\bibitem{zhao2025fastrsr}
Yuting Zhao, Yuheng Ji, Xiaoshuai Hao, and Shuxiao Li.
\newblock Fastrsr: Efficient and accurate road surface reconstruction from bird's eye view.
\newblock {\em arXiv preprint arXiv:2504.09535}, 2025.

\bibitem{hao2025really}
Xiaoshuai Hao, Yuting Zhao, Yuheng Ji, Luanyuan Dai, Peng Hao, Dingzhe Li, Shuai Cheng, and Rong Yin.
\newblock What really matters for robust multi-sensor hd map construction?
\newblock {\em arXiv preprint arXiv:2507.01484}, 2025.

\bibitem{symbolic_computing}
Artur~d'Avila Garcez, Marco Gori, Luis~C Lamb, Luciano Serafini, Michael Spranger, and Son~N Tran.
\newblock Neural-symbolic computing: An effective methodology for principled integration of machine learning and reasoning.
\newblock {\em arXiv preprint arXiv:1905.06088}, 2019.

\bibitem{symbolic}
Saeed Amizadeh, Hamid Palangi, Alex Polozov, Yichen Huang, and Kazuhito Koishida.
\newblock Neuro-symbolic visual reasoning: Disentangling.
\newblock In {\em International Conference on Machine Learning}, pages 279--290. Pmlr, 2020.

\bibitem{neuro-symbolic}
Minkyu Choi, Harsh Goel, Mohammad Omama, Yunhao Yang, Sahil Shah, and Sandeep Chinchali.
\newblock Towards neuro-symbolic video understanding.
\newblock In {\em European Conference on Computer Vision}, pages 220--236. Springer, 2024.

\bibitem{zhang2024take}
Mingyu Zhang, Jiting Cai, Mingyu Liu, Yue Xu, Cewu Lu, and Yong-Lu Li.
\newblock Take a step back: Rethinking the two stages in visual reasoning.
\newblock In {\em European Conference on Computer Vision}, pages 124--141. Springer, 2024.

\bibitem{visual_programming}
Tanmay Gupta and Aniruddha Kembhavi.
\newblock Visual programming: Compositional visual reasoning without training.
\newblock In {\em IEEE/CVF Conference on Computer Vision and Pattern Recognition}, pages 14953--14962, 2023.

\bibitem{llava-cot}
Guowei Xu, Peng Jin, Li~Hao, Yibing Song, Lichao Sun, and Li~Yuan.
\newblock Llava-o1: Let vision language models reason step-by-step.
\newblock {\em arXiv preprint arXiv:2411.10440}, 2024.

\bibitem{llamav-o1}
Omkar Thawakar, Dinura Dissanayake, Ketan More, Ritesh Thawkar, Ahmed Heakl, Noor Ahsan, Yuhao Li, Mohammed Zumri, Jean Lahoud, Rao~Muhammad Anwer, et~al.
\newblock Llamav-o1: Rethinking step-by-step visual reasoning in llms.
\newblock {\em arXiv preprint arXiv:2501.06186}, 2025.

\bibitem{guo2025deepseek}
Daya Guo, Dejian Yang, Haowei Zhang, Junxiao Song, Ruoyu Zhang, Runxin Xu, Qihao Zhu, Shirong Ma, Peiyi Wang, Xiao Bi, et~al.
\newblock Deepseek-r1: Incentivizing reasoning capability in llms via reinforcement learning.
\newblock {\em arXiv preprint arXiv:2501.12948}, 2025.

\bibitem{team2025kimi}
Kimi Team, Angang Du, Bofei Gao, Bowei Xing, Changjiu Jiang, Cheng Chen, Cheng Li, Chenjun Xiao, Chenzhuang Du, Chonghua Liao, et~al.
\newblock Kimi k1. 5: Scaling reinforcement learning with llms.
\newblock {\em arXiv preprint arXiv:2501.12599}, 2025.

\bibitem{mathvista}
Pan Lu, Hritik Bansal, Tony Xia, Jiacheng Liu, Chunyuan Li, Hannaneh Hajishirzi, Hao Cheng, Kai-Wei Chang, Michel Galley, and Jianfeng Gao.
\newblock Mathvista: Evaluating mathematical reasoning of foundation models in visual contexts.
\newblock {\em arXiv preprint arXiv:2310.02255}, 2023.

\bibitem{hao2023mixgen}
Xiaoshuai Hao, Yi~Zhu, Srikar Appalaraju, Aston Zhang, Wanqian Zhang, Bo~Li, and Mu~Li.
\newblock Mixgen: A new multi-modal data augmentation.
\newblock In {\em IEEE/CVF winter conference on applications of computer vision}, pages 379--389, 2023.

\bibitem{ma2025followyourmotion}
Yue Ma, Yulong Liu, Qiyuan Zhu, Ayden Yang, Kunyu Feng, Xinhua Zhang, Zhifeng Li, Sirui Han, Chenyang Qi, and Qifeng Chen.
\newblock Follow-your-motion: Video motion transfer via efficient spatial-temporal decoupled finetuning.
\newblock {\em arXiv preprint arXiv:2506.05207}, 2025.

\bibitem{ma2024followyouremoji}
Yue Ma, Hongyu Liu, Hongfa Wang, Heng Pan, Yingqing He, Junkun Yuan, Ailing Zeng, Chengfei Cai, Heung-Yeung Shum, Wei Liu, et~al.
\newblock Follow-your-emoji: Fine-controllable and expressive freestyle portrait animation.
\newblock In {\em SIGGRAPH Asia 2024 Conference Papers}, pages 1--12, 2024.

\bibitem{ma2023magicstick}
Yue Ma, Xiaodong Cun, Yingqing He, Chenyang Qi, Xintao Wang, Ying Shan, Xiu Li, and Qifeng Chen.
\newblock Magicstick: Controllable video editing via control handle transformations.
\newblock {\em arXiv preprint arXiv:2312.03047}, 2023.

\bibitem{ma2024followpose}
Yue Ma, Yingqing He, Xiaodong Cun, Xintao Wang, Siran Chen, Xiu Li, and Qifeng Chen.
\newblock Follow your pose: Pose-guided text-to-video generation using pose-free videos.
\newblock In {\em Proceedings of the AAAI Conference on Artificial Intelligence}, volume~38, pages 4117--4125, 2024.

\bibitem{ma2022visual}
Yue Ma, Yali Wang, Yue Wu, Ziyu Lyu, Siran Chen, Xiu Li, and Yu~Qiao.
\newblock Visual knowledge graph for human action reasoning in videos.
\newblock In {\em Proceedings of the 30th ACM International Conference on Multimedia}, pages 4132--4141, 2022.

\bibitem{yan2025eedit}
Zexuan Yan, Yue Ma, Chang Zou, Wenteng Chen, Qifeng Chen, and Linfeng Zhang.
\newblock Eedit: Rethinking the spatial and temporal redundancy for efficient image editing.
\newblock {\em arXiv preprint arXiv:2503.10270}, 2025.

\bibitem{feng2025follow}
Kunyu Feng, Yue Ma, Xinhua Zhang, Boshi Liu, Yikuang Yuluo, Yinhan Zhang, Runtao Liu, Hongyu Liu, Zhiyuan Qin, Shanhui Mo, et~al.
\newblock Follow-your-instruction: A comprehensive mllm agent for world data synthesis.
\newblock {\em arXiv preprint arXiv:2508.05580}, 2025.

\bibitem{long2025follow}
Yikuang Yuluo, Yue Ma, Kuan Shen, Tongtong Jin, Wang Liao, Yangpu Ma, and Fuquan Wang.
\newblock Follow-your-shape: Shape-aware image editing via trajectory-guided region control.
\newblock {\em arXiv preprint arXiv:2508.08134}, 2025.

\bibitem{super-clevr}
Zhuowan Li, Xingrui Wang, Elias Stengel-Eskin, Adam Kortylewski, Wufei Ma, Benjamin Van~Durme, and Alan~L Yuille.
\newblock Super-clevr: A virtual benchmark to diagnose domain robustness in visual reasoning.
\newblock In {\em IEEE/CVF conference on computer vision and pattern recognition}, pages 14963--14973, 2023.

\bibitem{geo170k}
Jiahui Gao, Renjie Pi, Jipeng Zhang, Jiacheng Ye, Wanjun Zhong, Yufei Wang, Lanqing Hong, Jianhua Han, Hang Xu, Zhenguo Li, et~al.
\newblock G-llava: Solving geometric problem with multi-modal large language model.
\newblock {\em arXiv preprint arXiv:2312.11370}, 2023.

\bibitem{kazemi2023geomverse}
Mehran Kazemi, Hamidreza Alvari, Ankit Anand, Jialin Wu, Xi~Chen, and Radu Soricut.
\newblock Geomverse: A systematic evaluation of large models for geometric reasoning.
\newblock {\em arXiv preprint arXiv:2312.12241}, 2023.

\bibitem{mavis}
Renrui Zhang, Xinyu Wei, Dongzhi Jiang, Ziyu Guo, Shicheng Li, Yichi Zhang, Chengzhuo Tong, Jiaming Liu, Aojun Zhou, Bin Wei, et~al.
\newblock Mavis: Mathematical visual instruction tuning with an automatic data engine.
\newblock {\em arXiv preprint arXiv:2407.08739}, 2024.

\bibitem{math360k}
Wenhao Shi, Zhiqiang Hu, Yi~Bin, Junhua Liu, Yang Yang, See-Kiong Ng, Lidong Bing, and Roy Ka-Wei Lee.
\newblock Math-llava: Bootstrapping mathematical reasoning for multimodal large language models.
\newblock {\em arXiv preprint arXiv:2406.17294}, 2024.

\bibitem{hong2021transformation}
Xin Hong, Yanyan Lan, Liang Pang, Jiafeng Guo, and Xueqi Cheng.
\newblock Transformation driven visual reasoning.
\newblock In {\em IEEE/CVF Conference on computer vision and pattern recognition}, pages 6903--6912, 2021.

\bibitem{ji2025visualtrans}
Yuheng Ji, Yipu Wang, Yuyang Liu, Xiaoshuai Hao, Yue Liu, Yuting Zhao, Huaihai Lyu, and Xiaolong Zheng.
\newblock Visualtrans: A benchmark for real-world visual transformation reasoning.
\newblock {\em arXiv preprint arXiv:2508.04043}, 2025.

\bibitem{sci}
Pan Lu, Swaroop Mishra, Tanglin Xia, Liang Qiu, Kai-Wei Chang, Song-Chun Zhu, Oyvind Tafjord, Peter Clark, and Ashwin Kalyan.
\newblock Learn to explain: Multimodal reasoning via thought chains for science question answering.
\newblock {\em Advances in Neural Information Processing Systems}, pages 2507--2521, 2022.

\bibitem{ai2d}
Aniruddha Kembhavi, Mike Salvato, Eric Kolve, Minjoon Seo, Hannaneh Hajishirzi, and Ali Farhadi.
\newblock A diagram is worth a dozen images.
\newblock In {\em European Conference on Computer Vision}, pages 235--251, 2016.

\bibitem{hu2023look}
Yingdong Hu, Fanqi Lin, Tong Zhang, Li~Yi, and Yang Gao.
\newblock Look before you leap: Unveiling the power of gpt-4v in robotic vision-language planning.
\newblock {\em arXiv preprint arXiv:2311.17842}, 2023.

\bibitem{hao2025tla}
Peng Hao, Chaofan Zhang, Dingzhe Li, Xiaoge Cao, Xiaoshuai Hao, Shaowei Cui, and Shuo Wang.
\newblock Tla: Tactile-language-action model for contact-rich manipulation.
\newblock {\em arXiv preprint arXiv:2503.08548}, 2025.

\bibitem{programs}
Justin Johnson, Bharath Hariharan, Laurens Van Der~Maaten, Judy Hoffman, Li~Fei-Fei, C~Lawrence~Zitnick, and Ross Girshick.
\newblock Inferring and executing programs for visual reasoning.
\newblock In {\em IEEE international conference on computer vision}, pages 2989--2998, 2017.

\bibitem{suris2023vipergpt}
D{\'\i}dac Sur{\'\i}s, Sachit Menon, and Carl Vondrick.
\newblock Vipergpt: Visual inference via python execution for reasoning.
\newblock In {\em IEEE/CVF International Conference on Computer Vision}, pages 11888--11898, 2023.

\bibitem{cot}
Jason Wei, Xuezhi Wang, Dale Schuurmans, Maarten Bosma, Fei Xia, Ed~Chi, Quoc~V Le, Denny Zhou, et~al.
\newblock Chain-of-thought prompting elicits reasoning in large language models.
\newblock {\em Advances in neural information processing systems}, pages 24824--24837, 2022.

\bibitem{insight-v}
Yuhao Dong, Zuyan Liu, Hai-Long Sun, Jingkang Yang, Winston Hu, Yongming Rao, and Ziwei Liu.
\newblock Insight-v: Exploring long-chain visual reasoning with multimodal large language models.
\newblock {\em arXiv preprint arXiv:2411.14432}, 2024.

\bibitem{deepseek-r1}
Daya Guo, Dejian Yang, Haowei Zhang, Junxiao Song, Ruoyu Zhang, Runxin Xu, Qihao Zhu, Shirong Ma, Peiyi Wang, Xiao Bi, et~al.
\newblock Deepseek-r1: Incentivizing reasoning capability in llms via reinforcement learning.
\newblock {\em arXiv preprint arXiv:2501.12948}, 2025.

\bibitem{tang2025affordgrasp}
Yingbo Tang, Shuaike Zhang, Xiaoshuai Hao, Pengwei Wang, Jianlong Wu, Zhongyuan Wang, and Shanghang Zhang.
\newblock Affordgrasp: In-context affordance reasoning for open-vocabulary task-oriented grasping in clutter.
\newblock {\em arXiv preprint arXiv:2503.00778}, 2025.

\bibitem{zhang2025mapnav}
Lingfeng Zhang, Xiaoshuai Hao, Qinwen Xu, Qiang Zhang, Xinyao Zhang, Pengwei Wang, Jing Zhang, Zhongyuan Wang, Shanghang Zhang, and Renjing Xu.
\newblock Mapnav: A novel memory representation via annotated semantic maps for vlm-based vision-and-language navigation.
\newblock {\em arXiv preprint arXiv:2502.13451}, 2025.

\bibitem{post_training}
Komal Kumar, Tajamul Ashraf, Omkar Thawakar, Rao~Muhammad Anwer, Hisham Cholakkal, Mubarak Shah, Ming-Hsuan Yang, Phillip~HS Torr, Salman Khan, and Fahad~Shahbaz Khan.
\newblock Llm post-training: A deep dive into reasoning large language models.
\newblock {\em arXiv preprint arXiv:2502.21321}, 2025.

\bibitem{post_training_2}
Tianzhe Chu, Yuexiang Zhai, Jihan Yang, Shengbang Tong, Saining Xie, Dale Schuurmans, Quoc~V Le, Sergey Levine, and Yi~Ma.
\newblock Sft memorizes, rl generalizes: A comparative study of foundation model post-training.
\newblock {\em arXiv preprint arXiv:2501.17161}, 2025.

\bibitem{cot_sft}
Jason Wei, Xuezhi Wang, Dale Schuurmans, Maarten Bosma, Fei Xia, Ed~Chi, Quoc~V Le, Denny Zhou, et~al.
\newblock Chain-of-thought prompting elicits reasoning in large language models.
\newblock {\em Advances in neural information processing systems}, pages 24824--24837, 2022.

\bibitem{math_sft}
Ke~Wang, Houxing Ren, Aojun Zhou, Zimu Lu, Sichun Luo, Weikang Shi, Renrui Zhang, Linqi Song, Mingjie Zhan, and Hongsheng Li.
\newblock Mathcoder: Seamless code integration in llms for enhanced mathematical reasoning.
\newblock {\em arXiv preprint arXiv:2310.03731}, 2023.

\bibitem{rl_1}
Daniel~M Ziegler, Nisan Stiennon, Jeffrey Wu, Tom~B Brown, Alec Radford, Dario Amodei, Paul Christiano, and Geoffrey Irving.
\newblock Fine-tuning language models from human preferences.
\newblock {\em arXiv preprint arXiv:1909.08593}, 2019.

\bibitem{rlhf}
Long Ouyang, Jeffrey Wu, Xu~Jiang, Diogo Almeida, Carroll Wainwright, Pamela Mishkin, Chong Zhang, Sandhini Agarwal, Katarina Slama, Alex Ray, et~al.
\newblock Training language models to follow instructions with human feedback.
\newblock {\em Advances in neural information processing systems}, pages 27730--27744, 2022.

\bibitem{rl_2}
Zhiqing Sun, Sheng Shen, Shengcao Cao, Haotian Liu, Chunyuan Li, Yikang Shen, Chuang Gan, Liang-Yan Gui, Yu-Xiong Wang, Yiming Yang, et~al.
\newblock Aligning large multimodal models with factually augmented rlhf.
\newblock {\em arXiv preprint arXiv:2309.14525}, 2023.

\bibitem{rl_math}
Haipeng Luo, Qingfeng Sun, Can Xu, Pu~Zhao, Jianguang Lou, Chongyang Tao, Xiubo Geng, Qingwei Lin, Shifeng Chen, and Dongmei Zhang.
\newblock Wizardmath: Empowering mathematical reasoning for large language models via reinforced evol-instruct.
\newblock {\em arXiv preprint arXiv:2308.09583}, 2023.

\bibitem{rl_3}
Simon Zhai, Hao Bai, Zipeng Lin, Jiayi Pan, Peter Tong, Yifei Zhou, Alane Suhr, Saining Xie, Yann LeCun, Yi~Ma, et~al.
\newblock Fine-tuning large vision-language models as decision-making agents via reinforcement learning.
\newblock {\em Advances in Neural Information Processing Systems}, pages 110935--110971, 2025.

\bibitem{flan}
Jason Wei, Maarten Bosma, Vincent~Y Zhao, Kelvin Guu, Adams~Wei Yu, Brian Lester, Nan Du, Andrew~M Dai, and Quoc~V Le.
\newblock Finetuned language models are zero-shot learners.
\newblock {\em arXiv preprint arXiv:2109.01652}, 2021.

\bibitem{llama3_1}
Abhimanyu Dubey, Abhinav Jauhri, Abhinav Pandey, Abhishek Kadian, Ahmad Al-Dahle, Aiesha Letman, Akhil Mathur, Alan Schelten, Amy Yang, Angela Fan, et~al.
\newblock The llama 3 herd of models.
\newblock {\em arXiv preprint arXiv:2407.21783}, 2024.

\bibitem{nemotron}
Bo~Adler, Niket Agarwal, Ashwath Aithal, Dong~H Anh, Pallab Bhattacharya, Annika Brundyn, Jared Casper, Bryan Catanzaro, Sharon Clay, Jonathan Cohen, et~al.
\newblock Nemotron-4 340b technical report.
\newblock {\em arXiv preprint arXiv:2406.11704}, 2024.

\bibitem{dpo}
Rafael Rafailov, Archit Sharma, Eric Mitchell, Christopher~D Manning, Stefano Ermon, and Chelsea Finn.
\newblock Direct preference optimization: Your language model is secretly a reward model.
\newblock {\em Advances in Neural Information Processing Systems}, pages 53728--53741, 2023.

\bibitem{tulu3}
Nathan Lambert, Jacob Morrison, Valentina Pyatkin, Shengyi Huang, Hamish Ivison, Faeze Brahman, Lester James~V Miranda, Alisa Liu, Nouha Dziri, Shane Lyu, et~al.
\newblock T$\backslash$" ulu 3: Pushing frontiers in open language model post-training.
\newblock {\em arXiv preprint arXiv:2411.15124}, 2024.

\bibitem{deepseekv3}
Aixin Liu, Bei Feng, Bing Xue, Bingxuan Wang, Bochao Wu, Chengda Lu, Chenggang Zhao, Chengqi Deng, Chenyu Zhang, Chong Ruan, et~al.
\newblock Deepseek-v3 technical report.
\newblock {\em arXiv preprint arXiv:2412.19437}, 2024.

\bibitem{grpo}
Zhihong Shao, Peiyi Wang, Qihao Zhu, Runxin Xu, Junxiao Song, Xiao Bi, Haowei Zhang, Mingchuan Zhang, YK~Li, Y~Wu, et~al.
\newblock Deepseekmath: Pushing the limits of mathematical reasoning in open language models.
\newblock {\em arXiv preprint arXiv:2402.03300}, 2024.

\bibitem{lu2021inter}
Pan Lu, Ran Gong, Shibiao Jiang, Liang Qiu, Siyuan Huang, Xiaodan Liang, and Song-Chun Zhu.
\newblock Inter-gps: Interpretable geometry problem solving with formal language and symbolic reasoning.
\newblock {\em arXiv preprint arXiv:2105.04165}, 2021.

\bibitem{zhang2024lmms}
Kaichen Zhang, Bo~Li, Peiyuan Zhang, Fanyi Pu, Joshua~Adrian Cahyono, Kairui Hu, Shuai Liu, Yuanhan Zhang, Jingkang Yang, Chunyuan Li, et~al.
\newblock Lmms-eval: Reality check on the evaluation of large multimodal models.
\newblock {\em arXiv preprint arXiv:2407.12772}, 2024.

\bibitem{qwen2vl}
Peng Wang, Shuai Bai, Sinan Tan, Shijie Wang, Zhihao Fan, Jinze Bai, Keqin Chen, Xuejing Liu, Jialin Wang, Wenbin Ge, Yang Fan, Kai Dang, Mengfei Du, Xuancheng Ren, Rui Men, Dayiheng Liu, Chang Zhou, Jingren Zhou, and Junyang Lin.
\newblock Qwen2-vl: Enhancing vision-language model's perception of the world at any resolution.
\newblock {\em arXiv preprint arXiv:2409.12191}, 2024.

\bibitem{openr1}
Huggingface.
\newblock open-r1: Fully open reproduction of deepseek-r1.
\newblock \url{https://github.com/huggingface/open-r1}, 2025.
\newblock [Online; accessed: 2025-01-24].

\bibitem{vllm}
Woosuk Kwon, Zhuohan Li, Siyuan Zhuang, Ying Sheng, Lianmin Zheng, Cody~Hao Yu, Joseph~E. Gonzalez, Hao Zhang, and Ion Stoica.
\newblock Efficient memory management for large language model serving with pagedattention.
\newblock In {\em ACM SIGOPS 29th Symposium on Operating Systems Principles}, 2023.

\bibitem{Qwen2.5-VL}
Shuai Bai, Keqin Chen, Xuejing Liu, Jialin Wang, Wenbin Ge, Sibo Song, Kai Dang, Peng Wang, Shijie Wang, Jun Tang, Humen Zhong, Yuanzhi Zhu, Mingkun Yang, Zhaohai Li, Jianqiang Wan, Pengfei Wang, Wei Ding, Zheren Fu, Yiheng Xu, Jiabo Ye, Xi~Zhang, Tianbao Xie, Zesen Cheng, Hang Zhang, Zhibo Yang, Haiyang Xu, and Junyang Lin.
\newblock Qwen2.5-vl technical report.
\newblock {\em arXiv preprint arXiv:2502.13923}, 2025.

\bibitem{abdin2024phi}
Marah Abdin, Jyoti Aneja, Hany Awadalla, Ahmed Awadallah, Ammar~Ahmad Awan, Nguyen Bach, Amit Bahree, Arash Bakhtiari, Jianmin Bao, Harkirat Behl, et~al.
\newblock Phi-3 technical report: A highly capable language model locally on your phone.
\newblock {\em arXiv preprint arXiv:2404.14219}, 2024.

\bibitem{chen2024internvl}
Zhe Chen, Jiannan Wu, Wenhai Wang, Weijie Su, Guo Chen, Sen Xing, Muyan Zhong, Qinglong Zhang, Xizhou Zhu, Lewei Lu, et~al.
\newblock Internvl: Scaling up vision foundation models and aligning for generic visual-linguistic tasks.
\newblock In {\em IEEE/CVF conference on computer vision and pattern recognition}, pages 24185--24198, 2024.

\bibitem{meta2024llama3vision}
{Meta AI}.
\newblock Llama 3 at connect 2024: Vision for edge and mobile devices, 2024.
\newblock Accessed: 2025-02-15.

\bibitem{agrawal2024pixtral}
Pravesh Agrawal, Szymon Antoniak, Emma~Bou Hanna, Baptiste Bout, Devendra Chaplot, Jessica Chudnovsky, Diogo Costa, Baudouin De~Monicault, Saurabh Garg, Theophile Gervet, et~al.
\newblock Pixtral 12b.
\newblock {\em arXiv preprint arXiv:2410.07073}, 2024.

\bibitem{hurst2024gpt4o}
Aaron Hurst, Adam Lerer, Adam~P Goucher, Adam Perelman, Aditya Ramesh, Aidan Clark, AJ~Ostrow, Akila Welihinda, Alan Hayes, Alec Radford, et~al.
\newblock Gpt-4o system card.
\newblock {\em arXiv preprint arXiv:2410.21276}, 2024.

\bibitem{team2024gemini}
Gemini Team, Petko Georgiev, Ving~Ian Lei, Ryan Burnell, Libin Bai, Anmol Gulati, Garrett Tanzer, Damien Vincent, Zhufeng Pan, Shibo Wang, et~al.
\newblock Gemini 1.5: Unlocking multimodal understanding across millions of tokens of context.
\newblock {\em arXiv preprint arXiv:2403.05530}, 2024.

\bibitem{yue2024mmmu}
Xiang Yue, Yuansheng Ni, Kai Zhang, Tianyu Zheng, Ruoqi Liu, Ge~Zhang, Samuel Stevens, Dongfu Jiang, Weiming Ren, Yuxuan Sun, et~al.
\newblock Mmmu: A massive multi-discipline multimodal understanding and reasoning benchmark for expert agi.
\newblock In {\em Proceedings of the IEEE/CVF Conference on Computer Vision and Pattern Recognition}, pages 9556--9567, 2024.

\bibitem{realworldqa}
Grok-1.5 Team.
\newblock Grok-1.5 vision preview.
\newblock \url{https://x.ai/news/grok-1.5v}, 2024.
\newblock [Online; accessed: 2024-04-12].

\bibitem{mathvision}
Ke~Wang, Junting Pan, Weikang Shi, Zimu Lu, Houxing Ren, Aojun Zhou, Mingjie Zhan, and Hongsheng Li.
\newblock Measuring multimodal mathematical reasoning with math-vision dataset.
\newblock {\em Advances in Neural Information Processing Systems}, 37:95095--95169, 2024.

\bibitem{hiippala2021ai2d}
Tuomo Hiippala, Malihe Alikhani, Jonas Haverinen, Timo Kalliokoski, Evanfiya Logacheva, Serafina Orekhova, Aino Tuomainen, Matthew Stone, and John~A Bateman.
\newblock Ai2d-rst: a multimodal corpus of 1000 primary school science diagrams.
\newblock {\em Language Resources and Evaluation}, 55:661--688, 2021.

\bibitem{lu2022scienceqa}
Pan Lu, Swaroop Mishra, Tanglin Xia, Liang Qiu, Kai-Wei Chang, Song-Chun Zhu, Oyvind Tafjord, Peter Clark, and Ashwin Kalyan.
\newblock Learn to explain: Multimodal reasoning via thought chains for science question answering.
\newblock {\em Advances in Neural Information Processing Systems}, 35:2507--2521, 2022.

\bibitem{wei2024gita}
Yanbin Wei, Shuai Fu, Weisen Jiang, Zejian Zhang, Zhixiong Zeng, Qi~Wu, James Kwok, and Yu~Zhang.
\newblock Gita: Graph to visual and textual integration for vision-language graph reasoning.
\newblock {\em Advances in Neural Information Processing Systems}, 37:44--72, 2024.

\bibitem{chia2024puzzlevqa}
Yew~Ken Chia, Vernon Toh~Yan Han, Deepanway Ghosal, Lidong Bing, and Soujanya Poria.
\newblock Puzzlevqa: Diagnosing multimodal reasoning challenges of language models with abstract visual patterns.
\newblock {\em arXiv preprint arXiv:2403.13315}, 2024.

\bibitem{lu2021iconqa}
Pan Lu, Liang Qiu, Jiaqi Chen, Tony Xia, Yizhou Zhao, Wei Zhang, Zhou Yu, Xiaodan Liang, and Song-Chun Zhu.
\newblock Iconqa: A new benchmark for abstract diagram understanding and visual language reasoning.
\newblock {\em arXiv preprint arXiv:2110.13214}, 2021.

\bibitem{zhang2019raven}
Chi Zhang, Feng Gao, Baoxiong Jia, Yixin Zhu, and Song-Chun Zhu.
\newblock Raven: A dataset for relational and analogical visual reasoning.
\newblock In {\em Proceedings of the IEEE/CVF conference on computer vision and pattern recognition}, pages 5317--5327, 2019.

\bibitem{chen2021geoqa}
Jiaqi Chen, Jianheng Tang, Jinghui Qin, Xiaodan Liang, Lingbo Liu, Eric~P Xing, and Liang Lin.
\newblock Geoqa: A geometric question answering benchmark towards multimodal numerical reasoning.
\newblock {\em arXiv preprint arXiv:2105.14517}, 2021.

\end{thebibliography}

\clearpage

\section*{NeurIPS Paper Checklist}

\begin{enumerate}

\item {\bf Claims}
    \item[] Question: Do the main claims made in the abstract and introduction accurately reflect the paper's contributions and scope?
    \item[] Answer: \answerYes{} 
    \item[] Justification: 
    See Section~\ref{sec:intro}.


\item {\bf Limitations}
    \item[] Question: Does the paper discuss the limitations of the work performed by the authors?
    \item[] Answer: \answerYes{} 
    \item[] Justification: 
    See Appendix Section~\ref{sec6} Limitations and Societal Impact.


\item {\bf Theory assumptions and proofs}
    \item[] Question: For each theoretical result, does the paper provide the full set of assumptions and a complete (and correct) proof?
    \item[] Answer: \answerNA{} 
    \item[] Justification: {The paper does not include theoretical results or proofs but focuses on architectural innovation and empirical evaluation.}

    \item {\bf Experimental result reproducibility}
    \item[] Question: Does the paper fully disclose all the information needed to reproduce the main experimental results of the paper to the extent that it affects the main claims and/or conclusions of the paper (regardless of whether the code and data are provided or not)?
    \item[] Answer: \answerYes{} 
    \item[] Justification: See Section~\ref{main-sec4}.
    

\item {\bf Open access to data and code}
    \item[] Question: Does the paper provide open access to the data and code, with sufficient instructions to faithfully reproduce the main experimental results, as described in supplemental material?
    \item[] Answer: \answerYes{} 
    \item[] Justification: See supplementary material.


\item {\bf Experimental setting/details}
    \item[] Question: Does the paper specify all the training and test details (e.g., data splits, hyperparameters, how they were chosen, type of optimizer, etc.) necessary to understand the results?
    \item[] Answer: \answerYes{} 
    \item[] Justification: 
    See Section~\ref{main-sec4} Implementation details and Appendix Sec.~\ref{sec2}.


\item {\bf Experiment statistical significance}
    \item[] Question: Does the paper report error bars suitably and correctly defined or other appropriate information about the statistical significance of the experiments?
    \item[] Answer:  \answerYes{} 
    \item[] Justification: We use five various seeds to train the model and report the average accuracy of them to avoid randomness.


\item {\bf Experiments compute resources}
    \item[] Question: For each experiment, does the paper provide sufficient information on the computer resources (type of compute workers, memory, time of execution) needed to reproduce the experiments?
    \item[] Answer: \answerYes{} 
    \item[] Justification:
    See Section~\ref{main-sec4} Implementation details.
    
\item {\bf Code of ethics}
    \item[] Question: Does the research conducted in the paper conform, in every respect, with the NeurIPS Code of Ethics \url{https://neurips.cc/public/EthicsGuidelines}?
    \item[] Answer: \answerYes{} 
    \item[] Justification: 
    We have read the NeurIPS Code of Ethics, and our paper does not have these problems.

\item {\bf Broader impacts}
    \item[] Question: Does the paper discuss both potential positive societal impacts and negative societal impacts of the work performed?
    \item[] Answer: \answerYes{} 
    \item[] Justification: See Appendix~\ref{sec6} Limitations and Societal Impact.
    
\item {\bf Safeguards}
    \item[] Question: Does the paper describe safeguards that have been put in place for responsible release of data or models that have a high risk for misuse (e.g., pretrained language models, image generators, or scraped datasets)?
    \item[] Answer: \answerNA{} 
    \item[] Justification: 
    The paper poses no such risks.

\item {\bf Licenses for existing assets}
    \item[] Question: Are the creators or original owners of assets (e.g., code, data, models), used in the paper, properly credited and are the license and terms of use explicitly mentioned and properly respected?
    \item[] Answer: \answerYes{} 
    \item[] Justification: CC-BY 4.0.

\item {\bf New assets}
    \item[] Question: Are new assets introduced in the paper well documented and is the documentation provided alongside the assets?
    \item[] Answer: \answerNA{} 
    \item[] Justification: The paper does not release new assets. 

\item {\bf Crowdsourcing and research with human subjects}
    \item[] Question: For crowdsourcing experiments and research with human subjects, does the paper include the full text of instructions given to participants and screenshots, if applicable, as well as details about compensation (if any)? 
    \item[] Answer: \answerNA{} 
    \item[] Justification: The paper does not involve crowdsourcing nor research with human subjects.

\item {\bf Institutional review board (IRB) approvals or equivalent for research with human subjects}
    \item[] Question: Does the paper describe potential risks incurred by study participants, whether such risks were disclosed to the subjects, and whether Institutional Review Board (IRB) approvals (or an equivalent approval/review based on the requirements of your country or institution) were obtained?
    \item[] Answer:  \answerNA{} 
    \item[] Justification: The paper does not involve crowdsourcing nor research with human subjects.

\item {\bf Declaration of LLM usage}
    \item[] Question: Does the paper describe the usage of LLMs if it is an important, original, or non-standard component of the core methods in this research? Note that if the LLM is used only for writing, editing, or formatting purposes and does not impact the core methodology, scientific rigorousness, or originality of the research, declaration is not required.
    \item[] Answer:  \answerNo{} 
    \item[] Justification: The core methodology and experiments do not involve LLMs. Any language editing assistance does not impact the scientific contributions.

\end{enumerate}
\newpage
\appendix
\section*{Appendix}

This supplementary material provides additional details on the proposed method and experimental results that could not be included in the main manuscript due to page limitations.
Specifically, this appendix is organized as follows.

\begin{itemize}[left=1em]
\item Sec.~\ref{sec1} provides more details on the evaluation of reasoning tasks and discusses how we collected, filtered, and reconstructed a high-quality dataset.
\item Sec.~\ref{sec2} outlines the models and training processes, providing more detailed experimental specifics.
\item Sec.~\ref{sec3} presents comprehensive experimental results.
\item Sec.~\ref{app:cot-sft-pipeline} details the pipeline of CoT date generation.
\item Sec.~\ref{sec4} presents detailed composition of diffenernt mixed CoT datasets.
\item Sec.~\ref{app:cot-rl-comparison} shows the comparison of CoT quality before and after RL.
\item Sec.~\ref{sec5} includes more visualization cases.
\item Sec.~\ref{sec6} introduces the limitations of our Reason-RFT and its societal impact.
\end{itemize}

\section{Details of Evaluation Reasoning Tasks}
\label{sec1}

\subsection{Visual Counting}
\label{subsec:vc}

\begin{wrapfigure}{r}{0.50\linewidth}
    \centering
    \vspace{-1.5em}
    \includegraphics[width=0.98\linewidth]{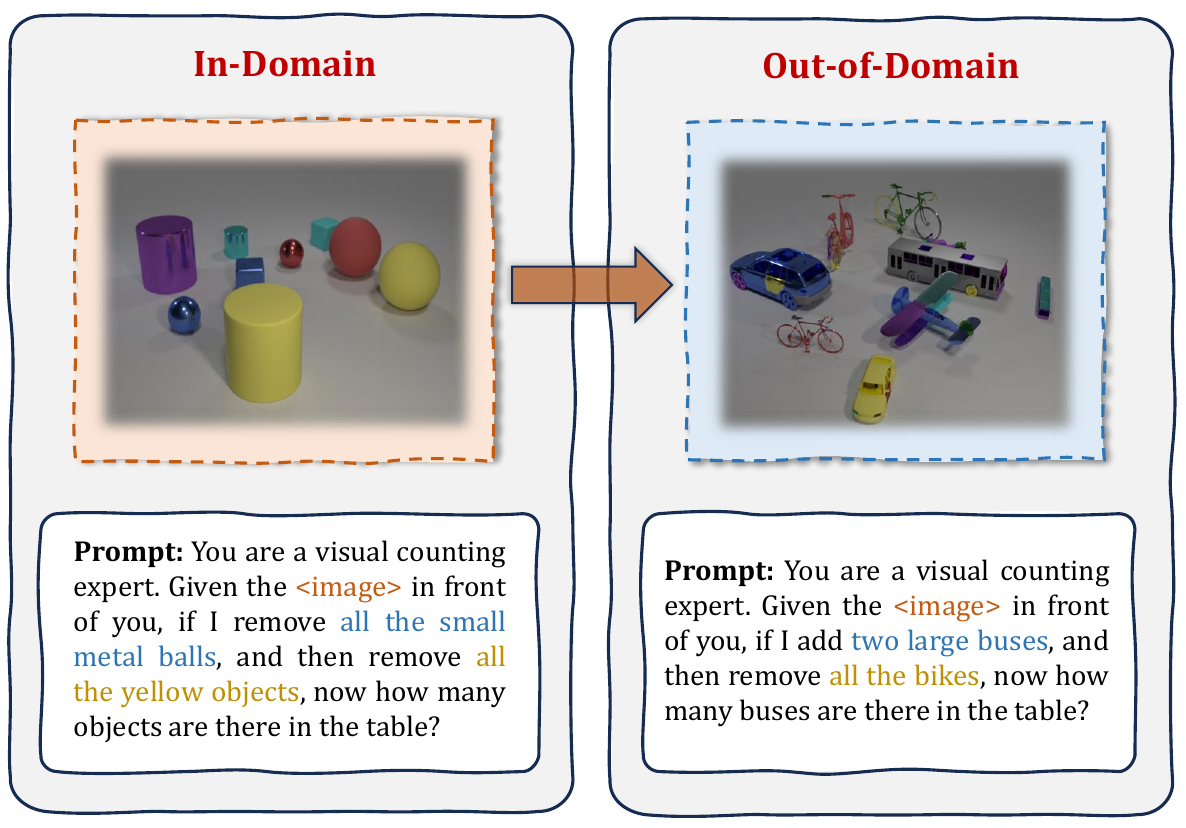}
    \caption{The sample of Visual Counting.}
    \label{fig:task1}
\vspace{-0.5em}
\end{wrapfigure}

\textbf{Task Definition} Visual Counting is a multi-modal reasoning task that evaluates the integration of linguistic, visual, and mathematical capabilities by requiring models to solve arithmetic problems in dynamic visual scenes composed of 3D blocks with diverse attributes, including color, size, material, and shape. The task consists of four distinct reasoning types: \textbf{1) Subtraction}, which involves counting objects after removing a specified subset based on given attributes; \textbf{2) Addition}, where models must compute totals after inserting new objects with defined quantities and properties; \textbf{3) Adversarial}, a challenging variant designed as trick questions in which operations are performed on one set of objects while the query targets an unrelated or unaffected subset, testing the model’s robustness against deceptive scenarios; and \textbf{4) Multi-Hop}, which requires sequential reasoning through multiple addition or subtraction steps to arrive at the final count. This task challenges models to perform attribute-based reasoning in dynamic visual contexts, emphasizing cross-modal understanding and reasoning capabilities. Some examples are shown in Fig.~\ref{fig:task1}.

\textbf{Dataset Preparation} For In-Domain (ID) dataset, we refined the original dataset from CLEVR-Math \cite{clevr-math} by filtering out low-quality or incorrect samples using GPT-4o, resulting in a clean dataset comprising 35K training samples and 1K test samples. These samples are categorized into four specific types: subtraction, addition, adversarial, and multihop-subtraction. To evaluate Domain-Shift (DS) generalization, we extended CLEVR-Math by enhancing the diversity of objects through the incorporation of 3D assets from Super-CLEVR~\cite{super-clevr}, which leads to the creation of Super-CLEVR-Math, an advanced benchmark with 1K test samples designed to assess model generalization under increased complexity. These test samples are also divided into four task types: addition, subtraction, subtraction-multihop and addition-subtraction. The test samples are further categorized into four task types: addition, subtraction, subtraction-multihop, and addition-subtraction. The first three constitute the DS-D subset, while addition-subtraction forms the DS-M subset. Notably, the mixed addition-subtraction type introduces a novel category consisting multi-steps of both addition and subtraction, which is not present in CLEVR-Math, further elevating the benchmark's challenge.

\textbf{Reward Design} Following the reward methodology of DeepSeek-R1 \cite{guo2025deepseek}, we define two distinct reward functions: Format Reward and Accuracy Reward. The Format Reward is assigned a value of 1 if the response adheres to the predefined template structure, specifically in the form of \texttt{<think>...</think><answer>...</answer>}; otherwise, it is assigned a value of 0. The Accuracy Reward is assigned a value of 1 if the numerical counting result in the response is correct; otherwise, it is assigned a value of 0. This dual-reward mechanism ensures both structural compliance and numerical accuracy in model responses.

\subsection{Structure Perception}
\label{subsec:sp}

\begin{wrapfigure}{r}{0.50\linewidth}
    \centering
    \includegraphics[width=0.98\linewidth]{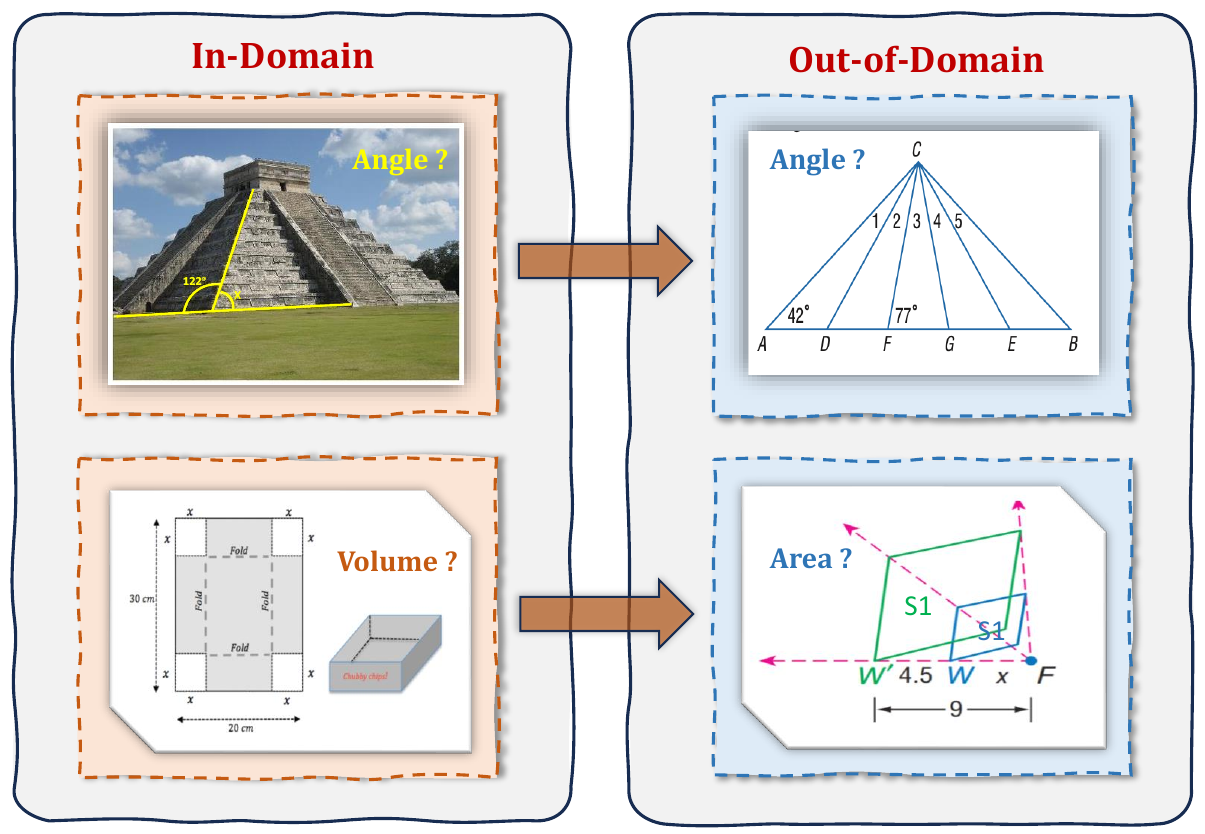}
    \caption{The sample of Structure Perception.}
    \label{fig:task2}
\vspace{-0.5em}
\end{wrapfigure}

\textbf{Task Definition} Structure Perception represents a complex class of visual mathematical reasoning tasks, which focuses on assessing the model's capacity to determine geometric structure relationships and perform calculations involving angles, lengths, areas, and other geometric properties. The task includes problems such as identifying congruent or similar shapes, calculating perimeters and areas, determining angles between lines or shapes, and solving problems related to geometric transformations (e.g., rotations, translations, and reflections). By combining mathematical rigor with visual reasoning, this task challenges models to demonstrate a deep understanding of geometric principles in both abstract and real-world scenarios. Some examples are shown in Fig.~\ref{fig:task2}.

\textbf{Dataset Preparation} For the ID dataset, we provide GeoMath-4K5, a dataset specifically designed for geometric problem solving, which is constructed based on Math360K \cite{math360k} and Geo170K \cite{geo170k}. To ensure data quality, we employed GPT-4o to filter out incorrect samples and removed those with answers that were neither numerical nor included in the provided options, thereby streamlining the validation process during training and testing. This refinement process resulted in a curated dataset consisting of 4.5K training samples and 820 test samples. For DS evaluation, we selected 800 samples from Geometry3K \cite{lu2021inter} (including 400 multiple-choice and 400 open-ended questions)
to comprehensively assess the model's generalization capabilities on geometry reasoning.

\textbf{Reward Design} We maintain the same Format Reward as used in the Visual Counting task above. The Accuracy Reward is extended to support the evaluation of both multiple-choice questions and mathematical expressions, ensuring comprehensive assessment across various problem types. Specifically, mathematical reward type is designed for Structure Perception tasks involving numerical answers, such as floating-point values or LaTeX-formatted expressions. It uses a tolerance-based evaluation to account for minor numerical deviations. The accuracy reward \( R_{\text{acc}}(a_i) \) is defined as:
\begin{small}
\begin{equation}
R_{\text{acc}}(a_i) = \frac{1}{2}\left[\cos\left(\pi \times \frac{|a_{\text{pred}} - a_{\text{gt}}| - \epsilon_1 \times |a_{\text{gt}}|}{(\epsilon_2-\epsilon_1) \times |a_{\text{gt}}|}\right)+1\right],
\end{equation}
\end{small}
where \( a_{\text{pred}} \) is the predicted answer, \( a_{\text{gt}} \) is the ground truth, \( \epsilon_1 \) is the tolerance threshold for an exact match (\textit{e.g.}, 0.05), and \( \epsilon_2 \) is the upper bound for partial rewards (\textit{e.g.}, 0.20). If \( |a_{\text{pred}} - a_{\text{gt}}| < \epsilon_1 \times |a_{\text{gt}}| \), the reward is 1 (exact match); if \( |a_{\text{pred}} - a_{\text{gt}}| > \epsilon_2 \times |a_{\text{gt}}| \), the reward is 0 (incorrect). This formulation ensures smooth transitions between full and partial rewards, enabling fair numerical evaluation.

\subsection{Spatial Transformation}
\label{subsec:st}

\begin{figure*}[t]
    \centering
    \includegraphics[width=0.98\linewidth]{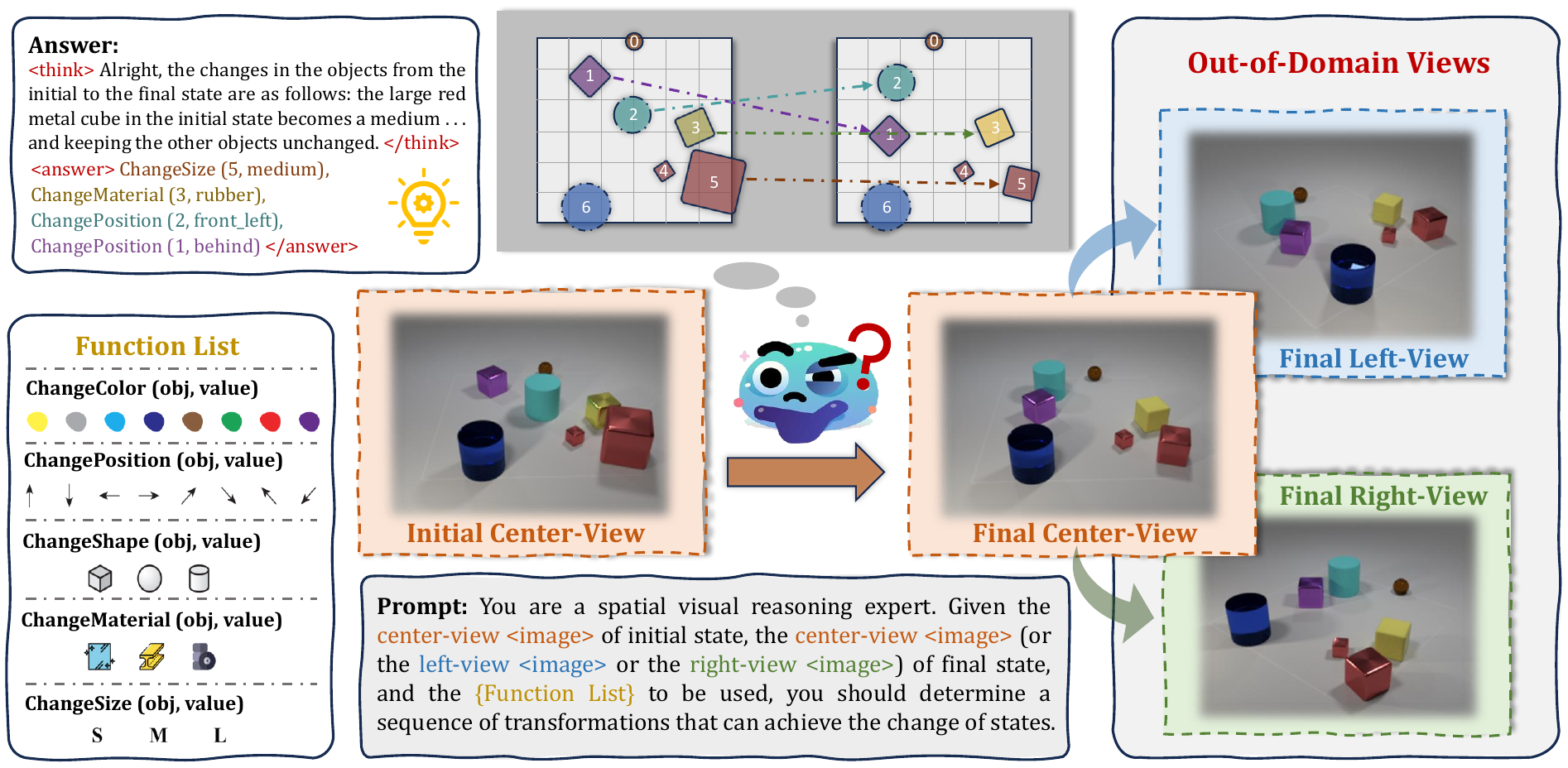}
    \caption{The sample of Spatial Transformation.}
    \label{fig:task3}
    \vspace{-0.5em}
\end{figure*}

\textbf{Task Definition} Spatial Transformation is a spatial-visual reasoning task designed to infer single-step or multi-step transformation actions by analyzing the initial and final visual states from multiple perspectives (\textit{e.g.}, center, left, right). The task utilizes transformation functions, including \texttt{change\_size}, \texttt{change\_color}, \texttt{change\_material}, \texttt{change\_shape}, and \texttt{change\_position}, to modify object properties such as size, color, material, shape, and position using predefined values. This task evaluates the model's ability to reason about spatial relationships and object transformations across diverse viewpoints in dynamic visual scenarios. Some examples are shown in Fig.~\ref{fig:task3}.

\textbf{Dataset Preparation} We generated 100K samples using the environment and configuration from Trance \cite{hong2021transformation}, with each sample comprising initial object attributes, front-view image of initial state, and images of final state captured from front, left, and right perspectives. To ensure high data quality, we implemented a rigorous filtering process: (1) removing samples containing occluded or invisible objects in either the initial or final states, (2) eliminating redundant actions within the transformation sequences, and (3) consolidating multi-step displacement actions, which collectively ensure the uniqueness and correctness of the solutions. The refined dataset consists of 60K training samples and 6K test samples. For the training set, we constructed the Trans-Center-60K dataset using the Center-Center configuration, which pairs front-view initial and final state images. For ID evaluation, we derived the Trans-Center-6K dataset from the 6K test samples under the same Center-Center configuration. To evaluate DS generalization, we constructed two additional datasets: Trans-Left-6K (DS-L) and Trans-Right-6K (DS-R), leveraging the Center-Left and Center-Right configurations to assess the model's generalization capabilities in spatial reasoning under viewpoint conditions.

\textbf{Reward Design} 
For the Format Reward, we adopted the same formulation as used in the Visual Counting task. As for the Accuracy Reward, a specialized design was developed to evaluate the sequence of transformation functions. Function-based type is designed for Spatial Transformation tasks requiring a sequence of transformation functions. The accuracy reward \( R_{\text{acc}}(a_i) \) evaluates the alignment between the predicted sequence \( T_{\text{pred}} \) and the ground truth \( T_{\text{gt}} \), computed as:
\begin{small}
\begin{equation}
\label{eq_task3_1}
R_{\text{acc}}(a_i) = \frac{\text{len}(T_{\text{pred}}^{f+o+v}) + \alpha \cdot \text{len}(T_{\text{pred}}^{f+o/v}) + \beta \cdot \text{len}(T_{\text{pred}}^{f})}{\max(\text{len}(T_{\text{pred}}), \text{len}(T_{\text{gt}}))},
\end{equation}
\end{small}
where \( T_{\text{pred}}^{f+o+v} \) is the subset of transformation steps with complete matches (w/ function, object, and value), \( T_{\text{pred}}^{f+o/v} \) are the subsets with partial and only-function matches (w/ function and object, or w/ function and value), \( T_{\text{pred}}^{f} \) is the subset with only-function matches. \( \alpha \) and \( \beta \) are the weighting coefficients for partial matches. This formulation ensures nuanced evaluation, rewarding both exact and partially correct responses while allowing flexible adjustment of partial match contributions.

\begin{figure*}[!h]
    \centering
    \vspace{7em}
    \includegraphics[width=1.0\linewidth]{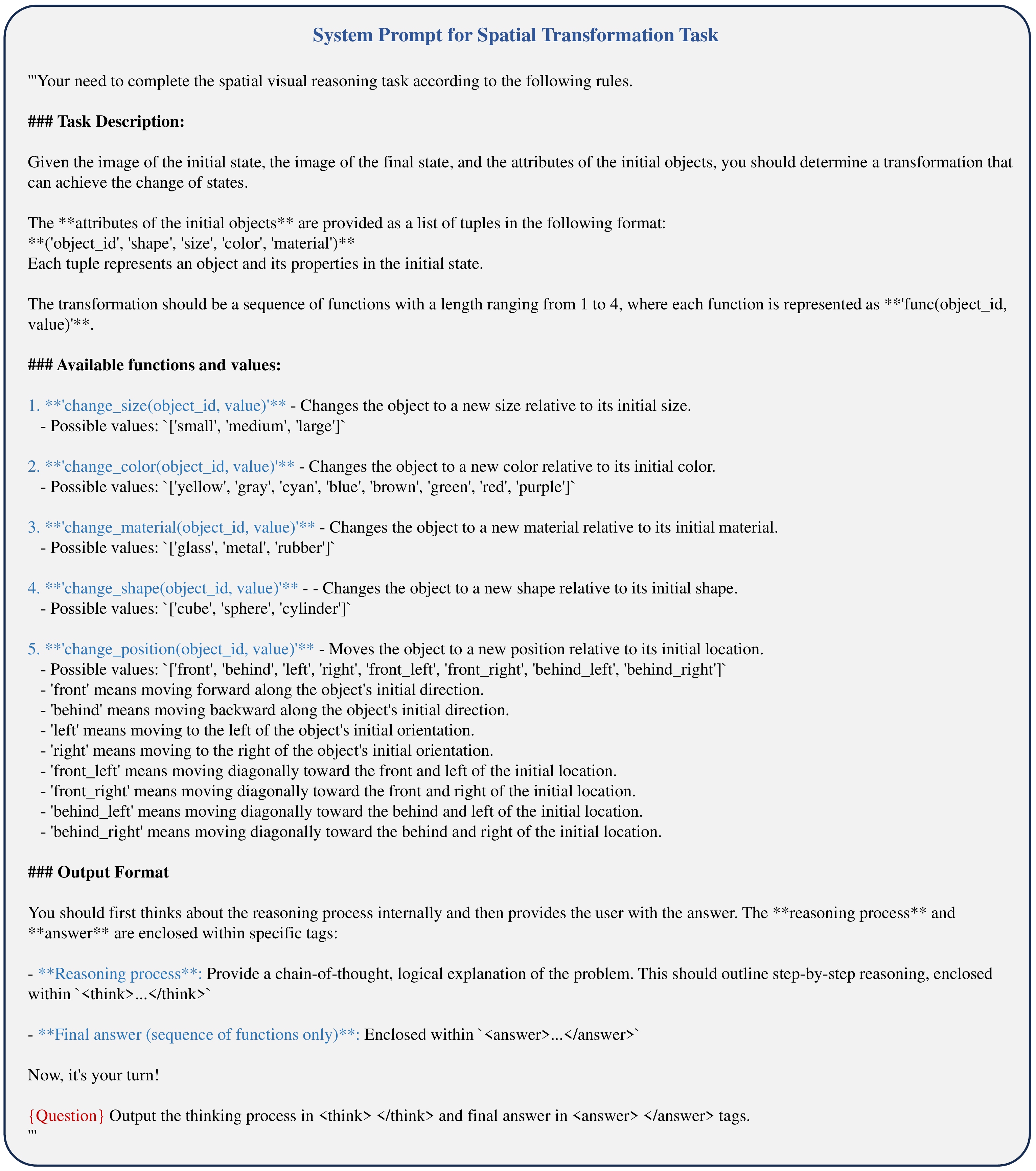}
    \caption{The system prompt used in Spatial Transformation task.}
    \label{fig:prompt_task3}
\end{figure*}

\begin{figure*}[!h]
    \centering
    \includegraphics[width=1.0\linewidth]{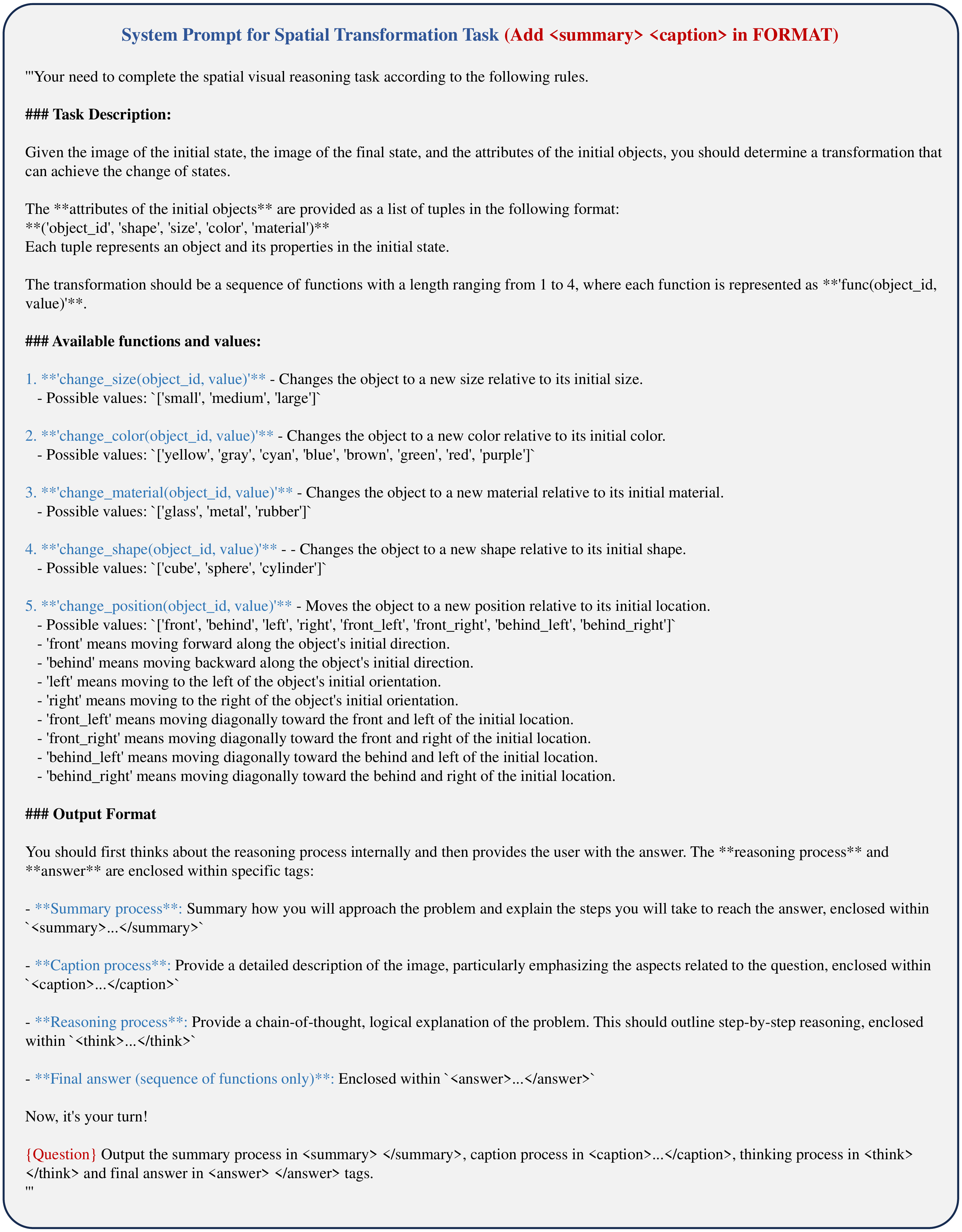}
    \caption{The system prompt used in Spatial Transformation task w/ $\texttt{<summary>}$ and $\texttt{<caption>}$ tags in format.}
    \label{fig:prompt_task3_caption}
\end{figure*}

\textbf{System Prompts} For the Spatial Transformation task, we designed two versions of the system prompt. The first version specifies the answer output format using the $\texttt{<think>}$ and $\texttt{<answer>}$ tags, while the second version includes additional outputs $\texttt{<summary>}$ and $\texttt{<caption>}$ for experiments on exploration of format reward design in the main paper. These two versions are illustrated in Fig. \ref{fig:prompt_task3} and Fig. \ref{fig:prompt_task3_caption}, respectively.

\section{Details of Models and Training}
\label{sec2}
We utilize Qwen2-VL-2B and Qwen2-VL-7B \cite{qwen2vl} as the backbone models for our experiments. Our implementation is built on the open-source frameworks Open-R1~\cite{openr1} and vLLM~\cite{vllm}, ensuring reproducibility and scalability. All experiments were conducted on a cluster of servers, each equipped with 8$\times$A800 GPUs. For the Visual Counting task and Spatial Transformation task, we trained the models for 1 epoch each on their respective training datasets, ensuring sufficient exposure to task-specific patterns while avoiding overfitting. For the Structure Perception task, due to its GeoMath training dataset consisting of a relatively small number of training samples (a total of 4,500), we extended the training duration to 5 epochs to allow the models to fully capture the underlying structural and geometric relationships. In the Reason-RFT training pipeline, all models underwent an initial CoT activation stage with 1,600 samples before proceeding to the RL phase. More details on training process of each models are shown in Tab.~\ref{tab:training_setting}

\begin{table*}[t]
\vspace{-1em}
    \caption{\textbf{Detailed configuration for each training stage of Reason-RFT.} The table presents the training parameters for the 2B model and 7B model across three visual reasoning tasks. The parameters marked with $^*$ correspond to Visual Counting / Structure Perception / Spatial Transformation.}
    \vspace{0.2cm}
    \label{tab:training_setting}
    \setlength{\tabcolsep}{12pt}
    \renewcommand{\arraystretch}{1.2}
    \resizebox{0.96\textwidth}{!}{%
    \begin{tabular}{@{}ll|c|c|c|c}
        \toprule
        & &  \multicolumn{2}{c|}{\textbf{Qwen2-VL-2B}} & \multicolumn{2}{c}{\textbf{Qwen2-VL-7B}} \\ \cmidrule(l){3-3} \cmidrule(l){4-4} \cmidrule(l){5-6}
        & & \textbf{Stage-1} & \textbf{Stage-2} & \textbf{Stage-1} & \textbf{Stage-2} \\
        \midrule 
        \multirow{2}{*}{\rotatebox[origin=c]{90}{\small \textit{Data}}}
        & \textbf{Dataset}  & CoT dataset & Non-CoT dataset & CoT dataset & Non-CoT dataset \\
        & \#Samples & 1.6K & 35K / 4.5K / 60K $^*$ & 1.6K & 35K / 4.5K / 60K $^*$\\
        \midrule 
        \multirow{2}{*}{\rotatebox[origin=c]{90}{\small \textit{Model}}}
        & \textbf{Trainable Part} & Full Model & Full Model & Full Model & Full Model \\
        & \#Tunable Parameters & 2.21B & 2.21B & 8.29B & 8.29B \\
        \midrule 
        \multirow{14}{*}{\rotatebox[origin=c]{90}{\small \textit{Training}}}
        & \textbf{Per-device Batch Size} & 8 & 1 & 8 & 1 \\
        & \textbf{Gradient Accumulation} & 2 & 2 & 2 & 2 \\   
        & \textbf{LR: $\{\psi_v^{\text{ViT}}, \phi_v^{\text{LLM}}\}$} & 1$\times 10^{-5}$ & 1 $\times 10^{-6}$ & 1 $\times 10^{-5}$ & 1 $\times 10^{-6}$ \\
        & \textbf{Epoch} & 1 & 1 / 5 / 1 $^*$ & 1 & 1 / 5 / 1 $^*$\\
        & \textbf{Optimizer} & AdamW & AdamW & AdamW & AdamW \\
        & \textbf{Deepspeed} & Zero3 & Zero3 & Zero3 & Zero3 \\
        & \textbf{Weight Decay} & 0.1 & 0.0 & 0.1 & 0.0 \\
        & \textbf{Warmup Ratio} & 0.03 & 0.00 & 0.03 & 0.00 \\
        & \textbf{LR Schedule} & Cosine & Cosine & Cosine & Cosine \\
        & \textbf{Max Seq. Length} & 32768 & 4096 & 32768 & 4096 \\
        & \textbf{Max Compl. Length} & -- & 512 & -- & 512 \\
        & \textbf{Num. of Compl.} & -- & 8 & -- & 4 \\
        & \textbf{GPU Nums} & 1 $\times$ 8 & 1 $\times$ 8 & 1 $\times$ 8 & 1 $\times$ 8 \\
        \bottomrule
    \end{tabular}
    }
    \vspace{-1.5em}
\end{table*}

\section{More Experiment Results}
\label{sec3}

\begin{wraptable}{r}{0.55\linewidth}
\centering
\vspace{-2em}
\caption{Results of various mixed CoT activation datasets on the Structure Perception task.}
\scalebox{0.75}{
\begin{tabular}{cc|ccc}
\toprule
\textbf{Setting} & \textbf{CoT Activation Data} & \textbf{ID} & \textbf{DS} & \textbf{AVG} \\ \midrule

Baseline                          & GeoMath-only data                           &  59.27 &  49.25  & 54.26   \\ \midrule
(a)                               & Mixed Specific-Domain data                    &  50.61 &  45.35  & 48.02   \\
(b)                               & Mixed General-Domain data                     &  42.51 &  40.25  & 41.38 \\ \bottomrule
\end{tabular}}
\label{cot_data}
\vspace{-1em}
\end{wraptable}

\textbf{Exploration on COT Activation Data} To investigate the impact of differently composed CoT activation data on Reason-RFT, we construct two distinct datasets: a mixed domain-specific dataset, which integrates relevant yet distinct data from in-domain tasks, and a mixed general-domain dataset, encompassing a broader range of visual reasoning tasks (\textit{e.g.}, graph topology, visual puzzles). The detailed dataset composition is provided in Appendix Sec.~\ref{sec4}. Using these datasets, we perform Reason-RFT training on Structure Perception task, with the results detailed in Tab.~\ref{cot_data}. From this, two key points emerge: (1) As the proportion of in-domain training data decreases, the model's performance on specific tasks declines; (2) Models trained on more diverse visual reasoning domain data may also exhibit a reduction in domain-specific performance.

\textbf{Results on Different Backbones}
\label{app:more-backbones}
We further validate the effectiveness of \emph{Reason-RFT} on stronger or alternative vision--language backbones. We report results on three visual reasoning tasks: Visual Counting (T1), Structure Perception (T2), and Spatial Transformation (T3) in the combined \mbox{Tab.~\ref{tab:backbones_combined}}. \emph{Reason\mbox{-}RFT} achieves the strongest averages across backbones and tasks, with especially large margins on domain-shifted splits while keeping in-domain (ID) performance near ceiling. On \textbf{\emph{Qwen2.5\mbox{-}VL\mbox{-}3B}}, for \textbf{T1} (Visual Counting) ID is already saturated (99.0 with Reason-RFT-Zero vs.\ 98.8 with Reason-RFT), yet \emph{Reason-RFT} markedly improves robustness on DS: \(+\!9.2\) on DS-D (68.7 vs.\ 59.5 vs.\ CoT-SFT) and \(+\!5.6\) on DS-M (54.8 vs.\ 49.2 vs.\ CoT-SFT), with an especially large \(+\!44.0\) over Reason-RFT-Zero on DS-M (54.8 vs.\ 10.8), yielding the best T1 AVG (74.1); for \textbf{T2} (Structure Perception) it is best on both ID/DS (59.0/56.6), beating CoT-SFT by \(+\!2.9\) (ID) and \(+\!7.2\) (DS) and Reason-RFT-Zero by \(+\!4.2\) (ID) and \(+\!2.1\) (DS), indicating that RL enhances stepwise structural reasoning rather than overfitting; for \textbf{T3} (Spatial Transformation) ANS-SFT attains the top ID (91.1) but is less robust, whereas \emph{Reason-RFT} trades a modest \(-4.4\) on ID (86.7) for substantial DS gains of \(+\!8.2\) on DS-L (55.2 vs.\ 47.0) and \(+\!7.6\) on DS-R (54.4 vs.\ 46.8), delivering the best AVG (65.4) and a superior ID/DS Pareto. On \textbf{\emph{InternVL3\mbox{-}2B}} for \textbf{T1}, \emph{Reason-RFT} is best on ID/DS-D/DS-M (99.10/69.80/55.90) and AVG (74.93), improving over ANS-SFT, CoT-SFT, and Reason-RFT-Zero by \(+\!23.16\), \(+\!8.53\), and \(+\!18.06\), respectively, with the largest domain-shift margin on DS-M of \(+\!43.60\) over Reason-RFT-Zero (55.90 vs.\ 12.30). Taken together, these trends across architecturally distinct backbones indicate that the benefits of \emph{Reason-RFT} are backbone-agnostic, improving both coherence-driven reasoning and out-of-distribution reliability.

\begin{table*}[t]
    \scriptsize
    \centering
    \caption{\textbf{Results on different backbones across three tasks.}
    Best is \textbf{bold}; second-best is \underline{underlined}. ``ID'' denotes in-domain; ``DS-*'' denotes domain-shifted splits. Missing results are shown as ``--''.}
    \vspace{0.5em}
    \scalebox{0.88}{
    \begin{tabular}{l|l|cccc|ccc|cccc}
        \toprule
        \multicolumn{1}{l|}{\multirow{2}{*}{\textbf{Backbone}}} & \multicolumn{1}{c|}{\multirow{2}{*}{\textbf{Method}}} & \multicolumn{4}{c|}{\textbf{Visual Counting (T1)}} & \multicolumn{3}{c|}{\textbf{Structure Perception (T2)}} & \multicolumn{4}{c}{\textbf{Spatial Transformation (T3)}} \\
        \cmidrule(lr){3-6}\cmidrule(lr){7-9}\cmidrule(lr){10-13}
        & & \textbf{ID} & \textbf{DS-D} & \textbf{DS-M} & \textbf{AVG} & \textbf{ID} & \textbf{DS} & \textbf{AVG} & \textbf{ID} & \textbf{DS-L} & \textbf{DS-R} & \textbf{AVG} \\
        \midrule
        \multirow{5}{*}{\textbf{Qwen2.5-VL-3B}} 
            & Zero-Shot             & 75.9 & 50.9 & 4.4  & 43.7 & 36.8 & 37.4 & 37.1 & 8.6  & 8.3  & 8.3  & 8.4  \\
            & $+$ ANS-SFT           & 97.4 & 51.5 & 6.0  & 51.6 & 53.0 & 31.8 & 42.4 & \textbf{91.1} & 47.0 & 46.8 & \underline{61.6} \\
            & $+$ CoT-SFT           & 89.2 & \underline{59.5} & \underline{49.2} & \underline{66.0} & \underline{56.1} & 49.4 & 52.7 & 81.6 & 46.1 & 44.2 & 57.3 \\
            & $+$ Reason-RFT-Zero   & \underline{99.0} & 58.9 & 10.8 & 56.2 & 54.8 & \underline{54.5} & \underline{54.6} & 68.5 & \underline{49.5} & \underline{48.0} & 55.3 \\
        \rowcolor[HTML]{DAEFF9}
            & $+$ Reason-RFT        & 98.8 & \textbf{68.7} & \textbf{54.8} & \textbf{74.1} & \textbf{59.0} & \textbf{56.6} & \textbf{57.8} & \underline{86.7} & \textbf{55.2} & \textbf{54.4} & \textbf{65.4} \\
        \midrule
        \multirow{5}{*}{\textbf{InternVL3-2B}}
            & Zero-Shot             & 79.30 & 51.20 & 5.10  & 45.20 & -- & -- & -- & -- & -- & -- & -- \\
            & $+$ ANS-SFT           & 96.80 & 52.00 & 6.50  & 51.77 & -- & -- & -- & -- & -- & -- & -- \\
            & $+$ CoT-SFT           & 88.90 & \underline{60.10} & \underline{50.20} & \underline{66.40} & -- & -- & -- & -- & -- & -- & -- \\
            & $+$ Reason-RFT-Zero   & 98.90 & 59.40 & 12.30 & 56.87 & -- & -- & -- & -- & -- & -- & -- \\
        \rowcolor[HTML]{DAEFF9}
            & $+$ Reason-RFT        & \textbf{99.10} & \textbf{69.80} & \textbf{55.90} & \textbf{74.93} & -- & -- & -- & -- & -- & -- & -- \\
        \bottomrule
    \end{tabular}}
    \label{tab:backbones_combined}
    \vspace{-1.75em}
\end{table*}

\begin{wraptable}{r}{0.55\linewidth}
\centering
\vspace{-1em}
\caption{Evaluation results on general benchmarks.}
 \scalebox{0.68}{
\begin{tabular}{l|cccc}
\toprule
\multicolumn{1}{l|}{\multirow{2}{*}{\textbf{Method}}} & \multicolumn{4}{c}{\textbf{General}}   \\ 
\cmidrule(l){2-5}  
& \textbf{MMMU}    & \textbf{RealWorldQA}    & \textbf{MathVision}  & \textbf{AI2D}  \\ \midrule
\rowcolor[HTML]{F2F2F2} \multicolumn{5}{l}{\textbf{Qwen2VL-2B-Instruct}} \\ \midrule
Zero-Shot            & 39.89          & \textbf{61.31}           & 12.50                  &72.50                      \\
$+$ ANS-SFT         & \underline{40.56}          & 48.76           & \textbf{15.79}                  & 68.20                      \\
$+$ CoT-SFT         & 34.00          & 37.78           & 12.99                  & 65.36                      \\
$+$ Reason-RFT-Zero &   39.30        & 42.81           & 13.00                  & \underline{74.61}                  \\
\rowcolor[HTML]{DAEFF9} {$+$ Reason-RFT} & \textbf{41.14}          & \underline{53.06}           & \underline{14.82}              & \textbf{75.24}           \\ \midrule
\rowcolor[HTML]{F2F2F2} \multicolumn{5}{l}{\textbf{Qwen2VL-7B-Instruct}} \\ \midrule
Zero-Shot             & \textbf{54.10}          & \textbf{67.19}           & \underline{16.30}                  & \textbf{83.00}           \\
$+$ ANS-SFT           & 42.66          & 48.10           & 9.12                  & 78.30           \\
$+$ CoT-SFT           & 44.67          & 36.46           & 15.30                  & 73.25           \\
$+$ Reason-RFT-Zero   & 46.44          & 45.10           & 10.86                  & 75.28           \\
\rowcolor[HTML]{DAEFF9} {$+$ Reason-RFT} & \underline{50.04}      & \underline{61.31}    & \textbf{17.60}           & \underline{81.70}       \\
\bottomrule
\end{tabular}}
\label{add_sub}
\end{wraptable}

\textbf{Evaluation on General Benchmarks} Although \textit{Reason-RFT} is primarily designed to enhance domain-specific visual reasoning abilities, we conduct a thorough evaluation on general benchmarks to verify whether our approach compromises the model's general reasoning capabilities. Tab.~\ref{add_sub} presents the results on four widely adopted datasets: {MMMU}~\cite{yue2024mmmu}, {RealWorldQA}~\cite{realworldqa}, {MathVision}~\cite{mathvision}, and {AI2D}~\cite{hiippala2021ai2d}. Across all tasks and model scales, \textit{Reason-RFT} consistently maintains or even improves general performance. For instance, on the 2B model, \textit{Reason-RFT} achieves the highest scores on {MMMU} (41.14) and {AI2D} (75.24), outperforming both zero-shot baselines and other supervised fine-tuning approaches such as ANS-SFT and CoT-SFT. Notably, it also improves performance on the challenging {MathVision} task (14.82), demonstrating its robustness in spatial reasoning. For the larger 7B model, \textit{Reason-RFT} again surpasses ANS-SFT and CoT-SFT by large margins, particularly on {RealWorldQA} (61.31) and {MMMU} (50.04), while maintaining strong results on {AI2D} (81.70). These results suggest that \textit{Reason-RFT} not only scales effectively with model size but also introduces no observable performance degradation on general benchmarks. In summary, the empirical evidence supports that \textit{Reason-RFT} enhances domain-specific reasoning while preserving—if not enhancing—general visual-language reasoning capabilities. This confirms the robustness and transferability of our method, making it a strong alternative to conventional fine-tuning paradigms.

\begin{figure*}[t]
    \centering
    \includegraphics[width=1.0\linewidth]{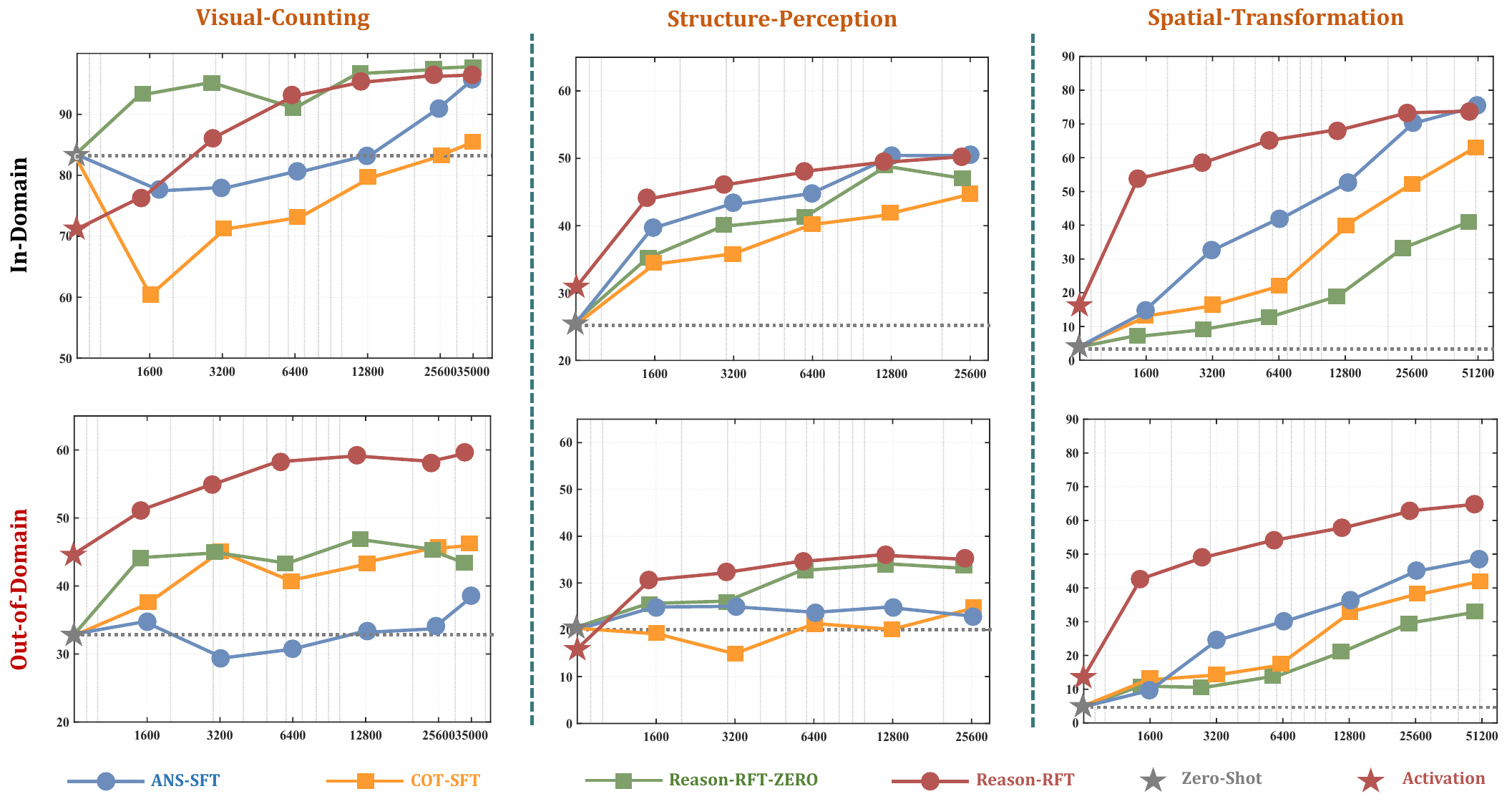}
    \caption{Results of all methods on Qwen2VL-2B-Instruct, ID and DS performance at different training checkpoints.}
    \label{fig:eval_1}
\end{figure*}

\begin{figure*}[t]
    \centering
    \includegraphics[width=1.0\linewidth]{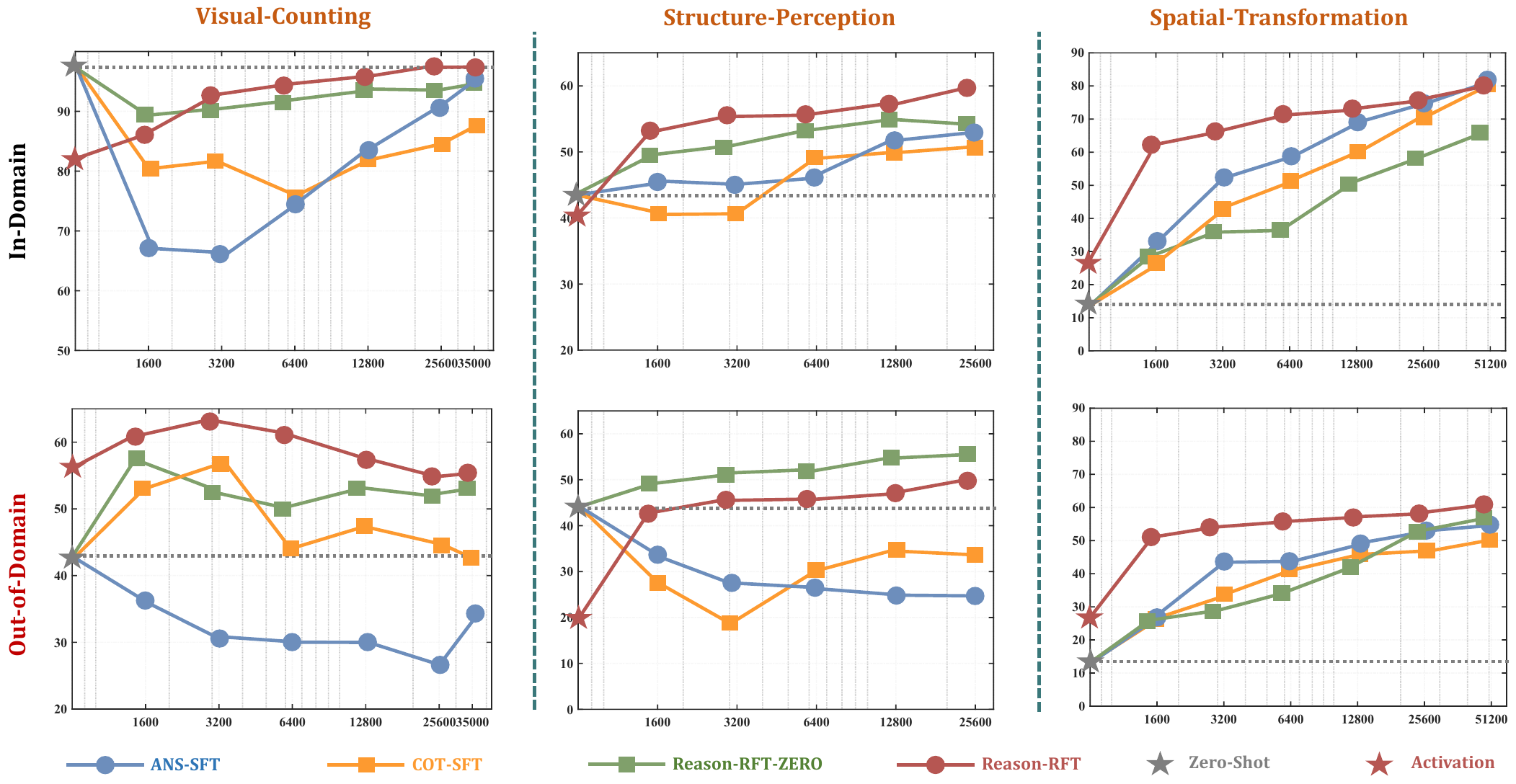}
    \caption{Results of all methods on Qwen2VL-7B-Instruct, ID and DS performance at different training checkpoints.}
    \label{fig:eval_2}
\end{figure*}

\textbf{Performance at Different Training Steps} Fig.~\ref{fig:eval_1} and Fig.~\ref{fig:eval_2} illustrate the ID and DS performance of all the training methods across three visual reasoning tasks, evaluated at various training sample sizes. This analysis helps us understand how each method scales with training data. More detail evaluation results for each subset of three tasks are in Tab.~\ref{tab:visual_counting_id_2b} - Tab.~\ref{tab:trance_7b}.
We systematically varied the number of training samples, from minimal to substantial, allowing us to identify performance thresholds and data efficiency for each method in both ID and DS contexts.
Key findings from this analysis include: Data Efficiency of Reason-RFT: \textit{Reason-RFT} demonstrates exceptional data efficiency, achieving approximately 70\% of the performance of \textit{Reason-RFT-Zero} with only 3\% of the training data (1,600 samples), and 82.5\% with just 9\%.
Robust Generalization to DS scenarios: In the 7B model, \textit{Reason-RFT} achieves over 92\% of \textit{Reason-RFT-Zero}'s performance using just 3\% of the training data, showcasing its strong generalization capabilities.
Comparison Across Methods: \textit{Reason-RFT} consistently outperforms other methods, particularly in data-constrained scenarios, indicating its suitability for applications with limited data availability.
Performance Saturation: As training sample size increases, some methods experience performance plateaus, suggesting that beyond a certain point, additional data yields diminishing returns.

In conclusion, the evaluation of performance across different training samples not only highlights the strengths of \textit{Reason-RFT} in terms of data efficiency and generalization but also provides critical insights into the performance dynamics of various methods. These findings are essential for practitioners aiming to maximize performance while effectively managing training resources.

\section{More Details on CoT Data Construction}
\label{app:cot-sft-pipeline}

This section expands the pipeline of CoT generation by detailing both the automated and manual components used to construct our \emph{CoT-SFT} corpus.

\textbf{(1) Automated Generation.}
We instantiate CoT drafts using reasoning-guided prompt templates such as
\emph{``Let's break down the problem step by step\ldots''} and
\emph{``To answer this, we need to consider\ldots''}.
Templates are combined with model prompting (GPT-4o~\cite{hurst2024gpt4o} and Gemini-Pro~\cite{team2024gemini}) under temperature-controlled sampling
\((T{=}0.7,~\text{top-}k{=}50,~\text{top-}p{=}0.9)\).
To increase coverage and depth, we insert hand-crafted, subtask-specific few-shot exemplars that bias toward explicit intermediate justifications and error-checking behavior.

\textbf{(2) Automated Filtering.}
Each generated CoT is screened by two criteria:

\textbf{\emph{Length range.}}
For each subtask \(s\), we compute a target trajectory length \(\bar{L}_s\) from a balanced mixture of 50\% human-written and 50\% model-generated samples:
\[
\bar{L}_s \;=\; \tfrac{1}{2}\big(\bar{L}^{\text{human}}_{s} + \bar{L}^{\text{model}}_{s}\big),
\]
\noindent where \(\bar{L}^{\text{human}}_{s}\) and \(\bar{L}^{\text{model}}_{s}\) are computed as sample means over their respective sets after basic de-duplication.
A candidate with length \(L_i\) is retained iff
\[
0.6\,\bar{L}_s \;\le\; L_i \;\le\; 1.4\,\bar{L}_s,
\]
\noindent where \(L_i\) is measured in tokens by our training-time tokenizer; the factors \(0.6\) and \(1.4\) were selected via a small pilot study to trim outliers while preserving diversity.

\textbf{\emph{Inconsistency.}}
We discard the responses that contradict the known ground truth, including self-inconsistent counts, incompatible algebraic steps, or reasoning that invalidates later conclusions.

For reference, the empirical trajectory-length statistics (mean \(\mu\), stdev \(\sigma\)) across tasks are:
\begin{center}
 \scalebox{0.8}{
\begin{tabular}{lcc}
\toprule
\textbf{Task} & \(\mu\) (tokens) & \(\sigma\) (tokens) \\
\midrule
Visual Counting & 70  & 30 \\
Structural Perception & 180 & 80 \\
Spatial Transformation & 400 & 120 \\
\bottomrule
\end{tabular}}
\end{center}
\noindent where  \(\mu\) and \(\sigma\) are computed over the curated pool \emph{after} automated filtering and \emph{before} manual review.

\textbf{(3) Human Verification.}
We randomly sample \(10\%\) of CoT drafts from eachtask for manual review, focusing on (i) step-to-step coherence, (ii) logical validity, and (iii) alignment between the reasoning chain and the final answer.
Typical failure modes include:
(i) a correct final answer supported by an incorrect chain (e.g., deriving triangle area via the Pythagorean theorem);
(ii) internal contradictions, such as stating ``There are 3 red blocks on the left and 2 on the right'' and later concluding the total is 6.
A follow-up quality audit found that, prior to human verification, approximately \(3.8\%\) of samples contained critical logical flaws; after verification, the residual error rate fell below \(1\%\), indicating high post-cleanup reliability.

\textbf{Discussion.}
Automated generation with calibrated sampling and subtask-specific few-shots provides diverse yet structured CoTs; the length- and consistency-based filters remove overly terse/verbose or self-contradictory drafts; targeted human verification further suppresses high-severity errors. Together, these stages yield a CoT--SFT dataset with improved coherence and faithfulness, while maintaining scalability and reproducibility.

\section{Detail on Mixed CoT Datasets}

As shown in Tab. \ref{tab:datasets}, we presents a comprehensive overview of the datasets utilized for all of our visual reasoning experiments, categorized into three experimental groups. All of them are CoT-annotated by GPT-4o~\cite{hurst2024gpt4o} The Main Experiment section includes three large-scale datasets: Visual-Counting (35,000 samples) for quantitative analysis, Structure-Perception (4,500 samples) for structural understanding, and Spatial-Transformation (60,000 samples) assessing spatial reasoning capabilities. 
For Ablation Studies, two mixed-domain subsets were constructed: (1) The Mixed General-Domain set comprises 11 CoT-annotated datasets spanning scientific reasoning (AI2D~\cite{ai2d}, ScienceQA~\cite{lu2022scienceqa}), topological graph problems (GVLQA series~\cite{wei2024gita}), and pattern recognition (PuzzleVQA~\cite{chia2024puzzlevqa}, IconQA~\cite{lu2021iconqa}, Raven~\cite{zhang2019raven}). (2) The Mixed Specific-Domain set focuses exclusively on geometric reasoning, featuring GeoQA~\cite{chen2021geoqa}, GeomVerse~\cite{kazemi2023geomverse}, and Geometry3K~\cite{lu2021inter} with progressively complex problem structures. All datasets were standardized to ensure training compatibility.

\label{sec4}
\begin{table}[h]
\centering
\caption{Datasets Overview for Visual Reasoning Tasks}
 \scalebox{0.85}{
\begin{tabular}{llll}
\toprule
\textbf{Dataset Name} & \textbf{Samples} & \textbf{Reasoning Type} & \textbf{Description} \\
\midrule
\rowcolor[HTML]{F2F2F2} \multicolumn{4}{l}{\textbf{Main Experiment}} \\ \midrule
Visual-Counting & 35,000 & Visual Counting & Full dataset for visual counting task \\
Structure-Perception & 4,500 & Structure Perception & Full dataset for structural perception tasks \\
Spatial-Transformation & 60,000 & Spatial Transformation & Full dataset for spatial transformation tasks \\
\midrule
\rowcolor[HTML]{F2F2F2} \multicolumn{4}{l}{\textbf{Ablation Experiment (Mixed General-Domain)}} \\ \midrule
AI2D~\cite{ai2d} & 1,467 & Scientific Reasoning & Scientific diagram interpretation \\
ScienceQA~\cite{lu2022scienceqa} & 2,112 & Scientific Reasoning & Science question answering \\
GVLQA-connectivity~\cite{wei2024gita} & 1,199 & Topological Reasoning & Graph connectivity problems \\
GVLQA-cycle~\cite{wei2024gita} & 1,194 & Topological Reasoning & Cycle detection in graphs \\
GVLQA-hamilton~\cite{wei2024gita} & 1,158 & Topological Reasoning & Hamiltonian path problems \\
GVLQA-topology~\cite{wei2024gita} & 1,070 & Topological Reasoning & General topology questions \\
GVLQA-matching~\cite{wei2024gita} & 1,193 & Topological Reasoning & Graph matching tasks \\
PuzzleVQA~\cite{chia2024puzzlevqa} & 1,618 & Pattern/Puzzle & Visual puzzle solving \\
IconQA~\cite{lu2021iconqa} & 5,270 & Pattern/Puzzle & Icon-based question answering \\
Raven~\cite{zhang2019raven} & 982 & Pattern/Puzzle & Raven's Progressive Matrices \\
\midrule
\rowcolor[HTML]{F2F2F2} \multicolumn{4}{l}{\textbf{Ablation Experiment (Mixed Specific-Domain)}} \\ \midrule
GeoQA~\cite{chen2021geoqa} & 1,500 & Geometric Reasoning & Geometric problem solving \\
GeomVerse~\cite{kazemi2023geomverse} & 2,841 & Geometric Reasoning & Advanced geometry challenges \\
Geometry3K~\cite{lu2021inter} & 3,794 & Geometric Reasoning & Comprehensive geometry problems \\
\bottomrule
\end{tabular}}
\label{tab:datasets}
\vspace{-1em}
\end{table}

\section{Comparison of CoT Quality Before and After RL}\label{app:cot-rl-comparison}

\textbf{Setting.}
We compare the \textbf{Qwen2VL-3B} model trained with \emph{Reason-RFT} (Stage~2, post-RL) against the same backbone trained with only \emph{CoT-SFT} (Stage~1, pre-RL) on the Structure Perception task. Unless otherwise noted, statistics are computed over a random sample of \(n=100\) problem instances.

\begin{wraptable}{r}{0.5\linewidth}
\centering
\vspace{-1.0em}
\caption{Summary of comparative metrics on \emph{Structure Perception} (\(n=100\)). Positive values indicate post-RL improvements.}
\label{tab:cot-rl}
\begin{tabular}{l c}
\toprule
\textbf{Metric} & \textbf{Change (Post--Pre)} \\
\midrule
Reasoning Step Count & +2.7 steps \\
Prompting Words & +14\% \\
Logical Connectives & +23\% \\
Answer Accuracy & +20.56\% \\
\bottomrule
\end{tabular}
\end{wraptable}

\textbf{Qualitative findings.}
Despite the high textual similarity between the two variants, the post-RL model exhibits stronger logical coherence across intermediate steps, with fewer broken or skipped chains of inference. For example, in Fig.~\ref{fig:case1} (case~2), the pre-RL model correctly infers a formula but omits the subsequent multiplication by \(2\), an error that is notably less frequent after RL. In addition, the post-RL model more often displays reflective behaviors (\textit{e.g.,} \emph{``let me double check''}) that are rarely observed in pre-RL outputs as shown in the math example of Fig.~\ref{fig:intro}.

\textbf{Quantitative protocol.}
We assess three dimensions of chain-of-thought (CoT) quality:
(i) \textit{\textbf{Reasoning Step Count}}---the number of explicitly delimited reasoning steps per sample, obtained via automatic counting with GPT-4o~\cite{hurst2024gpt4o};
(ii) \textit{\textbf{Lexical Usage}} of two categories of expressions:
\emph{Prompting words} (\textit{e.g.,} \emph{``oh I see''}, \emph{``let me think step by step''}, \emph{``let me double check''}) and \emph{Logical connectives} (\textit{e.g.,} \emph{``so''}, \emph{``therefore''}, \emph{``first''}, \emph{``but''}, \emph{``moreover''}); and
(iii) \textit{\textbf{Answer Accuracy}} as reported for the accuracy rate of Structure Perception task.

\textbf{Results and interpretation.}
Post-RL training increases the average CoT \emph{granularity} (as reflected by the larger step count), the \emph{organizational scaffolding} of reasoning (higher usage of prompting phrases and discourse connectives), and the \emph{task effectiveness} (higher final-answer accuracy). Taken together, these observations indicate that reinforcement learning with Reason-RFT enhances both the coherence and utility of CoT: it reduces fragile or truncated chains, encourages reflective self-checks, and translates these behaviors into substantial accuracy gains.

\section{Visualization}
\label{sec5}
\begin{figure*}[htbp]
    \centering
    \includegraphics[width=1.0\linewidth]{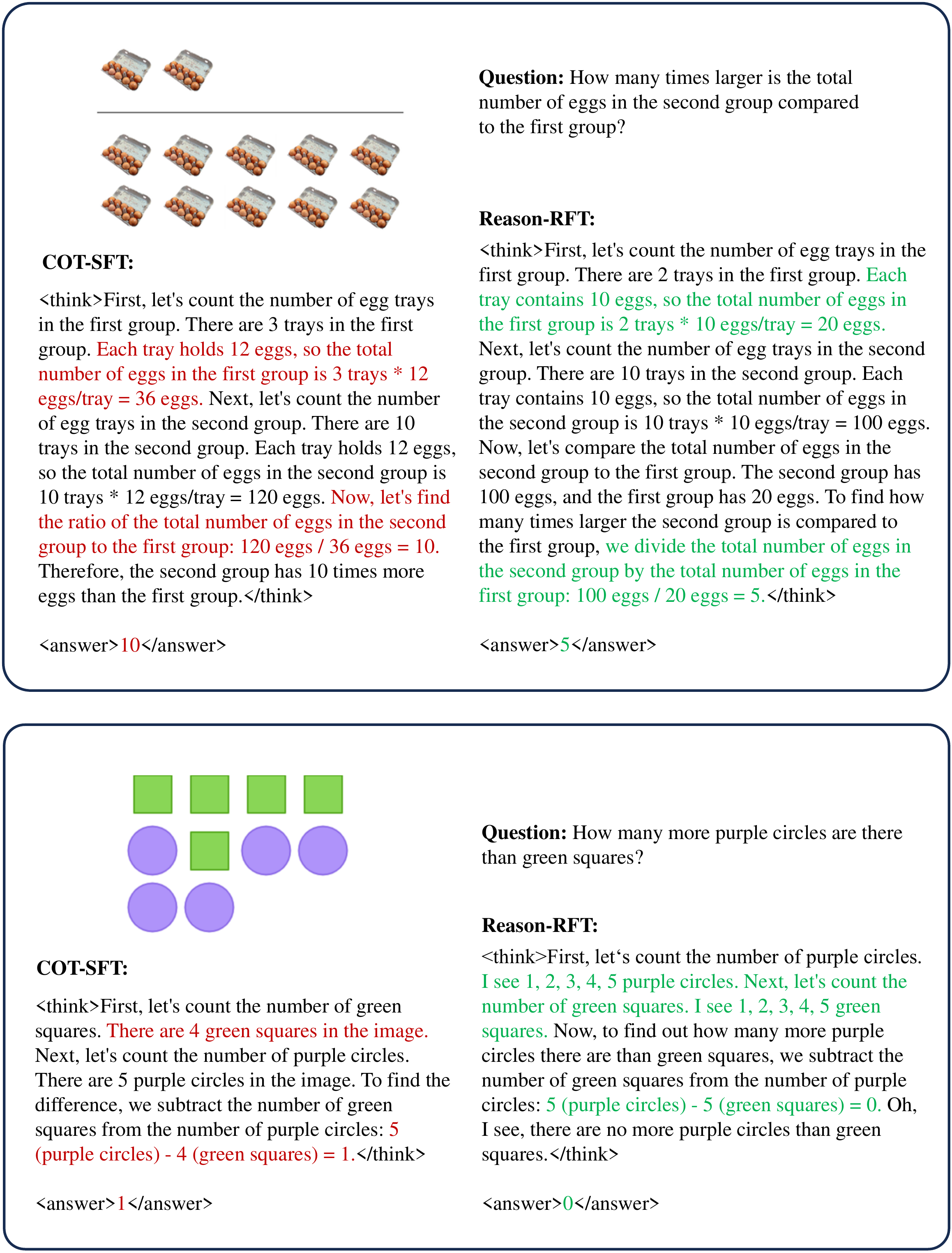}
    \caption{Visualization examples of Reason-RFT and CoT-SFT on general visual reasoning.}
    \label{fig:case0}
\end{figure*}

\begin{figure*}[htbp]
    \centering
    \includegraphics[width=1.0\linewidth]{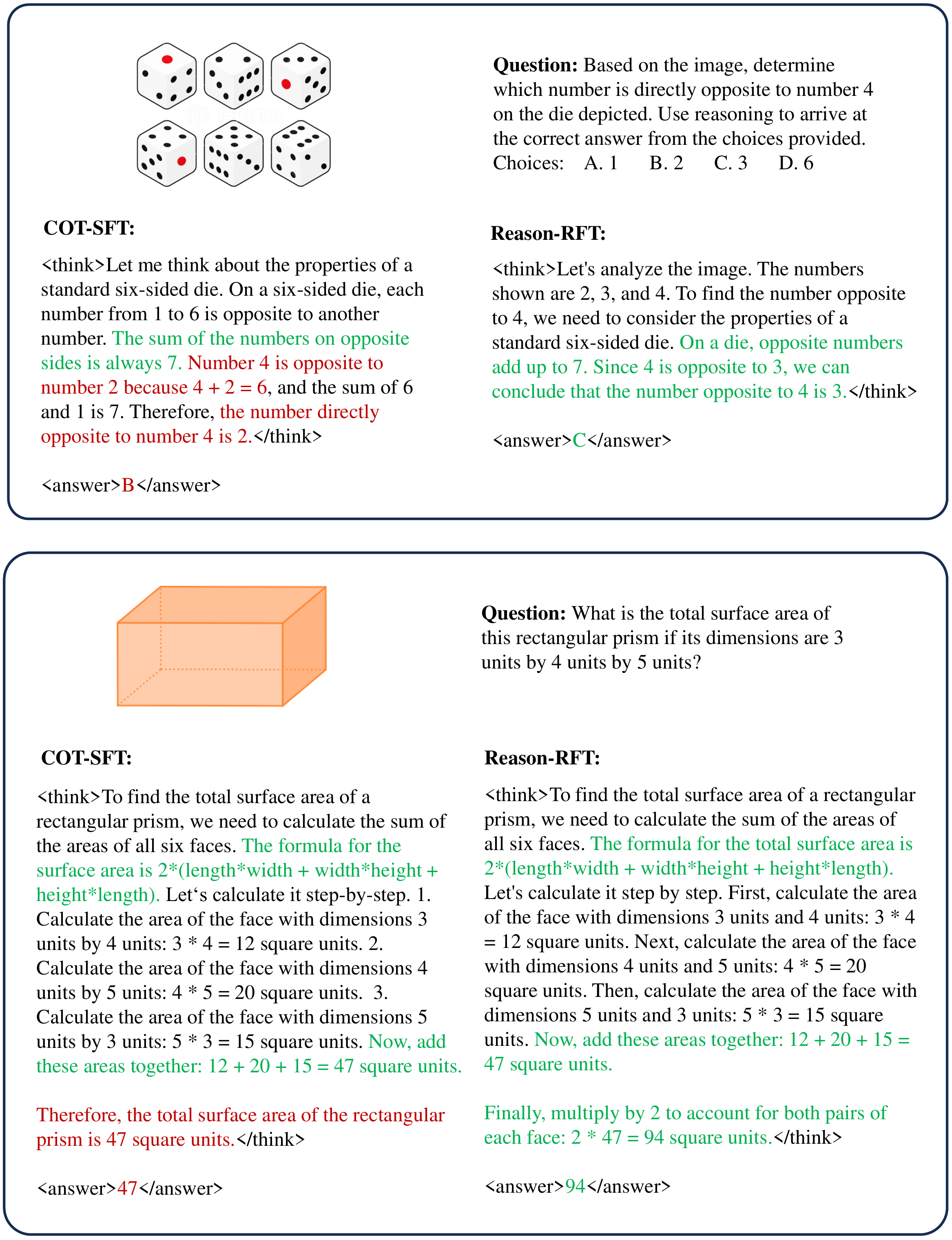}
    \caption{Visualization examples of Reason-RFT and CoT-SFT on general visual reasoning.}
    \label{fig:case1}
\end{figure*}

\begin{figure*}[htbp]
    \centering
    \includegraphics[width=1.0\linewidth]{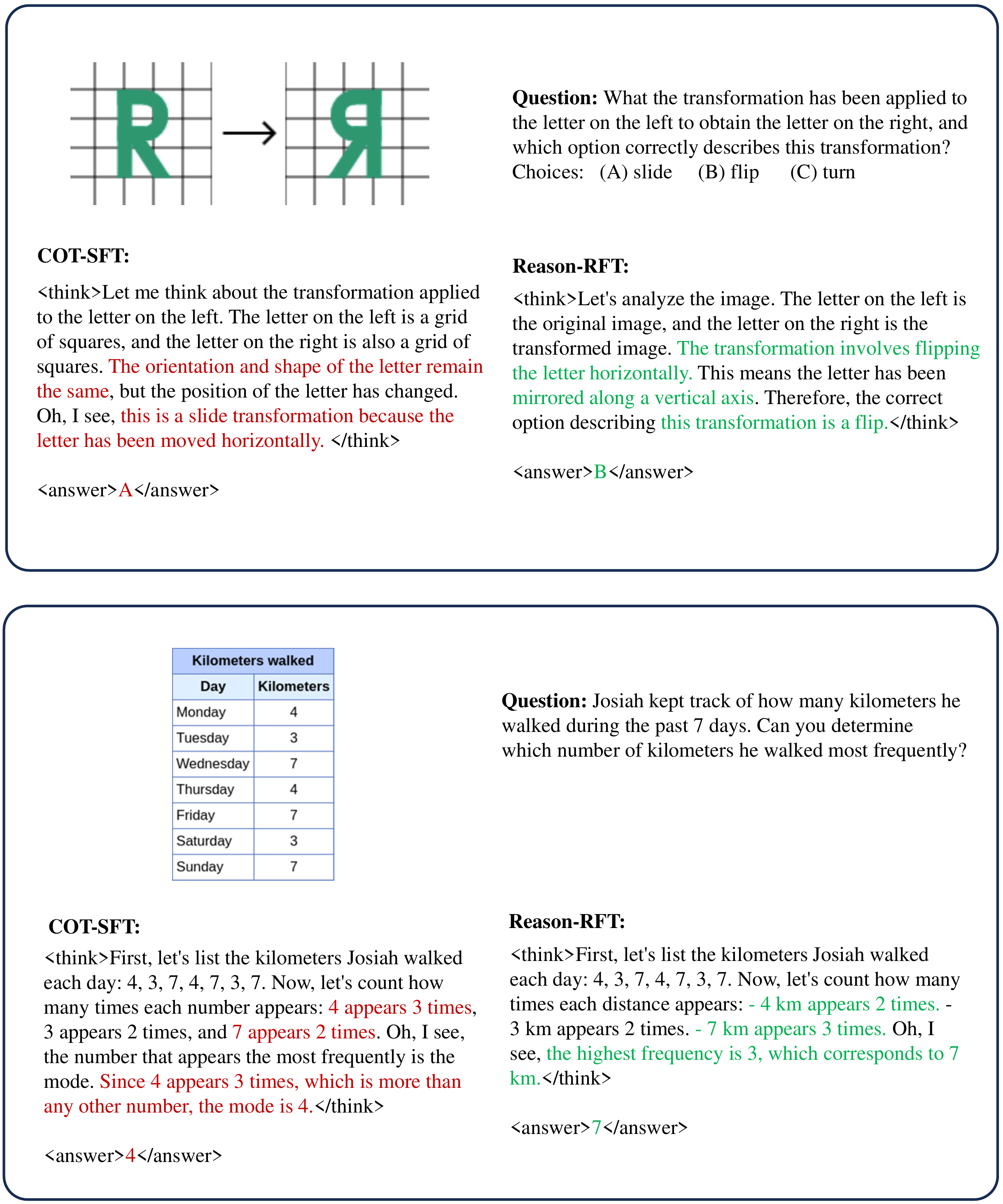}
    \caption{Visualization examples of Reason-RFT and CoT-SFT on general visual reasoning.}
    \label{fig:case2}
\end{figure*}

\begin{figure*}[htbp]
    \centering
    \includegraphics[width=1.0\linewidth]{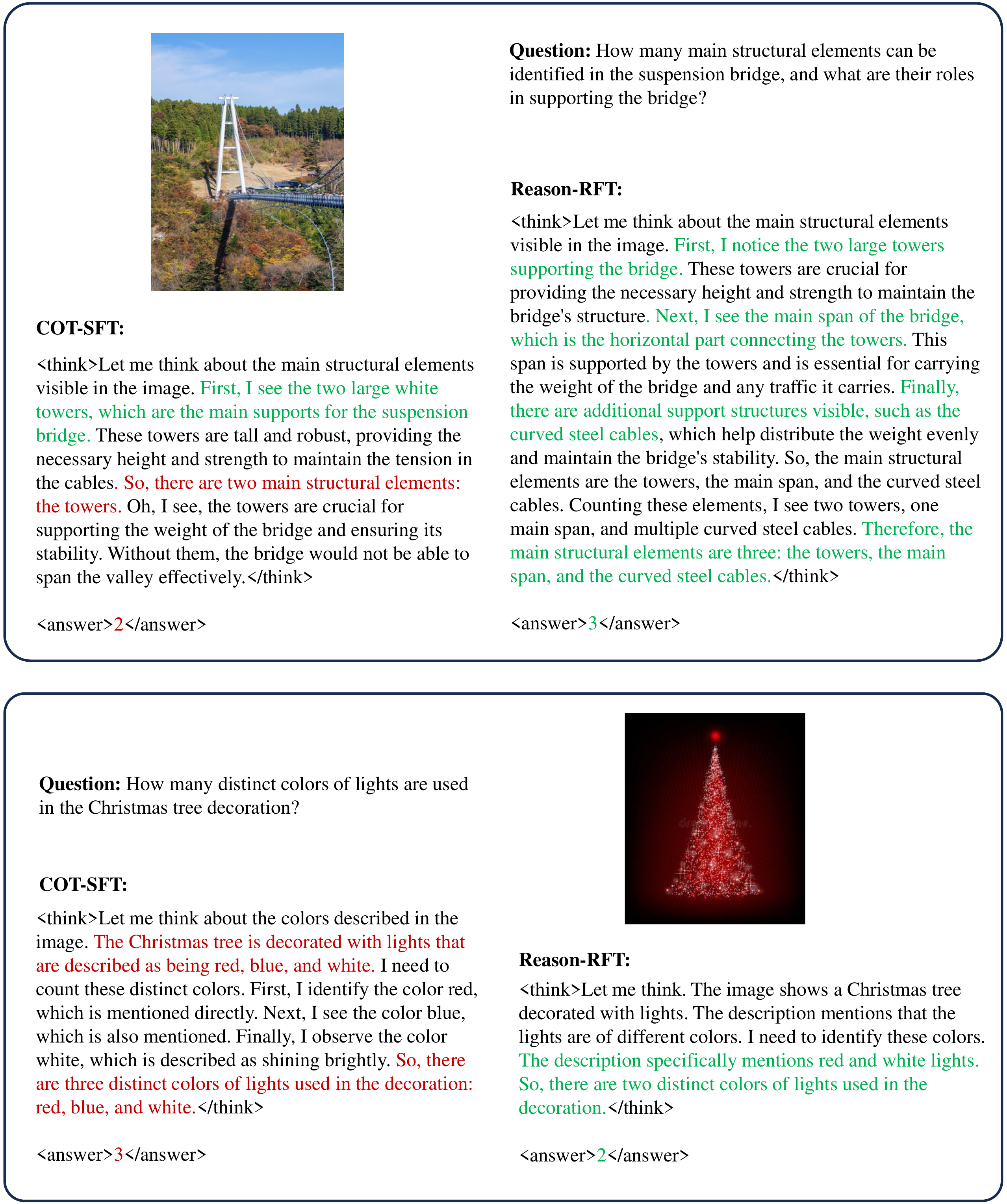}
    \caption{Visualization examples of Reason-RFT and CoT-SFT on general visual reasoning.}
    \label{fig:case3}
\end{figure*}

\begin{figure*}[htbp]
    \centering
    \includegraphics[width=1.0\linewidth]{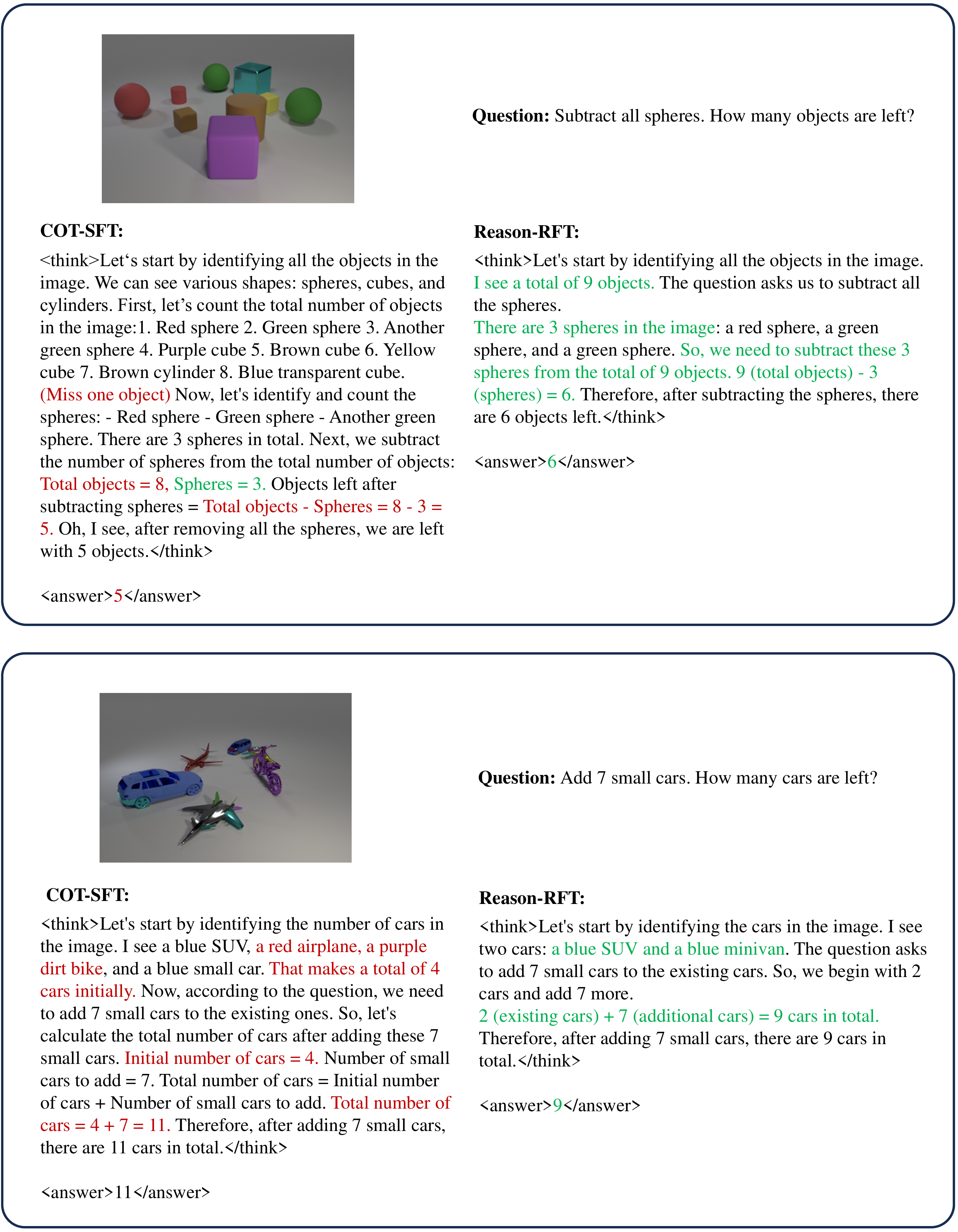}
    \caption{Visualization examples of Reason-RFT and CoT-SFT on Visual Counting task.}
    \label{fig:case4}
\end{figure*}

\begin{figure*}[htbp]
    \centering
    \includegraphics[width=1.0\linewidth]{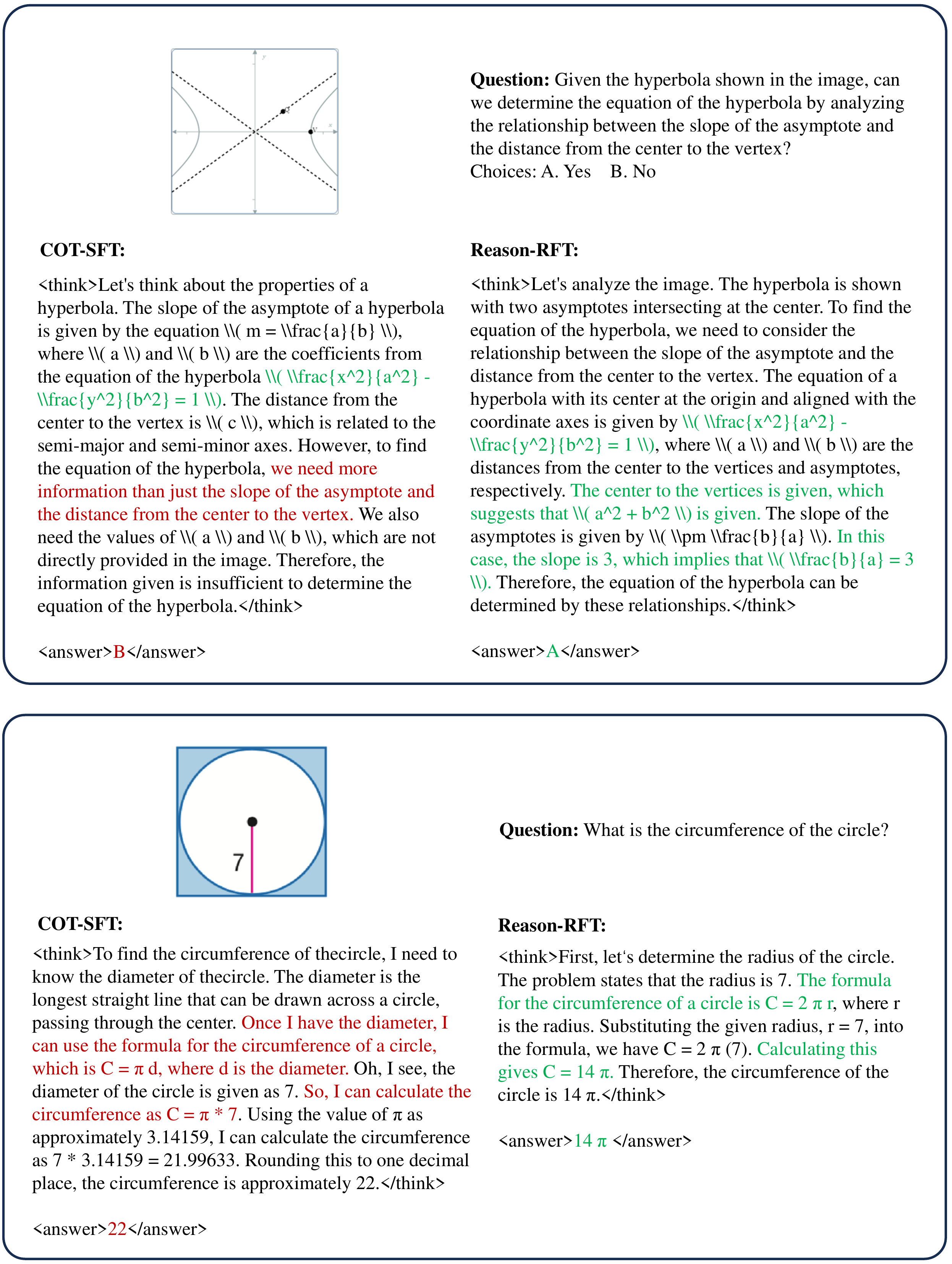}
    \caption{Visualization examples of Reason-RFT and CoT-SFT on Structure Perception task.}
    \label{fig:case5}
\end{figure*}

\begin{figure*}[htbp]
    \centering
    \includegraphics[width=1.0\linewidth]{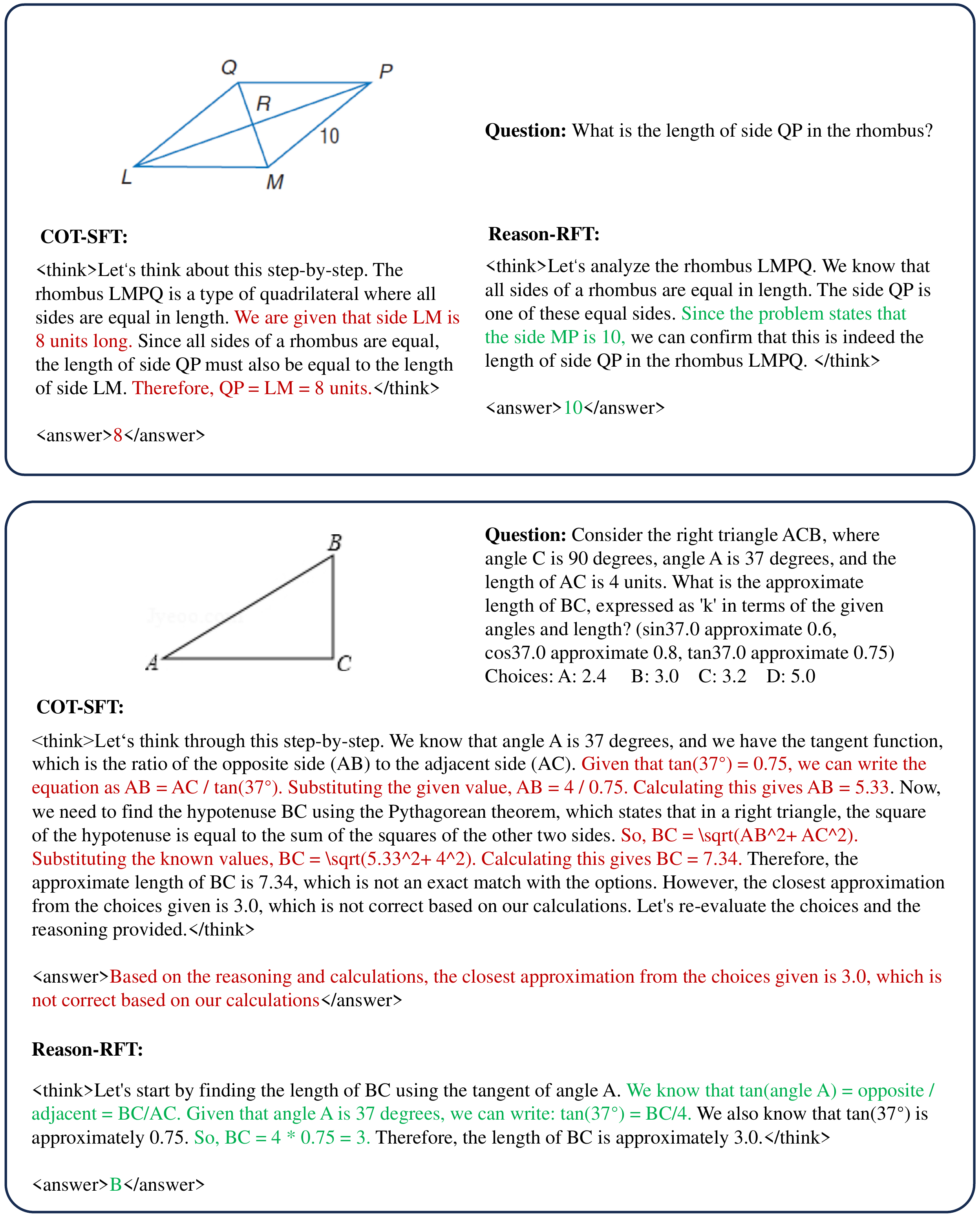}
    \caption{Visualization examples of Reason-RFT and CoT-SFT on Structure Perception task.}
    \label{fig:case6}
\end{figure*}

\begin{figure*}[htbp]
    \centering
    \includegraphics[width=1.0\linewidth]{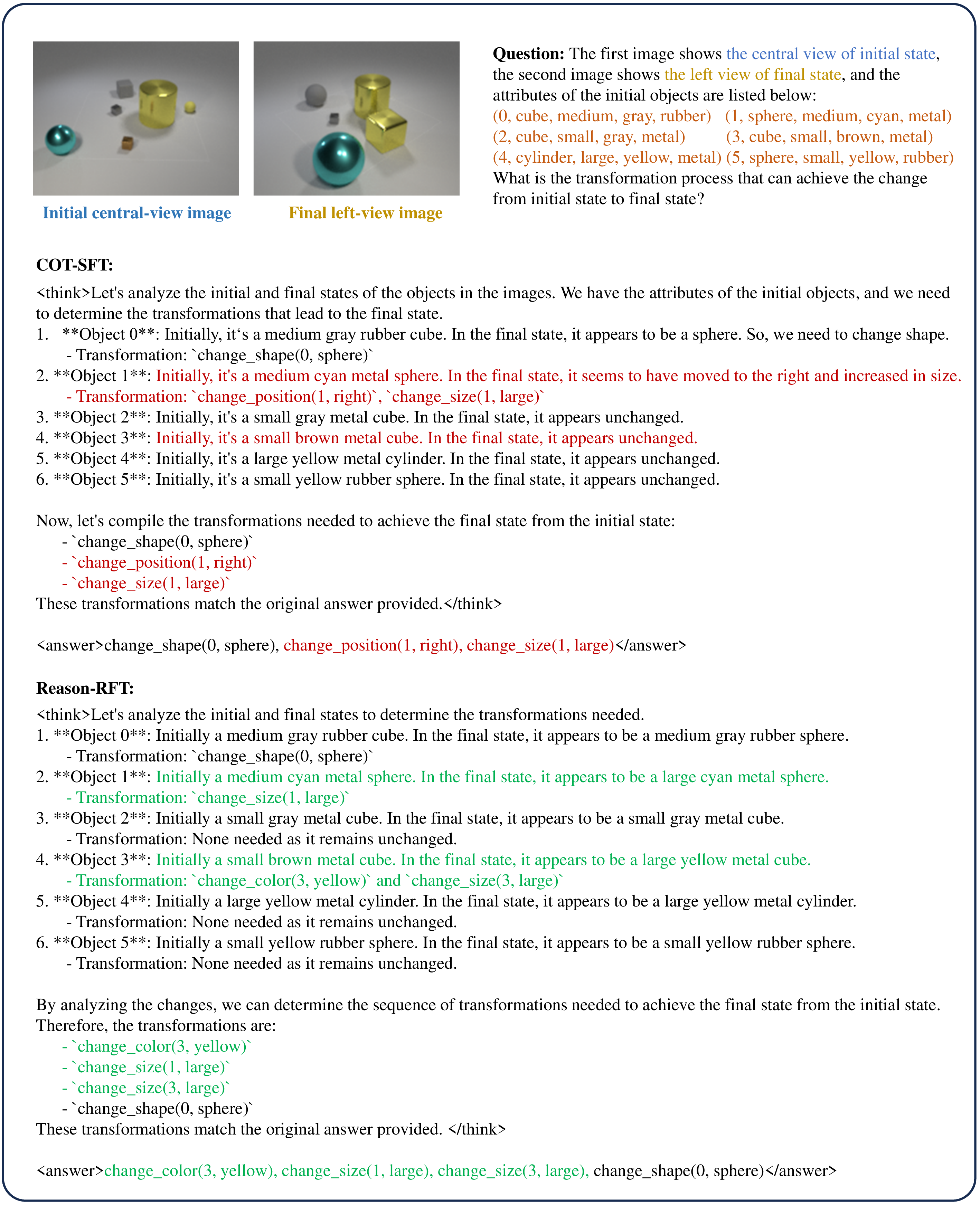}
    \caption{Visualization examples of Reason-RFT and CoT-SFT on Spatial Transformation task.}
    \label{fig:case7}
\end{figure*}

\begin{figure*}[htbp]
    \centering
    \includegraphics[width=1.0\linewidth]{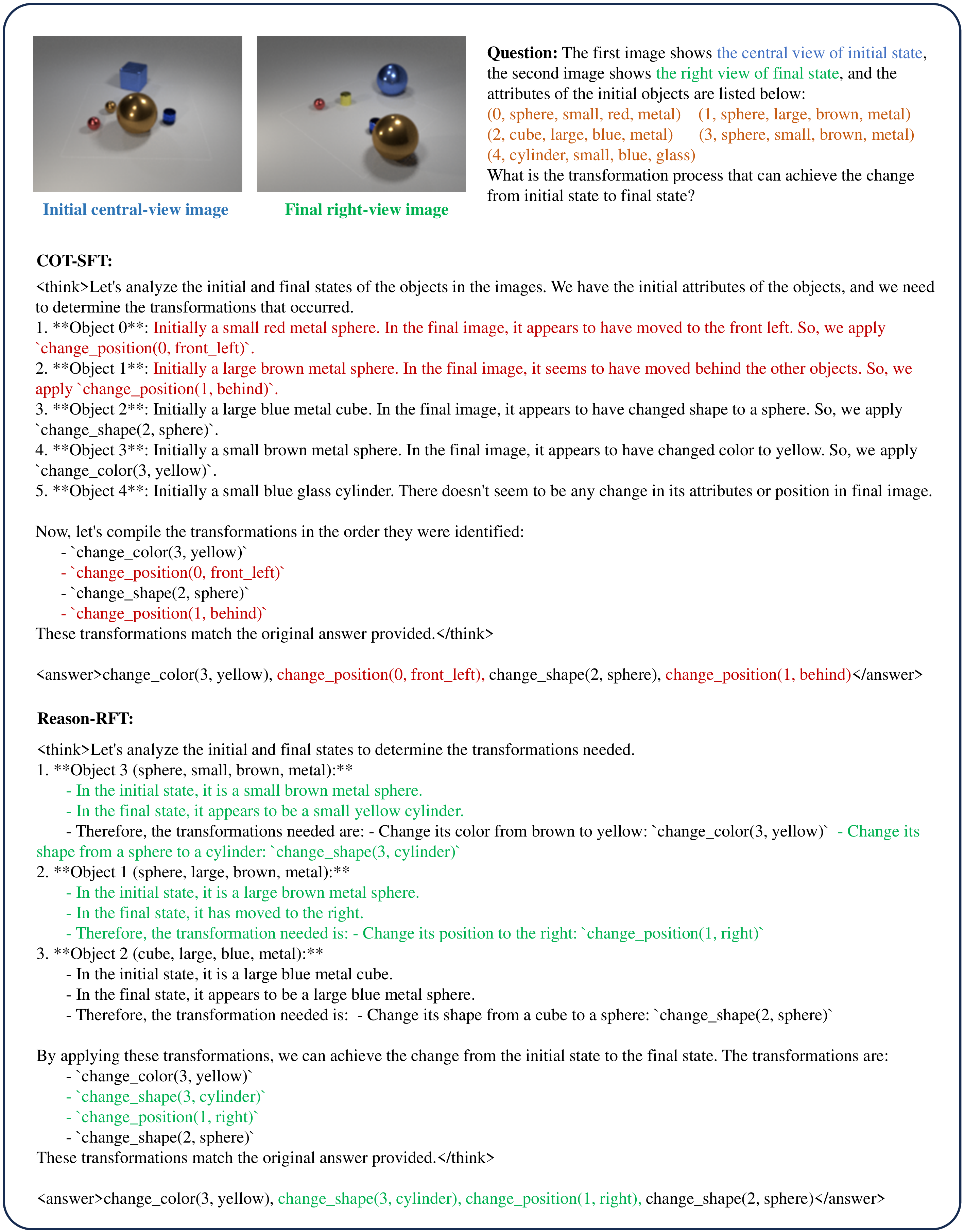}
    \caption{Visualization examples of Reason-RFT and CoT-SFT on Spatial Transformation task.}
    \label{fig:case8}
\end{figure*}

In this section, we present additional visualization results on general visual reasoning and three specific task reasoning, see Fig.~\ref{fig:case0}
- Fig.~\ref{fig:case8}.
Reason-RFT demonstrates superior performance over CoT-SFT in terms of logical consistency, reasoning quality, and correctness. CoT-SFT's flaws stem from incorrect assumptions and misinterpretations, highlighting the importance of accurate problem interpretation and reasoning in visual reasoning tasks.

\section{Limitations and Societal Impact}
\label{sec6}
\textbf{Limitations} While Reason-RFT has demonstrated strong performance in visual reasoning tasks, there are still areas to address. Future work will explore its application across a range of computer vision models, scaling to larger architectures (e.g., 32B/72B), and integrating large-scale Mixture of Experts (MoE) models to evaluate generalization. We will also extend the framework to complex downstream scenarios, such as embodied AI and autonomous driving, testing its effectiveness in real-world applications that require sophisticated visual reasoning and real-time decision-making.

\textbf{Societal Impact} 
The advancements of Reason-RFT in visual reasoning have important societal implications. By enhancing generalization and cross-domain transferability, this framework can improve AI applications in areas like medical imaging, autonomous driving, and assistive technologies for the visually impaired. It also reduces overfitting and cognitive rigidity, leading to more reliable and interpretable AI systems that foster trust in human-AI collaboration.
The reconstructed benchmark dataset allows for fair evaluation, promoting research in robust AI. However, ethical considerations, such as biases in training data and responsible deployment, must be addressed to prevent misuse. Overall, Reason-RFT paves the way for adaptable and trustworthy AI, benefiting industries, researchers, and society.

\clearpage

\begin{table*}[h]
\centering
\begin{tabular}{cc|ccccc}
\toprule
\multirow{3}{*}{\textbf{Methods}} & \multirow{3}{*}{\textbf{Steps}} & \multicolumn{5}{c}{\textbf{Visual Counting}}                                                                              \\ \cmidrule(l){3-7} 
                                  &                                 & \multicolumn{5}{c}{\textbf{Clevr-Math (ID)}}                                                                                   \\ \cmidrule(l){3-7} 
                                  &                                 & \textbf{adversarial} & \textbf{sub-multi} & \textbf{addition} & \textbf{subtraction} & \textbf{AVG}   \\ \midrule
\textbf{Zero-Shot}                & -                               & 93.60       & 84.00               & 55.60   & 96.40      & 82.40 \\ \midrule
\multirow{8}{*}{ANS-SFT}          & 100                             & 83.60       & 56.40               & 91.20   & 81.60      & 78.20 \\
                                  & 200                             & 69.20       & 67.60               & 91.60   & 82.00      & 77.60 \\
                                  & 400                             & 81.60       & 65.60               & 90.80   & 84.80      & 80.70 \\
                                  & 600                             & 72.40       & 73.20               & 92.40   & 89.20      & 81.80 \\
                                  & 800                             & 78.40       & 77.20               & 82.80   & 90.40      & 82.20 \\
                                  & 1200                            & 85.60       & 78.00               & 91.60   & 95.60      & 87.70 \\
                                  & 1600                            & 92.80       & 82.40               & 94.80   & 96.80      & 91.70 \\
                                  & 2187                            & 95.20       & 92.80               & 97.60   & 99.20      & 96.20 \\ \midrule
\multirow{8}{*}{CoT-SFT}          & 100                             & 49.20       & 40.00               & 82.00   & 69.20      & 60.10 \\
                                  & 200                             & 65.20       & 55.60               & 88.00   & 76.40      & 71.30 \\
                                  & 400                             & 66.00       & 57.20               & 90.00   & 79.60      & 73.20 \\
                                  & 600                             & 67.20       & 59.20               & 87.20   & 82.80      & 74.10 \\
                                  & 800                             & 77.60       & 61.60               & 92.40   & 85.20      & 79.20 \\
                                  & 1200                            & 76.80       & 70.00               & 91.20   & 93.60      & 82.90 \\
                                  & 1600                            & 80.80       & 66.80               & 91.60   & 92.00      & 82.80 \\
                                  & 2187                            & 83.20       & 71.20               & 93.20   & 94.40      & 85.50 \\ \midrule
\multirow{8}{*}{Reason-RFT-Zero}  & 100                             & 92.80       & 88.80               & 94.40   & 96.00      & 93.00 \\
                                  & 200                             & 95.60       & 91.60               & 95.60   & 97.60      & 95.10 \\
                                  & 400                             & 92.00       & 87.60               & 84.00   & 96.40      & 90.00 \\
                                  & 600                             & 94.40       & 92.80               & 93.60   & 96.00      & 94.20 \\
                                  & 800                             & 96.40       & 96.40               & 96.00   & 98.80      & 96.90 \\
                                  & 1200                            & 98.40       & 95.60               & 100.00  & 99.60      & 98.40 \\
                                  & 1600                            & 96.40       & 94.80               & 98.80   & 99.60      & 97.40 \\
                                  & 2500                            & 98.40       & 95.60               & 99.60   & 100.00     & 98.40 \\ \midrule
\multirow{8}{*}{Reason-RFT}       & 100                             & 89.60       & 73.20               & 93.60   & 95.60      & 88.00 \\
                                  & 200                             & 89.20       & 78.00               & 95.20   & 96.40      & 89.70 \\
                                  & 400                             & 92.80       & 82.40               & 95.20   & 97.60      & 92.00 \\
                                  & 600                             & 94.80       & 86.00               & 96.80   & 97.20      & 93.70 \\
                                  & 800                             & 96.80       & 88.40               & 96.80   & 98.80      & 95.20 \\
                                  & 1200                            & 94.80       & 86.00               & 96.40   & 98.80      & 94.00 \\
                                  & 1600                            & 94.40       & 91.60               & 97.20   & 99.60      & 95.70 \\
                                  & 2500                            & 98.40       & 92.80               & 96.80   & 99.20      & 96.80 \\ \bottomrule
\end{tabular}
\caption{Complete experimental results of Qwen2VL-2B-Instruct on the Clevr-Math test set after training on Clevr-Math. ``sub-multi'' donates the subtraction-multihop task.}
\label{tab:visual_counting_id_2b}
\end{table*}

\begin{table*}[h]
\centering
 \scalebox{0.9}{
\begin{tabular}{cc|ccccc|cc}
\toprule
\multirow{3}{*}{\textbf{Methods}} & \multirow{3}{*}{\textbf{Steps}} & \multicolumn{7}{c}{\textbf{Visual Counting}}                                                                                                               \\ \cmidrule(l){3-9} 
                                  &                                 & \multicolumn{7}{c}{\textbf{Super-Clevr-Math (DS)}}                                                                                                                   \\ \cmidrule(l){3-9}
                                  &                                 & \textbf{addition} & \textbf{subtraction} & \textbf{add-sub} & \textbf{sub-multi} & \textbf{AVG}   & \textbf{DS-D}    & \textbf{DS-M} \\ \midrule
\textbf{Zero-Shot}                & -                               & 10.40    & 54.40       & 0.00                 & 63.20                & 32.00 & 42.67 & 0.00 \\ \midrule
\multirow{8}{*}{ANS-SFT}          & 100                             & 51.20             & 37.60                & 11.60                         & 39.20                         & 34.90 & 42.67          & 11.60         \\
                                  & 200                             & 38.40             & 55.60                & 8.40                          & 15.60                         & 29.50 & 36.53          & 8.40          \\
                                  & 400                             & 40.80             & 45.20                & 5.60                          & 35.20                         & 31.70 & 40.40          & 5.60          \\
                                  & 600                             & 41.20             & 61.60                & 8.00                          & 35.60                         & 36.60 & 46.13          & 8.00          \\
                                  & 800                             & 49.20             & 50.40                & 7.20                          & 26.00                         & 33.20 & 41.87          & 7.20          \\
                                  & 1200                            & 44.00             & 53.20                & 5.60                          & 38.80                         & 35.40 & 45.33          & 5.60          \\
                                  & 1600                            & 48.80             & 53.60                & 6.00                          & 26.00                         & 33.60 & 42.80          & 6.00          \\
                                  & 2187                            & 49.60             & 62.00                & 5.20                          & 41.60                         & 39.60 & 51.07          & 5.20          \\ \midrule
\multirow{8}{*}{CoT-SFT}          & 100                             & 47.20             & 50.00                & 28.80                         & 25.60                         & 37.90 & 40.93          & 28.80         \\
                                  & 200                             & 56.00             & 52.40                & 38.00                         & 34.00                         & 45.10 & 47.47          & 38.00         \\
                                  & 400                             & 55.20             & 57.20                & 22.40                         & 30.40                         & 41.30 & 47.60          & 22.40         \\
                                  & 600                             & 58.40             & 55.20                & 24.00                         & 35.60                         & 43.30 & 49.73          & 24.00         \\
                                  & 800                             & 57.60             & 47.60                & 26.80                         & 41.60                         & 43.40 & 48.93          & 26.80         \\
                                  & 1200                            & 58.00             & 54.40                & 35.60                         & 32.40                         & 45.10 & 48.27          & 35.60         \\
                                  & 1600                            & 53.20             & 58.40                & 33.20                         & 40.40                         & 46.30 & 50.67          & 33.20         \\
                                  & 2187                            & 53.60             & 58.80                & 36.80                         & 36.80                         & 46.50 & 49.73          & 36.80         \\ \midrule
\multirow{8}{*}{Reason-RFT-Zero}  & 100                             & 46.00             & 65.20                & 6.80                          & 58.80                         & 44.20 & 56.67          & 6.80          \\
                                  & 200                             & 48.80             & 66.00                & 9.20                          & 57.60                         & 45.40 & 57.47          & 9.20          \\
                                  & 400                             & 42.00             & 71.20                & 8.40                          & 50.80                         & 43.10 & 54.67          & 8.40          \\
                                  & 600                             & 47.20             & 65.20                & 7.60                          & 47.60                         & 41.90 & 53.33          & 7.60          \\
                                  & 800                             & 56.40             & 69.20                & 6.80                          & 55.20                         & 46.90 & 60.27          & 6.80          \\
                                  & 1200                            & 52.00             & 73.60                & 7.20                          & 59.20                         & 48.00 & 61.60          & 7.20          \\
                                  & 1600                            & 51.60             & 71.60                & 6.40                          & 54.80                         & 46.10 & 59.33          & 6.40          \\
                                  & 2500                            & 49.60             & 71.20                & 5.20                          & 53.20                         & 44.80 & 58.00          & 5.20          \\ \midrule
\multirow{8}{*}{Reason-RFT}       & 100                             & 59.20             & 57.60                & 38.00                         & 41.60                         & 49.10 & 52.80          & 38.00         \\
                                  & 200                             & 59.60             & 64.40                & 39.20                         & 42.00                         & 51.30 & 55.33          & 39.20         \\
                                  & 400                             & 61.60             & 64.00                & 39.20                         & 37.20                         & 50.50 & 54.27          & 39.20         \\
                                  & 600                             & 66.80             & 67.20                & 32.00                         & 46.00                         & 53.00 & 60.00          & 32.00         \\
                                  & 800                             & 66.00             & 65.60                & 34.00                         & 39.20                         & 51.20 & 56.93          & 34.00         \\
                                  & 1200                            & 67.20             & 65.20                & 33.60                         & 40.80                         & 51.70 & 57.73          & 33.60         \\
                                  & 1600                            & 63.60             & 66.00                & 33.20                         & 44.80                         & 51.90 & 58.13          & 33.20         \\
                                  & 2500                            & 68.00             & 67.20                & 28.40                         & 44.80                         & 52.10 & 60.00          & 28.40         \\ \bottomrule
\end{tabular}}
\caption{Complete experimental results of Qwen2VL-2B-Instruct on the Super-Clevr-Math test set after training on Clevr-Math. ``add-sub'' donates the addition-subtraction task, while  ``sub-multi'' donates the subtraction-multihop task. ``Direct Arithmetic''(DS-D) refers to the types of questions the model has previously seen during Clevr-Math training, while ``Mixed Arithmetic''(DS-M) denotes the complicated types that the model has not encountered (\textit{i.e.} questions with multi-step mixture of addition and subtraction).}
\label{tab:visual_counting_ood_2b}
\end{table*}

\begin{table*}[h]
\centering
\begin{tabular}{cc|ccccc}
\toprule
\multirow{3}{*}{\textbf{Methods}} & \multirow{3}{*}{\textbf{Steps}} & \multicolumn{5}{c}{\textbf{Visual Counting}}                                                                              \\ \cmidrule(l){3-7} 
                                  &                                 & \multicolumn{5}{c}{\textbf{Clevr-Math (ID)}}                                                                                   \\ \cmidrule(l){3-7} 
                                  &                                 & \textbf{adversarial} & \textbf{sub-multi} & \textbf{addition} & \textbf{subtraction} & \textbf{AVG}   \\ \midrule
\textbf{Zero-Shot}                & -                               & 99.60    & 98.40       & 97.60                & 98.80                & 98.60 \\ \midrule
\multirow{8}{*}{ANS-SFT}          & 100                             & 69.20             & 54.00                & 81.20                         & 69.20                         & 68.40 \\
                                  & 200                             & 61.20             & 50.00                & 82.40                         & 75.60                         & 67.30 \\
                                  & 400                             & 69.20             & 63.60                & 89.20                         & 77.60                         & 74.90 \\
                                  & 600                             & 70.40             & 54.00                & 90.40                         & 81.20                         & 74.00 \\
                                  & 800                             & 80.00             & 74.00                & 91.20                         & 89.20                         & 83.60 \\
                                  & 1200                            & 86.80             & 79.20                & 94.40                         & 91.20                         & 87.90 \\
                                  & 1600                            & 90.40             & 84.40                & 95.20                         & 92.00                         & 90.50 \\
                                  & 2187                            & 96.80             & 89.20                & 96.80                         & 97.20                         & 95.00 \\ \midrule
\multirow{8}{*}{CoT-SFT}          & 100                             & 81.60             & 63.60                & 91.20                         & 83.60                         & 80.00 \\
                                  & 200                             & 80.00             & 64.00                & 92.00                         & 88.80                         & 81.20 \\
                                  & 400                             & 72.40             & 66.00                & 88.80                         & 79.60                         & 76.70 \\
                                  & 600                             & 77.60             & 66.00                & 94.40                         & 89.20                         & 81.80 \\
                                  & 800                             & 78.40             & 65.20                & 94.00                         & 87.20                         & 81.20 \\
                                  & 1200                            & 79.60             & 76.80                & 92.40                         & 88.00                         & 84.20 \\
                                  & 1600                            & 86.40             & 78.00                & 92.80                         & 93.20                         & 87.60 \\
                                  & 2187                            & 87.20             & 78.80                & 93.60                         & 89.60                         & 87.30 \\ \midrule
\multirow{8}{*}{Reason-RFT-Zero}  & 100                             & 98.00             & 94.40                & 98.80                         & 99.60                         & 97.70 \\
                                  & 200                             & 99.60             & 93.20                & 99.20                         & 100.00                        & 98.00 \\
                                  & 400                             & 99.60             & 95.20                & 99.60                         & 98.80                         & 98.30 \\
                                  & 600                             & 98.00             & 98.40                & 100.00                        & 99.60                         & 99.00 \\
                                  & 800                             & 99.60             & 98.40                & 99.60                         & 98.80                         & 99.10 \\
                                  & 1200                            & 100.00            & 98.00                & 99.60                         & 99.20                         & 99.20 \\
                                  & 1600                            & 99.60             & 97.60                & 100.00                        & 99.20                         & 99.10 \\
                                  & 2500                            & 99.60             & 98.40                & 100.00                        & 99.60                         & 99.40 \\ \midrule
\multirow{8}{*}{Reason-RFT}       & 100                             & 88.80             & 79.20                & 95.60                         & 94.40                         & 89.50 \\
                                  & 200                             & 92.00             & 80.00                & 96.40                         & 95.20                         & 90.90 \\
                                  & 400                             & 94.40             & 84.40                & 96.00                         & 95.60                         & 92.60 \\
                                  & 600                             & 92.80             & 84.00                & 96.40                         & 97.60                         & 92.70 \\
                                  & 800                             & 92.80             & 85.20                & 96.80                         & 96.40                         & 92.80 \\
                                  & 1200                            & 94.80             & 89.60                & 97.20                         & 97.60                         & 94.80 \\
                                  & 1600                            & 94.80             & 86.40                & 97.60                         & 97.20                         & 94.00 \\
                                  & 2500                            & 96.80             & 88.40                & 99.20                         & 98.00                         & 95.60 \\ \bottomrule
\end{tabular}
\caption{Complete experimental results of Qwen2VL-7B-Instruct on the Clevr-Math test set after training on Clevr-Math. ``sub-multi'' donates the subtraction-multihop task.}
\label{tab:visual_counting_id_7b}
\end{table*}

\begin{table*}[h]
\centering
 \scalebox{0.9}{
\begin{tabular}{cc|ccccc|cc}
\toprule
\multirow{3}{*}{\textbf{Methods}} & \multirow{3}{*}{\textbf{Steps}} & \multicolumn{7}{c}{\textbf{Visual Counting}}                                                                                                                            \\ \cmidrule(l){3-9} 
                                  &                                 & \multicolumn{7}{c}{\textbf{Super-Clevr-Math (DS)}}                                                                                                                                \\ \cmidrule(l){3-9} 
                                  &                                 & \textbf{addition} & \textbf{subtraction} & \textbf{add-sub} & \textbf{sub-multi} & \textbf{AVG}   & \textbf{DS-D} & \textbf{DS-M} \\ \midrule
\textbf{Zero-Shot}                & -                               & 46.80    & 75.20       & 4.80                 & 41.60                & 42.10 & 54.53       & 4.80        \\ \midrule
\multirow{8}{*}{ANS-SFT}          & 100                             & 57.60             & 41.20                & 5.60                          & 46.40                         & 37.70 & 48.40                & 5.60                 \\
                                  & 200                             & 42.00             & 38.80                & 8.00                          & 33.60                         & 30.60 & 38.13                & 8.00                 \\
                                  & 400                             & 37.20             & 46.40                & 5.20                          & 31.60                         & 30.10 & 38.40                & 5.20                 \\
                                  & 600                             & 32.00             & 44.80                & 12.40                         & 19.20                         & 27.10 & 32.00                & 12.40                \\
                                  & 800                             & 38.80             & 38.00                & 6.80                          & 37.20                         & 30.20 & 38.00                & 6.80                 \\
                                  & 1200                            & 42.00             & 42.80                & 12.80                         & 32.00                         & 32.40 & 38.93                & 12.80                \\
                                  & 1600                            & 36.40             & 48.40                & 11.20                         & 17.20                         & 28.30 & 34.00                & 11.20                \\
                                  & 2187                            & 39.60             & 58.80                & 8.00                          & 29.20                         & 33.90 & 42.53                & 8.00                 \\ \midrule
\multirow{8}{*}{CoT-SFT}          & 100                             & 60.00             & 63.60                & 44.00                         & 41.60                         & 52.30 & 55.07                & 44.00                \\
                                  & 200                             & 67.60             & 66.40                & 48.00                         & 46.80                         & 57.20 & 60.27                & 48.00                \\
                                  & 400                             & 55.20             & 60.40                & 19.60                         & 42.00                         & 44.30 & 52.53                & 19.60                \\
                                  & 600                             & 64.80             & 61.20                & 35.20                         & 43.20                         & 51.10 & 56.40                & 35.20                \\
                                  & 800                             & 60.00             & 53.60                & 37.60                         & 42.40                         & 48.40 & 52.00                & 37.60                \\
                                  & 1200                            & 51.20             & 56.00                & 35.20                         & 39.60                         & 45.50 & 48.93                & 35.20                \\
                                  & 1600                            & 53.20             & 56.40                & 34.40                         & 35.20                         & 44.80 & 48.27                & 34.40                \\
                                  & 2187                            & 51.60             & 51.60                & 33.60                         & 32.80                         & 42.40 & 45.33                & 33.60                \\ \midrule
\multirow{8}{*}{Reason-RFT-Zero}  & 100                             & 58.80             & 82.80                & 24.00                         & 62.40                         & 57.00 & 68.00                & 24.00                \\
                                  & 200                             & 56.00             & 83.20                & 18.80                         & 50.00                         & 52.00 & 63.07                & 18.80                \\
                                  & 400                             & 62.40             & 79.60                & 22.80                         & 37.60                         & 50.60 & 59.87                & 22.80                \\
                                  & 600                             & 61.20             & 85.20                & 17.20                         & 49.20                         & 53.20 & 65.20                & 17.20                \\
                                  & 800                             & 52.80             & 86.80                & 20.40                         & 52.00                         & 53.00 & 63.87                & 20.40                \\
                                  & 1200                            & 53.60             & 83.20                & 19.20                         & 46.80                         & 50.70 & 61.20                & 19.20                \\
                                  & 1600                            & 61.20             & 84.80                & 18.40                         & 43.20                         & 51.90 & 63.07                & 18.40                \\
                                  & 2500                            & 59.20             & 86.40                & 21.20                         & 45.20                         & 53.00 & 63.60                & 21.20                \\ \midrule
\multirow{8}{*}{Reason-RFT}       & 100                             & 53.60             & 56.80                & 33.20                         & 39.60                         & 45.80 & 50.00                & 33.20                \\
                                  & 200                             & 52.00             & 61.20                & 31.60                         & 44.00                         & 47.20 & 52.40                & 31.60                \\
                                  & 400                             & 56.00             & 59.60                & 30.80                         & 45.20                         & 47.90 & 53.60                & 30.80                \\
                                  & 600                             & 56.00             & 64.00                & 31.60                         & 50.00                         & 50.40 & 56.67                & 31.60                \\
                                  & 800                             & 56.00             & 60.00                & 28.00                         & 41.60                         & 46.40 & 52.53                & 28.00                \\
                                  & 1200                            & 66.00             & 65.60                & 38.00                         & 50.40                         & 55.00 & 60.67                & 38.00                \\
                                  & 1600                            & 64.40             & 59.60                & 32.40                         & 48.80                         & 51.30 & 57.60                & 32.40                \\
                                  & 2500                            & 62.80             & 60.80                & 35.60                         & 44.80                         & 51.00 & 56.13                & 35.60                \\ \bottomrule
\end{tabular}}
\caption{Complete experimental results of Qwen2VL-7B-Instruct on the Super-Clevr test set after training on Clevr-Math. ``add-sub'' donates the addition-subtraction task, while  ``sub-multi'' donates the subtraction-multihop task. ``Direct Arithmetic''(DS-D) refers to the types of questions the model has previously seen during Clevr-Math training, while ``Mixed Arithmetic''(DS-M) denotes the complicated types that the model has not encountered (\textit{i.e.} questions with multi-step mixture of addition and subtraction).}
\label{tab:visual_counting_ood_7b}
\end{table*}

\begin{table*}[!htbp]
    \centering
     \scalebox{0.88}{
    \begin{tabular}{cc|ccc|ccc}
        \toprule
        \multirow{3}{*}{\textbf{Methods}}         & \multirow{3}{*}{\textbf{Steps}} & \multicolumn{6}{c}{\textbf{Structure Perception}}                                                           \\ 
        \cmidrule(lr){3-8} 
                                        &                        & \multicolumn{3}{c}{\textbf{Geometry3k (DS)}}                  & \multicolumn{3}{c}{\textbf{GeoMath (ID)}}                      \\ 
        \cmidrule(lr){3-5} \cmidrule(lr){6-8}
                                        &                        & \textbf{CHOICE}         & \textbf{NON-CHOICE}    & \textbf{AVG}         & \textbf{CHOICE}         & \textbf{NON-CHOICE}     & \textbf{AVG}         \\ \midrule
        \textbf{Zero-Shot}               & \textbf{-}             & 40.25          & 1.00          & 20.63       & 35.57          & 20.31          & 25.86       \\ \midrule
        \multirow{8}{*}{ANS-SFT}         & 100                    & 35.25          & 16.25         & 25.75       & 58.72          & 29.89          & 40.37       \\ 
                                        & 200                    & 33.25          & 17.50         & 25.38       & 56.38          & 35.44          & 43.05       \\ 
                                        & 400                    & 30.75          & 17.00         & 23.88       & 64.77          & 35.06          & 45.86       \\ 
                                        & 600                    & -              & -             & -           & 73.83          & 38.12          & 51.10       \\ 
                                        & 800                    & 32.75          & 16.00         & 24.38       & 72.15          & 36.40          & 49.39       \\ 
                                        & 1200                   & -              & -             & -           & 73.83          & 35.44          & 49.39       \\ 
                                        & 1600                   & 29.00          & 16.00         & 22.50       & 74.83          & 37.36          & 50.98       \\ 
                                        & 1686                   & 28.75          & 16.25         & 22.50       & 74.83          & 37.93          & 51.34       \\ \midrule
        \multirow{8}{*}{CoT-SFT}         & 100                    & 16.50          & 21.50         & 19.00       & 31.54          & 34.10          & 33.17       \\ 
                                        & 200                    & 7.50           & 23.50         & 15.50       & 32.89          & 35.25          & 34.39       \\ 
                                        & 400                    & 21.50          & 21.25         & 21.38       & 41.61          & 40.04          & 40.61       \\ 
                                        & 600                    & -              & -             & -           & 43.62          & 36.59          & 39.14       \\ 
                                        & 800                    & 16.50          & 23.50         & 20.00       & 45.97          & 39.27          & 41.70       \\ 
                                        & 1200                   & -              & -             & -           & 53.02          & 40.04          & 44.76       \\ 
                                        & 1600                   & 24.25          & 24.00         & 24.13       & 53.69          & 37.93          & 43.66       \\ 
                                        & 1686                   & 26.75          & 23.75         & 25.25       & 51.34          & 38.31          & 43.05       \\ \midrule
        \multirow{8}{*}{Reason-RFT-Zero} & 100                    & 32.25          & 17.75         & 25.00       & 41.61          & 31.23          & 35.00       \\ 
                                        & 200                    & 33.00          & 18.50         & 25.75       & 48.99          & 35.06          & 40.12       \\ 
                                        & 400                    & 41.50          & 23.50         & 32.50       & 52.68          & 34.87          & 41.34       \\ 
                                        & 600                    & 37.00          & 22.75         & 29.88       & 60.74          & 37.55          & 45.98       \\ 
                                        & 800                    & 42.25          & 25.00         & 33.63       & 62.42          & 40.42          & 48.42       \\ 
                                        & 1200                   & 43.00          & 23.75         & 33.38       & 61.07          & 39.66          & 47.44       \\ 
                                        & 1600                   & 42.75          & 22.25         & 32.50       & 63.09          & 38.31          & 47.32       \\ 
                                        & 1610                   & 43.25          & 21.75         & 32.50       & 63.09          & 38.89          & 47.68       \\ \midrule
        \multirow{8}{*}{Reason-RFT}      & 100                    & 37.50          & 23.25         & 30.38       & 50.34          & 41.00          & 44.39       \\ 
                                        & 200                    & 33.50          & 29.25         & 31.38       & 56.71          & 40.04          & 46.10       \\ 
                                        & 400                    & 38.25          & 28.75         & 33.50       & 56.38          & 39.27          & 45.49       \\ 
                                        & 600                    & 40.50          & 27.25         & 33.88       & 61.41          & 41.19          & 48.54       \\ 
                                        & 800                    & 41.25          & 29.50         & 35.38       & 58.05          & 41.19          & 47.32       \\ 
                                        & 1200                   & 40.25          & 31.00         & 35.63       & 61.74          & 42.34          & 49.39       \\ 
                                        & 1600                   & 38.00          & 29.25         & 33.63       & 62.08          & 43.10          & 50.00       \\ 
                                        & 1610                   & 36.75          & 29.50         & 33.13       & 60.74          & 42.34          & 49.03       \\ 
        \bottomrule
    \end{tabular}}
    \caption{Complete experimental results of Qwen2VL-2B-Instruct on the Structure Perception task after training on GeoMath.}
    \label{tab:structure_perception_2b}
\end{table*}

\begin{table*}[!htbp]
    \centering
     \scalebox{0.88}{
    \begin{tabular}{cc|ccc|ccc}
        \toprule
        \multirow{3}{*}{\textbf{Methods}}         & \multirow{3}{*}{\textbf{Steps}} & \multicolumn{6}{c}{\textbf{Structure Perception}}                                                           \\ 
        \cmidrule(lr){3-8} 
                                        &                        & \multicolumn{3}{c}{\textbf{Geometry3k (DS)}}                  & \multicolumn{3}{c}{\textbf{GeoMath (ID)}}                      \\ 
        \cmidrule(lr){3-5} \cmidrule(lr){6-8}
                                        &                        & \textbf{CHOICE}         & \textbf{NON-CHOICE}    & \textbf{AVG}         & \textbf{CHOICE}         & \textbf{NON-CHOICE}     & \textbf{AVG}         \\ \midrule
        \textbf{Zero-Shot}                        & -                      & 45.25          & 23.00          & 34.13       & 61.07          & 38.12          & 46.46       \\ \midrule
        \multirow{8}{*}{ANS-SFT}         & 100                    & 38.50          & 18.25          & 28.38       & 64.77          & 34.87          & 45.74       \\ 
                                        & 200                    & 32.50          & 22.75          & 27.63       & 69.46          & 35.25          & 47.68       \\ 
                                        & 400                    & -              & -              & -           & 72.48          & 40.42          & 52.07       \\ 
                                        & 600                    & 32.25          & 18.00          & 25.13       & 73.49          & 39.27          & 51.71       \\ 
                                        & 800                    & -              & -              & -           & 75.50          & 37.93          & 51.58       \\ 
                                        & 1200                   & 32.50          & 18.50          & 25.50       & 75.84          & 37.74          & 51.59       \\ 
                                        & 1600                   & 32.50          & 18.25          & 25.38       & 75.84          & 37.36          & 51.34       \\ 
                                        & 1686                   & 18.25          & 38.75          & 28.50       & 38.59          & 42.72          & 41.22       \\ \midrule
        \multirow{8}{*}{CoT-SFT}         & 100                    & 6.50           & 32.00          & 19.25       & 38.26          & 43.10          & 41.34       \\ 
                                        & 200                    & 27.00          & 34.50          & 30.75       & 56.71          & 44.64          & 49.03       \\ 
                                        & 400                    & -              & -              & -           & 52.68          & 44.06          & 47.19       \\ 
                                        & 600                    & 35.50          & 36.25          & 35.88       & 63.09          & 43.49          & 50.61       \\ 
                                        & 800                    & -              & -              & -           & 63.42          & 42.91          & 50.36       \\ 
                                        & 1200                   & 29.50          & 37.50          & 33.50       & 64.09          & 44.06          & 51.34       \\ 
                                        & 1600                   & 29.25          & 36.75          & 33.00       & 61.74          & 44.06          & 50.49       \\ 
                                        & 1686                   & 58.50          & 41.75          & 50.13       & 56.71          & 45.98          & 49.88       \\ \midrule
        \multirow{8}{*}{Reason-RFT-Zero} & 100                    & 59.00          & 44.25          & 51.63       & 63.42          & 45.21          & 51.83       \\ 
                                        & 200                    & 62.00          & 43.00          & 52.50       & 70.47          & 45.40          & 54.51       \\ 
                                        & 400                    & -              & -              & -           & 70.13          & 46.74          & 55.24       \\ 
                                        & 600                    & 64.75          & 45.25          & 55.00       & 70.47          & 49.23          & 56.95       \\ 
                                        & 800                    & -              & -              & -           & 66.11          & 46.17          & 53.42       \\ 
                                        & 1200                   & 69.00          & 43.25          & 56.13       & 71.14          & 45.59          & 54.88       \\ 
                                        & 1600                   & 66.25          & 43.25          & 54.75       & 69.80          & 46.55          & 55.00       \\ 
                                        & 1610                   & 46.75          & 37.50          & 42.13       & 67.79          & 45.79          & 53.79       \\ \midrule
        \multirow{8}{*}{Reason-RFT}      & 100                    & 53.00          & 37.00          & 45.00       & 72.82          & 46.93          & 56.34       \\ 
                                        & 200                    & 52.75          & 37.25          & 45.00       & 71.14          & 46.55          & 55.49       \\ 
                                        & 400                    & 51.50          & 37.00          & 44.25       & 73.49          & 48.28          & 57.44       \\ 
                                        & 600                    & 56.75          & 37.25          & 47.00       & 77.52          & 46.17          & 57.56       \\ 
                                        & 800                    & 59.00          & 40.00          & 49.50       & 79.87          & 48.08          & 59.63       \\ 
                                        & 1200                   & 56.00          & 39.50          & 47.75       & 74.50          & 49.62          & 58.66       \\ 
                                        & 1600                   & 59.00          & 39.50          & 49.25       & 78.52          & 48.28          & 59.27       \\ 
                                        & 1610                   & 59.00          & 39.50          & 49.25       & 78.52          & 48.28          & 59.27       \\ 
        \bottomrule
    \end{tabular}}
    \caption{Complete experimental results of Qwen2VL-7B-Instruct on the Structure Perception task after training on GeoMath.}
    \label{tab:structure_perception_7b}
\end{table*}

\begin{sidewaystable*}[htbp]
    \centering
    \scalebox{0.87}{
    \begin{tabular}{cc|ccccc|ccccc|ccccc}
        \toprule
        \multirow{3}{*}{\textbf{Method}}         & \multirow{3}{*}{\textbf{Steps}} & \multicolumn{15}{c}{\textbf{Spatial Transformation}} \\ \cmidrule(lr){3-17} 
                                        &                        & \multicolumn{5}{c}{\textbf{TRANCE (ID)}}   & \multicolumn{5}{c}{\textbf{TRANCE-L (DS-L)}}  & \multicolumn{5}{c}{\textbf{TRANCE-R (DS-R)}}  \\ 
        \cmidrule(lr){3-17}
                                      &                        & \textbf{Level-1} & \textbf{Level-2} & \textbf{Level-3} & \textbf{Level-4} & \textbf{AVG}    & \textbf{Level-1} & \textbf{Level-2} & \textbf{Level-3} & \textbf{Level-4} & \textbf{AVG}   & \textbf{Level-1} & \textbf{Level-2} & \textbf{Level-3} & \textbf{Level-4} & \textbf{AVG}\\ \midrule
        GPT-4o                        & /                      & 47.28   & 42.96   & 40.87   & 39.08   & 42.55  & 23.16   & 30.56   & 30.73   & 30.22   & 28.67 & 24.38   & 31.74   & 31.13   & 31.77   & 29.76\\ 
        Zero-Shot                     & /                      & 2.10    & 3.27    & 4.08    & 5.68    & 3.78   & 2.02    & 4.73    & 5.57    & 6.08    & 4.60  & 2.39    & 4.73    & 5.59    & 5.95    & 4.67\\ \midrule
        \multirow{7}{*}{ANS-SFT}      & 100                    & 15.90   & 19.12   & 14.67   & 13.12   & 15.70  & 10.60   & 11.33   & 10.01   & 9.07    & 10.25 & 11.08   & 12.17   & 10.17   & 10.07   & 10.87\\ 
                                      & 200                    & 23.97   & 29.56   & 33.98   & 33.95   & 30.37  & 13.25   & 26.76   & 31.26   & 32.58   & 25.96 & 13.53   & 26.54   & 30.53   & 31.20   & 25.45\\ 
                                      & 400                    & 44.95   & 42.58   & 40.75   & 33.65   & 40.48  & 26.03   & 35.98   & 34.26   & 29.38   & 31.41 & 24.06   & 35.62   & 35.46   & 31.73   & 31.72\\ 
                                      & 800                    & 62.10   & 56.55   & 53.01   & 47.55   & 54.80  & 24.70   & 42.27   & 43.05   & 42.65   & 38.17 & 26.09   & 38.98   & 42.95   & 42.55   & 37.64\\ 
                                      & 1600                   & 80.70   & 75.68   & 68.34   & 64.60   & 72.33  & 34.38   & 49.00   & 53.83   & 53.10   & 47.58 & 35.84   & 49.28   & 51.23   & 52.50   & 47.21\\ 
                                      & 3200                   & 82.85   & 80.30   & 78.00   & 71.60   & 78.19  & 36.22   & 52.61   & 55.23   & 54.27   & 49.58 & 38.51   & 52.78   & 54.47   & 53.98   & 49.94\\ 
                                      & final                  & 82.70   & 79.93   & 76.70   & 70.22   & 77.39  & 36.00   & 52.82   & 54.59   & 53.55   & 49.24 & 39.63   & 53.75   & 54.33   & 53.60   & 50.33\\ \midrule
        \multirow{7}{*}{COT-SFT}      & 100                    & 6.99    & 14.90   & 15.99   & 20.36   & 14.56  & 10.32   & 13.74   & 11.82   & 14.69   & 12.64 & 6.97    & 13.38   & 12.84   & 13.41   & 11.65\\ 
                                      & 200                    & 15.45   & 19.12   & 14.53   & 16.46   & 16.39  & 12.90   & 17.51   & 14.22   & 15.82   & 15.11 & 11.23   & 17.71   & 13.87   & 16.32   & 14.78\\ 
                                      & 400                    & 25.98   & 26.74   & 19.94   & 16.02   & 22.17  & 15.73   & 21.19   & 17.73   & 15.96   & 17.65 & 16.13   & 20.55   & 17.91   & 15.97   & 17.64\\ 
                                      & 800                    & 43.85   & 43.19   & 41.77   & 37.84   & 41.66  & 22.88   & 38.29   & 37.54   & 35.29   & 33.50 & 22.74   & 35.81   & 37.60   & 36.23   & 33.10\\ 
                                      & 1600                   & 52.82   & 61.06   & 54.38   & 45.85   & 53.53  & 28.65   & 43.95   & 40.99   & 40.13   & 38.43 & 29.00   & 41.23   & 40.78   & 39.26   & 37.57\\ 
                                      & 3200                   & 61.40   & 69.15   & 65.32   & 62.28   & 64.54  & 28.67   & 45.97   & 50.06   & 52.10   & 44.20 & 31.19   & 45.87   & 45.92   & 51.35   & 43.58\\ 
                                      & final                  & 67.47   & 67.52   & 62.78   & 59.70   & 64.37  & 28.87   & 44.41   & 49.16   & 50.30   & 43.19 & 30.20   & 44.77   & 47.15   & 49.33   & 42.86\\ \midrule
        \multirow{7}{*}{Reason-RFT-Zero} & 100                 & 8.44    & 17.96   & 20.69   & 26.22   & 18.33  & 8.53    & 17.42   & 21.16   & 25.05   & 18.04 & 8.08    & 18.12   & 21.09   & 25.70   & 18.25\\ 
                                      & 200                    & 9.59    & 18.76   & 22.97   & 28.73   & 20.01  & 9.49    & 20.08   & 23.19   & 27.00   & 19.94 & 9.72    & 18.97   & 22.82   & 28.50   & 20.00\\ 
                                      & 400                    & 12.35   & 21.47   & 27.01   & 26.25   & 21.77  & 11.10   & 21.47   & 25.73   & 25.30   & 20.90 & 10.54   & 21.19   & 25.44   & 25.60   & 20.69\\ 
                                      & 800                    & 18.47   & 32.08   & 32.77   & 27.85   & 27.79  & 15.40   & 29.12   & 30.93   & 27.38   & 25.71 & 15.52   & 27.75   & 31.50   & 27.88   & 25.66\\ 
                                      & 1600                   & 36.78   & 40.20   & 37.78   & 34.51   & 37.32  & 19.96   & 33.03   & 35.84   & 34.49   & 30.83 & 20.39   & 32.85   & 33.87   & 33.90   & 30.25\\ 
                                      & 3200                   & 43.72   & 46.89   & 44.07   & 40.50   & 43.80  & 18.67   & 34.11   & 37.85   & 39.69   & 32.58 & 18.08   & 34.01   & 37.27   & 40.27   & 32.41\\ 
                                      & final                  & 46.21   & 45.01   & 44.53   & 42.11   & 44.47  & 18.33   & 34.57   & 37.94   & 40.57   & 32.85 & 18.28   & 33.45   & 37.44   & 40.34   & 32.38\\ \midrule
        \multirow{7}{*}{Reason-RFT}      & 100                 & 53.52   & 55.47   & 58.91   & 53.35   & 55.31  & 31.84   & 47.02   & 50.62   & 50.39   & 44.97 & 31.29   & 46.08   & 48.61   & 49.85   & 43.96\\ 
                                      & 200                    & 54.97   & 59.77   & 63.67   & 59.46   & 59.47  & 35.72   & 49.28   & 54.85   & 54.67   & 48.63 & 36.74   & 52.27   & 53.68   & 54.16   & 49.21\\ 
                                      & 400                    & 63.80   & 66.97   & 68.47   & 64.70   & 65.99  & 39.74   & 55.94   & 61.27   & 57.94   & 53.72 & 41.10   & 56.35   & 59.16   & 57.22   & 53.46\\ 
                                      & 800                    & 64.33   & 68.13   & 66.88   & 63.15   & 65.62  & 47.40   & 61.64   & 63.00   & 58.45   & 57.62 & 46.76   & 60.60   & 61.40   & 59.91   & 57.17\\ 
                                      & 1600                   & 76.47   & 73.42   & 74.05   & 69.16   & 73.28  & 52.68   & 62.22   & 66.56   & 64.71   & 61.54 & 53.47   & 65.17   & 64.98   & 63.52   & 61.79\\ 
                                      & 3200                   & 72.88   & 74.85   & 75.77   & 72.45   & 73.99  & 52.58   & 63.60   & 68.51   & 66.01   & 62.68 & 52.67   & 65.60   & 67.60   & 65.66   & 62.88\\ 
                                      & final                  & 74.10   & 74.52   & 76.68   & 73.12   & 74.61  & 53.49   & 65.72   & 69.64   & 67.34   & 64.05 & 54.95   & 66.25   & 68.32   & 66.80   & 64.08\\ 
        \bottomrule
    \end{tabular}}
    \caption{Complete experimental results of Qwen2VL-2B-Instruct on the Spatial Transformation task after training on TRANCE.}
    \label{tab:trance_2b}
\end{sidewaystable*}

\begin{sidewaystable*}[htbp]
    \centering
    \scalebox{0.87}{
    \begin{tabular}{cc|ccccc|ccccc|ccccc}
        \toprule
        \multirow{3}{*}{\textbf{Method}}         & \multirow{3}{*}{\textbf{Steps}} & \multicolumn{15}{c}{\textbf{Spatial Transformation}} \\ \cmidrule(lr){3-17} 
                                        &                        & \multicolumn{5}{c}{\textbf{TRANCE (ID)}}   & \multicolumn{5}{c}{\textbf{TRANCE-L (DS-L)}}  & \multicolumn{5}{c}{\textbf{TRANCE-R (DS-R)}}  \\ 
        \cmidrule(lr){3-17}
                                      &                        & \textbf{Level-1} & \textbf{Level-2} & \textbf{Level-3} & \textbf{Level-4} & \textbf{AVG}    & \textbf{Level-1} & \textbf{Level-2} & \textbf{Level-3} & \textbf{Level-4} & \textbf{AVG}   & \textbf{Level-1} & \textbf{Level-2} & \textbf{Level-3} & \textbf{Level-4} & \textbf{AVG}\\ \midrule
        GPT-4o                        & /                      & 47.28   & 42.96   & 40.87   & 39.08   & 42.55  & 23.16   & 30.56   & 30.73   & 30.22   & 28.67  & 24.38   & 31.74   & 31.13   & 31.77   & 29.76\\ 
        Zero-Shot                     & /                      & 16.25   & 16.42   & 10.96   & 10.48   & 13.53  & 11.71   & 16.80   & 11.50   & 10.85   & 12.72  & 13.30   & 16.08   & 10.55   & 11.18   & 12.78\\ \midrule
        \multirow{7}{*}{ANS-SFT}      & 100                    & 40.30   & 37.05   & 30.67   & 28.35   & 34.09  & 32.07   & 31.12   & 26.00   & 26.80   & 29.00  & 26.38   & 29.71   & 27.74   & 26.48   & 27.58\\ 
                                      & 200                    & 65.18   & 53.33   & 49.43   & 45.15   & 53.27  & 33.29   & 45.14   & 45.61   & 45.52   & 42.39  & 35.17   & 43.43   & 45.97   & 43.02   & 41.90\\ 
                                      & 400                    & 65.33   & 59.35   & 57.17   & 50.77   & 58.16  & 32.40   & 44.13   & 47.69   & 46.23   & 42.61  & 32.10   & 44.88   & 47.04   & 45.25   & 42.32\\ 
                                      & 800                    & 78.90   & 70.67   & 63.97   & 62.10   & 68.91  & 34.08   & 50.62   & 51.99   & 52.95   & 47.41  & 34.22   & 50.40   & 50.62   & 52.88   & 47.03\\ 
                                      & 1600                   & 78.50   & 76.12   & 73.80   & 66.25   & 73.67  & 38.85   & 52.97   & 57.93   & 56.05   & 51.45  & 37.77   & 53.57   & 56.45   & 55.92   & 50.93\\ 
                                      & 3200                   & 83.80   & 83.23   & 82.83   & 78.17   & 82.01  & 40.10   & 56.02   & 61.02   & 59.90   & 54.26  & 40.78   & 55.06   & 61.67   & 60.98   & 54.62\\ 
                                      & final                  & 83.70   & 84.10   & 82.50   & 78.45   & 82.19  & 39.67   & 55.58   & 61.84   & 60.05   & 54.29  & 42.64   & 54.84   & 61.44   & 60.38   & 54.83\\ \midrule
        \multirow{7}{*}{COT-SFT}      & 100                    & 20.58   & 28.98   & 25.97   & 30.00   & 26.38  & 21.89   & 29.94   & 29.18   & 29.33   & 27.59  & 18.49   & 28.43   & 30.43   & 30.36   & 26.93\\ 
                                      & 200                    & 41.80   & 44.08   & 46.02   & 42.16   & 43.52  & 25.31   & 36.28   & 36.81   & 38.86   & 34.32  & 21.19   & 34.55   & 37.06   & 37.54   & 32.59\\ 
                                      & 400                    & 45.39   & 51.32   & 58.20   & 52.42   & 51.83  & 32.53   & 44.61   & 47.73   & 46.10   & 42.74  & 31.87   & 38.96   & 44.62   & 45.90   & 40.34\\ 
                                      & 800                    & 54.87   & 61.97   & 62.20   & 59.93   & 59.74  & 30.19   & 46.01   & 50.01   & 52.88   & 44.77  & 29.91   & 45.02   & 49.60   & 52.83   & 44.34\\ 
                                      & 1600                   & 71.27   & 71.14   & 72.82   & 69.93   & 71.29  & 28.82   & 46.43   & 51.01   & 58.94   & 46.30  & 29.08   & 45.25   & 52.24   & 58.31   & 46.22\\ 
                                      & 3200                   & 84.13   & 80.62   & 79.99   & 78.42   & 80.79  & 29.93   & 47.63   & 56.49   & 62.54   & 49.15  & 30.46   & 47.85   & 54.83   & 61.02   & 48.54\\ 
                                      & final                  & 86.50   & 79.43   & 80.54   & 78.77   & 81.31  & 28.07   & 47.54   & 54.42   & 61.58   & 47.90  & 29.69   & 45.32   & 54.69   & 61.48   & 47.80\\ \midrule
        \multirow{7}{*}{Reason-RFT-Zero} & 100                 & 23.59   & 31.62   & 33.22   & 31.27   & 29.93  & 15.88   & 26.86   & 28.13   & 30.19   & 25.27  & 15.21   & 27.29   & 27.54   & 29.88   & 24.98\\ 
                                      & 200                    & 35.06   & 39.45   & 36.80   & 34.77   & 36.52  & 20.39   & 30.22   & 31.15   & 31.20   & 28.24  & 18.10   & 29.27   & 30.57   & 30.81   & 27.19\\ 
                                      & 400                    & 25.28   & 40.78   & 41.70   & 35.35   & 35.78  & 20.20   & 39.28   & 35.43   & 33.44   & 32.09  & 21.72   & 39.47   & 37.63   & 32.79   & 32.90\\ 
                                      & 800                    & 50.18   & 51.55   & 50.43   & 46.06   & 49.56  & 35.44   & 46.15   & 45.65   & 39.55   & 41.70  & 33.90   & 45.89   & 46.48   & 41.57   & 41.96\\ 
                                      & 1600                   & 59.60   & 61.90   & 57.30   & 55.36   & 58.54  & 43.95   & 55.03   & 52.96   & 50.60   & 50.64  & 41.28   & 56.46   & 51.64   & 49.08   & 49.62\\ 
                                      & 3200                   & 62.50   & 68.53   & 68.79   & 66.22   & 66.51  & 42.54   & 58.05   & 58.97   & 60.10   & 54.92  & 42.56   & 56.93   & 60.02   & 60.21   & 54.93\\ 
                                      & final                  & 65.63   & 68.30   & 69.45   & 67.30   & 67.67  & 46.61   & 58.22   & 61.69   & 62.26   & 57.20  & 45.53   & 58.40   & 61.81   & 58.85   & 56.15\\ \midrule
        \multirow{7}{*}{Reason-RFT}   & 100                    & 60.07   & 62.95   & 68.53   & 62.78   & 63.58  & 37.09   & 52.54   & 60.21   & 54.99   & 51.21  & 35.26   & 52.56   & 57.30   & 55.42   & 50.14\\ 
                                      & 200                    & 67.69   & 66.98   & 69.53   & 66.00   & 67.55  & 36.26   & 53.94   & 60.20   & 60.05   & 52.61  & 34.33   & 54.21   & 58.25   & 60.79   & 51.90\\ 
                                      & 400                    & 74.72   & 71.31   & 73.62   & 69.14   & 72.20  & 36.11   & 56.00   & 62.25   & 63.01   & 54.34  & 36.95   & 55.72   & 60.82   & 64.88   & 54.59\\ 
                                      & 800                    & 72.57   & 71.04   & 75.58   & 73.26   & 73.11  & 38.93   & 56.86   & 64.10   & 63.69   & 55.90  & 35.06   & 54.39   & 60.82   & 64.88   & 53.79\\ 
                                      & 1600                   & 79.18   & 76.70   & 76.59   & 73.26   & 76.43  & 37.25   & 53.44   & 66.38   & 66.76   & 55.96  & 35.01   & 55.39   & 66.02   & 66.67   & 55.77\\ 
                                      & 3200                   & 79.70   & 79.22   & 80.13   & 76.61   & 78.92  & 37.63   & 55.73   & 65.24   & 68.02   & 56.66  & 37.18   & 56.34   & 63.90   & 68.89   & 56.58\\ 
                                      & final                  & 80.48   & 79.15   & 81.97   & 78.26   & 79.97  & 43.68   & 61.04   & 64.66   & 68.05   & 59.36  & 45.45   & 56.34   & 63.33   & 69.33   & 58.61\\ 
        \bottomrule
    \end{tabular}}
    \caption{Complete experimental results of Qwen2VL-7B-Instruct on the Spatial Transformation task after training on TRANCE.}
    \label{tab:trance_7b}
\end{sidewaystable*}

\end{document}